# A Constraint Satisfaction Framework for Executing Perceptions and Actions in Diagrammatic Reasoning

**Bonny Banerjee** BANERJEE.28@OSU.EDU
**B. Chandrasekaran** CHANDRA@CSE.OHIO-STATE.EDU
*Laboratory for Artificial Intelligence Research*
*Department of Computer Science & Engineering*
*The Ohio State University, Columbus, OH 43210, USA*

## Abstract

Diagrammatic reasoning (DR) is pervasive in human problem solving as a powerful adjunct to symbolic reasoning based on language-like representations. The research reported in this paper is a contribution to building a general purpose DR system as an extension to a SOAR-like problem solving architecture. The work is in a framework in which DR is modeled as a process where subtasks are solved, as appropriate, either by inference from symbolic representations or by interaction with a diagram, i.e., perceiving specified information from a diagram or modifying/creating objects in a diagram in specified ways according to problem solving needs. The perceptions and actions in most DR systems built so far are hand-coded for the specific application, even when the rest of the system is built using the general architecture. The absence of a general framework for executing perceptions/actions poses as a major hindrance to using them opportunistically – the essence of open-ended search in problem solving.

Our goal is to develop a framework for executing a wide variety of specified perceptions and actions across tasks/domains without human intervention. We observe that the domain/task-specific visual perceptions/actions can be transformed into domain/task-independent spatial problems. We specify a spatial problem as a quantified constraint satisfaction problem in the real domain using an open-ended vocabulary of properties, relations and actions involving three kinds of diagrammatic objects – points, curves, regions. Solving a spatial problem from this specification requires computing the equivalent simplified quantifier-free expression, the complexity of which is inherently doubly exponential. We represent objects as configuration of simple elements to facilitate decomposition of complex problems into simpler and *similar* subproblems. We show that, if the symbolic solution to a subproblem can be expressed concisely, quantifiers can be eliminated from spatial problems in low-order polynomial time using similar previously solved subproblems. This requires determining the similarity of two problems, the existence of a mapping between them computable in polynomial time, and designing a memory for storing previously solved problems so as to facilitate search. The efficacy of the idea is shown by time complexity analysis. We demonstrate the proposed approach by executing perceptions and actions involved in DR tasks in two army applications.

## 1. Introduction

The research reported in this paper is a contribution to building problem solving agents in artificial intelligence (AI) that use diagrams, much as people do, but most of AI does not, given the almost exclusive emphasis in AI on language-like or predicate-symbolic representations. Diagrammatic reasoning (DR) is an emerging area of research in a number of fields, including AI (Glasgow, Narayanan, & Chandrasekaran, 1995; Chandrasekaran, Kurup, & Banerjee, 2005), logic (Barwise & Etchemendy, 1998; Allwein & Barwise, 1999), and psychology (Tversky, 2000; Tricket & Trafton, 2006). While all research in DR is in one way or other dealing with diagrams, different research issues are addressed by different researchers. The research reported in this paper considers DR as a problem solving activity in which an agent (human or artificial) makes use of two forms of representation – a spatial representation in the form of 2D diagrams and a symbolic representation that contains information in a predicate-symbolic form similar to logic and natural language. A schematic DR architecture, as proposed by Chandrasekaran et al. (2002, 2004, 2005), is illustrated in Figure 1.

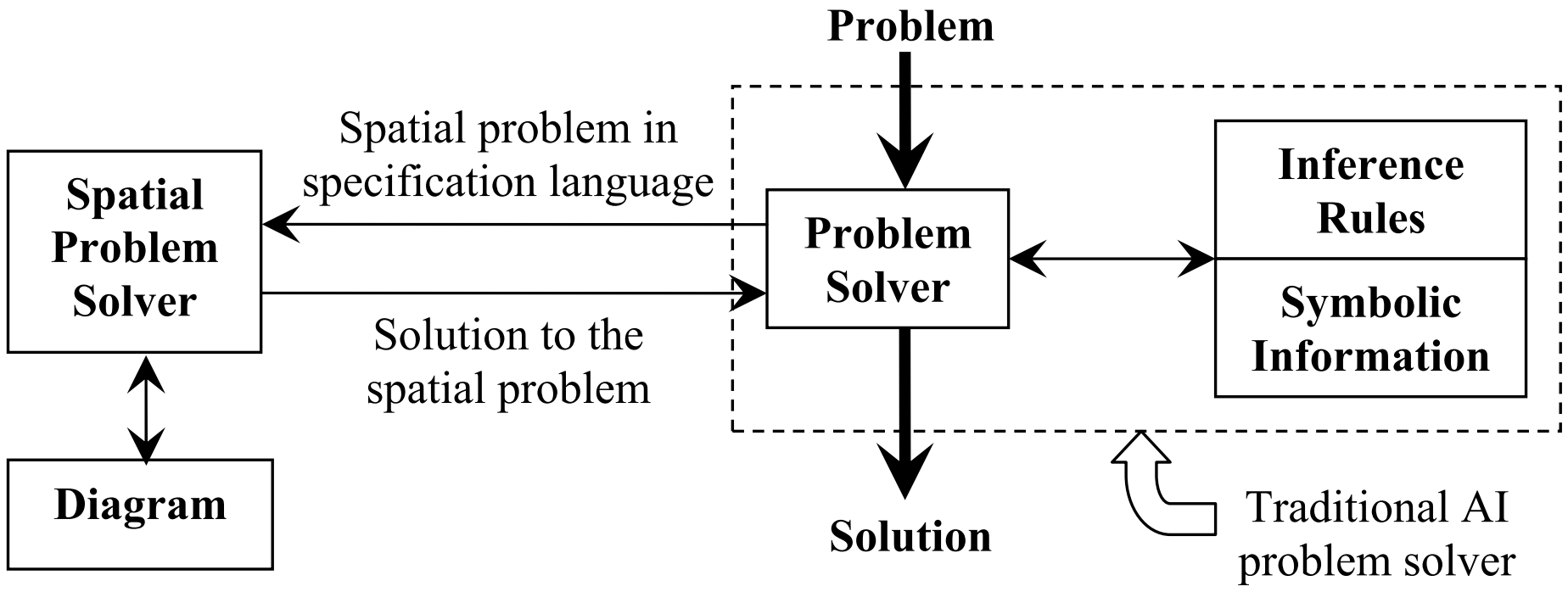

Figure 1: The diagrammatic reasoning architecture.

### 1.1 Diagrammatic Reasoning as a Problem Solving Activity

The DR architecture shares the idea of problem solving as search in problem state space (Laird, Rosenbloom, & Newell, 1986; Newell, 1990). In this approach, starting from an initial state, the agent applies operators to bring about state transitions to reach the goal state. A goal is either reached or decomposed into subgoals by the use of general and domain knowledge. Reaching a goal or subgoal requires information which is generated in the traditional problem solving architectures (e.g., SOAR in Laird, Newell, & Rosenbloom, 1987, ACT-R in Anderson, 1993) by inference using predicate-symbolic representation. In the DR architecture, the agent can extract information from diagrams by applying perception-like operations in addition to inference using predicate-symbolic representation to reach the goal/subgoal. The agent can also create or modify objects in a diagram that propose new states from which the goal might be reached with subsequent perceptions and inferences.

To illustrate our conceptualization of DR, let us consider a real-world problem. An army commander, planning strategic operations, uses a terrain map to chalk out a path for his troops to safely travel from one base camp location $L_1$ to another $L_2$ within a given time. The information he has is regarding the nature of the terrain (e.g., slow-go or no-go regions, altitude of different parts of the terrain, the speed at which his troops can travel in different kinds of terrain) and an estimate of the maximum firepower range of the enemy. The commander, being a veteran in the field, is well aware of the possibility that his troops might be ambushed along any path by the enemy who might be hiding in the neighboring regions. His problem solving might proceed as follows. A diagram consisting of the part of the terrain map of interest for this particular problem is given, along with the peripheries of the no-go regions and the two points, $L_1$ and $L_2$ (see Figure 2(a)). The commander draws one of the shorter paths from $L_1$ to $L_2$ maintaining a maximum distance from the neighboring no-go regions (see Figure 2(b)). He knows what kinds of spatial relations between points on the route and the points where the enemy could be hiding correspond to ambush potential. He then uses that knowledge to perceive (and mark) the portions of the path that are prone to ambush due to enemies hiding behind the neighboring no-go regions (see Figure 2(c)). If no such portion is found, the path is inferred to be safe. If the length of a safe path can be traversed in the given time, it is considered a suitable path for the operation. If the path drawn is not safe or does not satisfy the time constraint, another path is drawn (see Figure 2(d)) and analyzed. This procedure continues until all paths have been exhausted. If a suitable path is still not found, the least risky path might be considered for the operation. In the worst case, the commander might infer that this operation is not possible. A problem in a similar vein, as described above, has been considered by Forbus, Usher, and Chapman (2003).

In the above example, it is noteworthy how the problem solver (the commander) opportunistically brings together symbolic knowledge (such as, the firepower range of enemies) and perception and action on a diagram to solve a real-world problem. Such a phenomenon is characteristic of DR whenever it is used to solve problems in different domains, such as, economics, geometry, engineering, computer-aided design, military, and so on. We observe that executing the perceptions and actions require solving purely spatial problems with no involvement of domain knowledge. These spatial problems can be described in terms of diagrammatic objects, such as, points, curves, and regions, and spatial properties (e.g., *length* of a curve) and relations (e.g., point *on* a curve) involving them. For example, perceiving the portions of a path prone to ambush due to enemies hiding behind a mountain range requires computing the set of points $q$ on a curve (the path) $c_1$ such that $q$ is within a specified distance (the firepower range) $d$ from some point $p$ on a curve (the mountain range) $c_2$ (see Figure 22(b)). Formally, this can be written as

$$RiskyPortionsofPath(q, c_1, c_2, d) \equiv On(q, c_1) \wedge \exists p, On(p, c_2) \wedge DistanceLessThan(p, q, d)$$

$$DistanceLessThan(p, q, d) \equiv Distance(p, q) \leq d$$

where $p$ is a point. In this paper, we propose a general and efficient framework for spatial problem solving to autonomously execute perceptions and actions in DR.

### 1.2 What Do We Mean by a "Diagram"?

**Definition 1.** ***Diagram.*** *A diagram $\mathcal{D}$ is a set of labeled 2D objects $\{\mathcal{O}_1, \mathcal{O}_2, ...\mathcal{O}_n\}$ all located clearly inside (i.e., no intersection or touching) a common region (or bounding box) $\mathcal{B}$. The objects are of three types – points, curves, regions.*

**Definition 2.** ***Diagrammatic Object.*** *A diagrammatic object $\mathcal{O}$ is a 3-tuple $< \mathcal{L}, \mathcal{T}, \mathcal{E} >$ where $\mathcal{L}$ is a label, $\mathcal{T}$ is a type (point, curve or region), and $\mathcal{E}$ is its spatial extent. The spatial extent of a diagrammatic object is the set of points constituent of the object.*

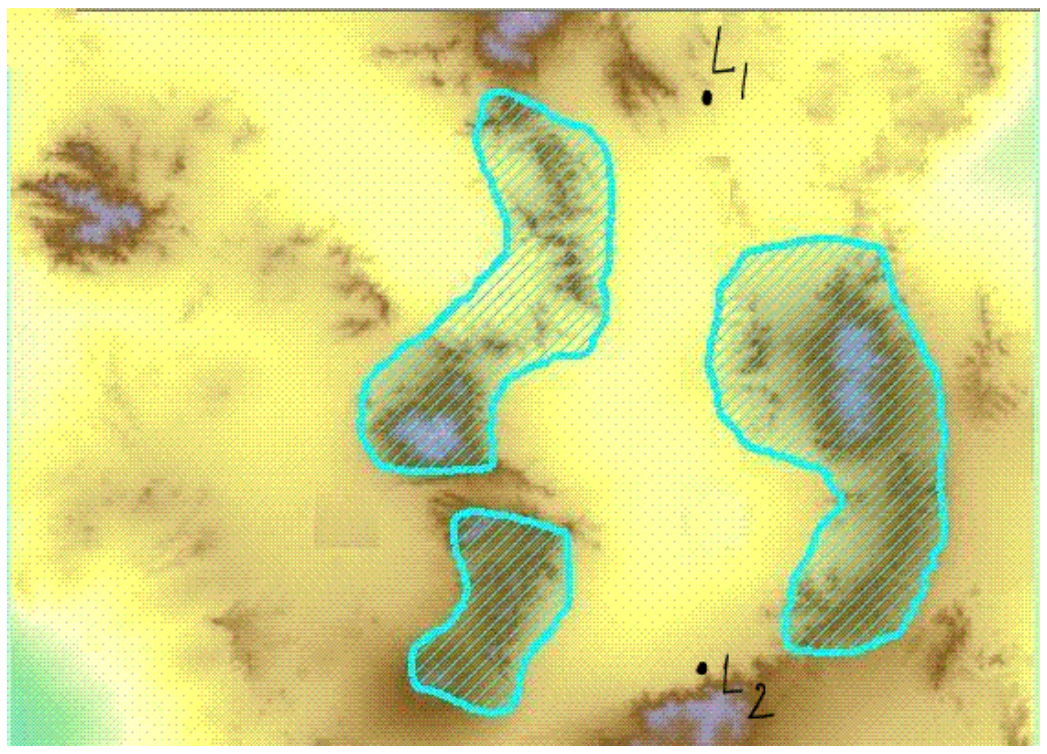

(a) The given diagram consisting of two points, $L_1$ and $L_2$, and three region obstacles.

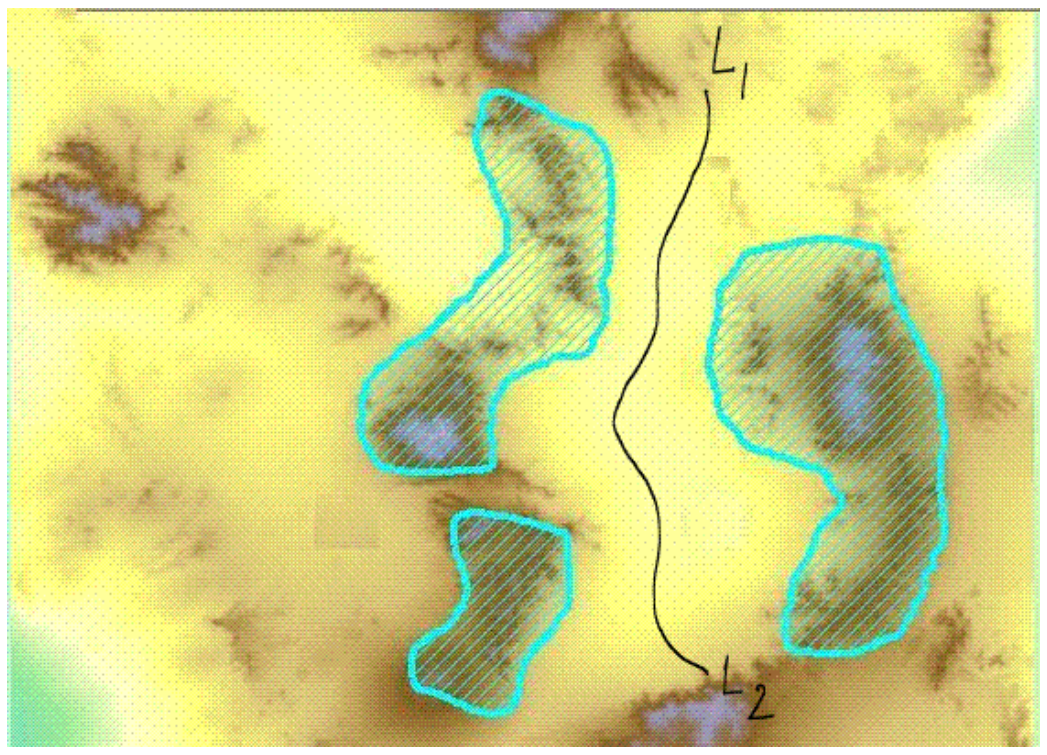

(b) One of the shorter paths between $L_1$ and $L_2$ avoiding the obstacles is drawn.

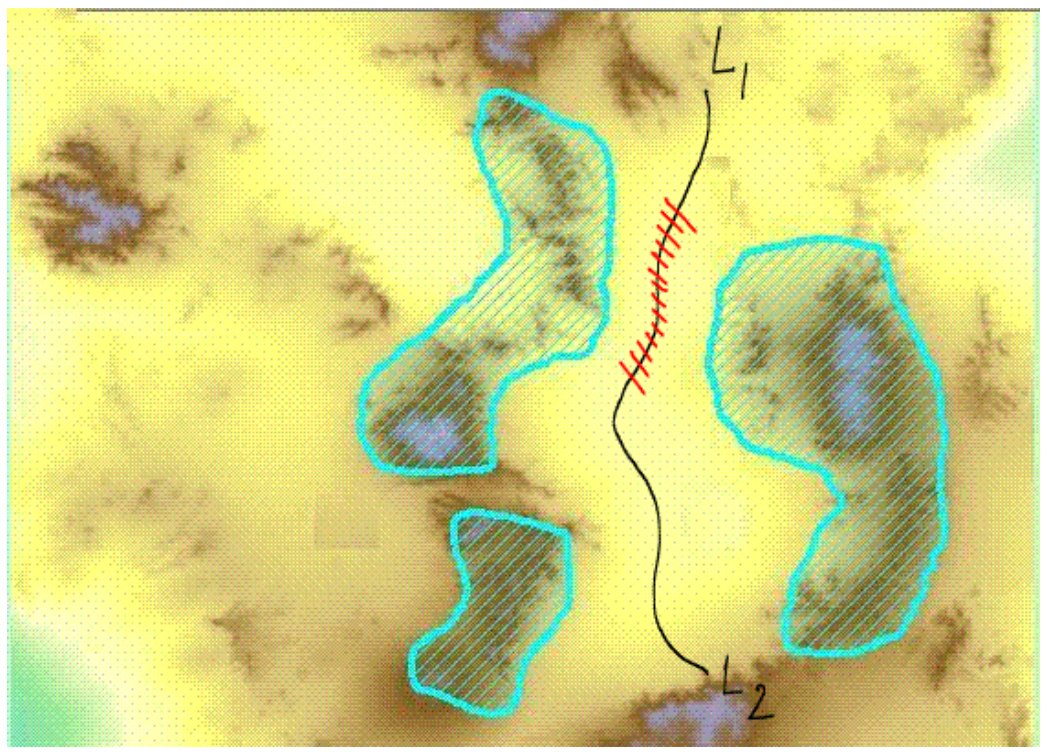

(c) Portions of the path prone to ambush are perceived and marked.

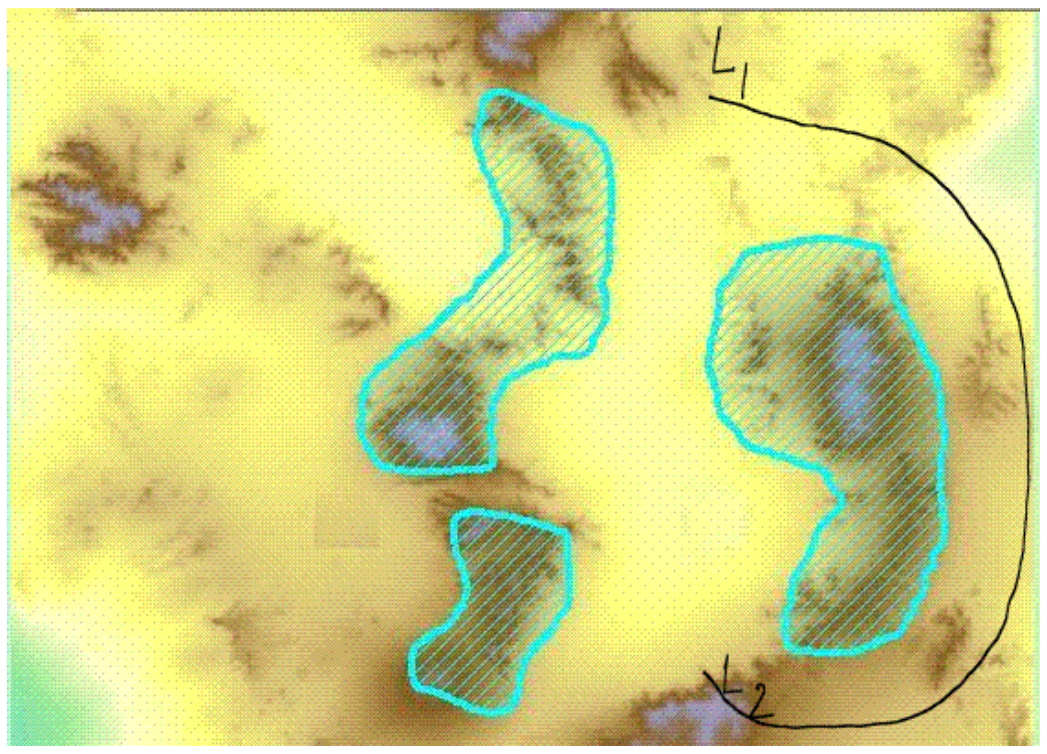

(d) Another path is drawn and will be analyzed for risk.

Figure 2: Diagrammatic reasoning by an army commander for finding a safe path for transporting his troops from $L_1$ to $L_2$ within a given time.

**Definition 3.** ***Diagrammatic Image.*** *The diagrammatic image, $\mathcal{I}$, of a diagram is the set of points constituent of all objects in the diagram. Thus, if $\mathcal{D}= \{\mathcal{O}_1, \mathcal{O}_2, ...\mathcal{O}_n\}$ is a diagram where $\mathcal{O}_i = <\mathcal{L}(\mathcal{O}_i), \mathcal{T}(\mathcal{O}_i), \mathcal{E}(\mathcal{O}_i)>$, then its diagrammatic image $\mathcal{I}(\mathcal{D})$ is given by*

$$\mathcal{I}(\mathcal{D}) = \bigcup_{i \leftarrow 1}^{n} \mathcal{E}(\mathcal{O}_i)$$

This definition of a diagram, due to Chandrasekaran et al. (2002, 2004, 2005), supports the functional representation of a diagram *in* an artificial agent. A diagram on an external medium (e.g., piece of paper, computer screen) is, at one level, an image consisting of pixels with different intensities. At another level, it is an interpreted representation consisting of spatial objects in some domain of interest. The abstract diagram is ideal, i.e., the points are dimensionless, curves have no thickness, etc. In an external diagram, points and curves consist of at least one pixel with finite dimensions. We will need to interchange between the two forms of diagrams for reasoning and interaction purposes. In the rest of the paper, the term “diagram” will refer to an abstract diagram only, unless otherwise stated. We are interested only in diagrams that are line drawings with no color or intensity variation. Such diagrams form a substantial class of diagrams in everyday use.

### 1.3 Perceptions and Actions in Diagrammatic Reasoning

**Definition 4.** ***Perception.*** *A perception is an act of extracting a new piece of information from a diagram. The new piece of information satisfies constraints specified in terms of properties and relations among existing objects in the diagram and is a boolean or a real number or a diagrammatic object(s). Thus, a perception $\mathcal{P}$ is a mapping from a diagram $\mathcal{D}$ to a set of booleans $\{True, False\}$ or real numbers $\Re$ or a set of diagrammatic objects $\mathcal{D}'$ satisfying constraints $\mathcal{C}$.*

$$\mathcal{P} : \mathcal{D} \xrightarrow{\mathcal{C}} \{True, False\} \cup \Re \cup \mathcal{D}' \ , \quad \mathcal{I}(\mathcal{D}') \subseteq \mathcal{I}(\mathcal{D})$$

**Definition 5.** ***Action.*** *An action is an act of introducing a new object(s), or modifying or deleting an existing object(s) in a diagram satisfying constraints specified in terms of properties and relations among existing objects. Thus, an action $\mathcal{A}$ is a mapping from a diagram $\mathcal{D}$ to a new set of diagrammatic objects $\mathcal{D}'$ satisfying constraints $\mathcal{C}$.*

$$\mathcal{A} : \mathcal{D} \xrightarrow{\mathcal{C}} \mathcal{D}' \ , \quad \mathcal{I}(\mathcal{D}) \neq \mathcal{I}(\mathcal{D}')$$

In the last couple of decades, numerous DR systems have been built for different applications in different domains. In the following we review some well-known DR systems where a problem solving agent reasons using diagrams. This review will help realize the role of perception and action in DR, and the spatial problems implicit in such perceptions and actions.

Sketchy (Pisan, 1995) is a computer implementation of a model of graph understanding. It recognizes the diagrammatic objects - points, lines, regions, and a vocabulary of properties and relations that includes *coordinate at point, right of, above, inside, steeper, bigger, vertical, change in slope, touches, intersects, on line, on border, forms border*, etc. for representing conceptual relationships in domains, such as, thermodynamics and economics. A domain translator is responsible for converting domain-specific conceptual questions into domain-independent graphical relations. Examples of perception from a supply-demand graph in economics include how price effects the supply, demand, and market price of the product, which requires solving visual problems, such as, “At what point is supply equal to demand?” (corresponding spatial problem: compute the intersection of two curves), “What is the price for the supply line when the quantity is 350?” (corresponding spatial problem: compute a point on a curve whose one coordinate is given), “Are the quantity and price directly proportional?” (corresponding spatial problem: check whether the slope of a curve between two points is a positive constant or not), “Are the quantity and price inversely proportional?” (corresponding spatial problem: check whether the slope of a curve between two points is a negative constant or not), etc. Actions in this model are not required due to the nature of its task. Examples of graphs understood by sketchy are shown in Figure 3.

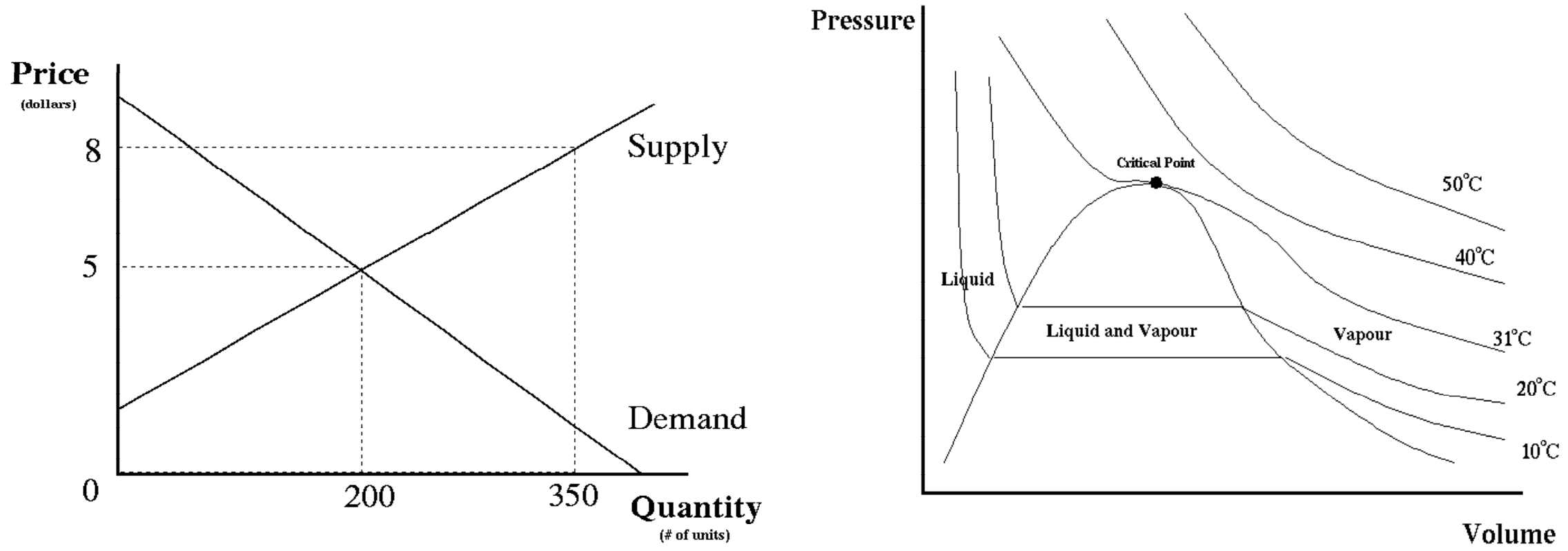

(a) Graph from economics
(b) Graph from thermodynamics

Figure 3: Examples of graphs understood by SKETCHY. Reproduced with permission from Pisan (1994).

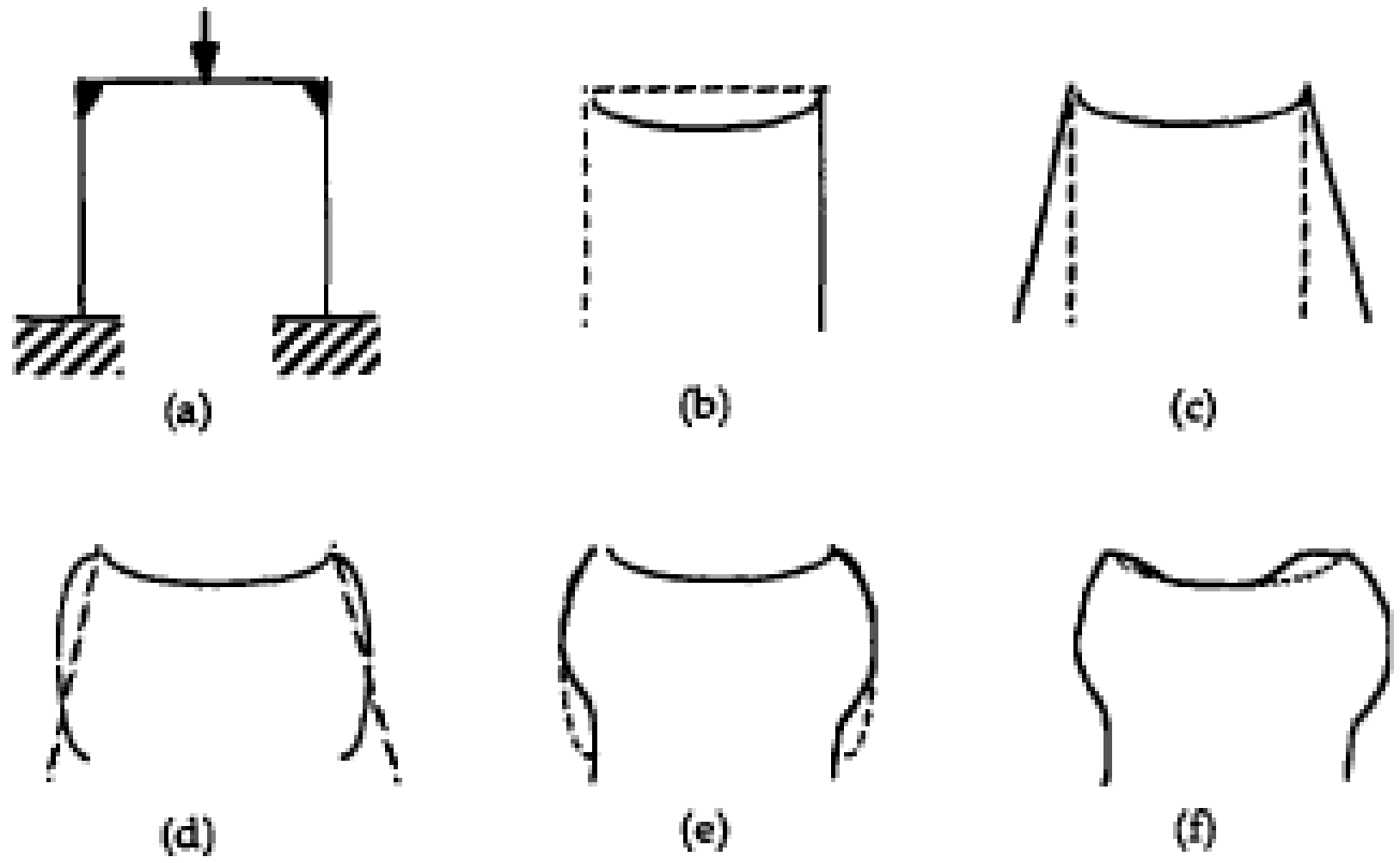

Figure 4: An example of a deflected frame analysis (from civil engineering) by REDRAW. Reproduced with permission from Tessler et al. (1995).

The REDRAW system (Tessler et al., 1995) combines diagrammatic and symbolic reasoning to qualitatively determine the deflected shape of a frame structure under a load, a structural analysis problem in civil engineering. It uses a vocabulary of properties and relations including *get-angular-displacement, get-displacement, symmetrical-p, connected-to, near, left, above, rotate, bend, translate, smooth*, etc. on three kinds of diagrammatic objects – lines, splines, circles. Though most properties and relations are domain-independent, some, such as, *bend* reflect the assumptions implicit in the domain and the task and can be defined accordingly. Perceptions and actions are called inspection and manipulation operators in the system. The underlying representation is a combination of a grid-based and Cartesian coordinates – shapes are represented using the grid where each element in the grid corresponds to a point in the diagram while lines are represented by a set of coordinate points. Examples of perception and action include deflecting a beam in the same direction as the load, checking whether a beam and column are perpendicular at a particular rigid joint, etc. which require solving visual problems, such as, "Bend Beam3 in the negative direction of the $y$-axis" (corresponding spatial problem: compute a curve with a given slope at a given point), "Make the angle between Beam3 and Column3 at Joint3 90 degrees without modifying Beam3" (corresponding spatial problem: compute a curve such that it makes a particular angle at a given point with a given curve), "Get the angle between Beam3 and Column3 at the ends connected by Joint3" (corresponding spatial problem: compute the angle between two curves at a given point), etc. An example of a deflected frame analysis by REDRAW is shown in Figure 4.

The ARCHIMEDES system (Lindsay, 1998) assists a human in demonstrating theorems in Euclidean geometry by modifying/creating diagrams according to his instructions and thereafter perceiving/inferencing from the diagram. It operates on two diagrammatic objects - points and line segments, and recognizes shapes, such as, square, triangle, path, etc. The underlying representation is array- or grid-based. The perceptions, called retrieval processes, are of different classes, such as, verify relationship, test for a condition, etc. The actions, called construction processes, are also of different classes, such as, create an object with certain properties, transform an object, etc. Executing the perceptions and actions require solving spatial problems, such as, create a segment parallel to a given segment through a given point, rotate an object and check whether it coincides with another object, etc. An example of a geometry theorem demonstrated by ARCHIMEDES is shown in Figure 5.

The DIAMOND (Jamnik, 2001), a system for proving mathematical theorems, uses a sequence of actions on diagrams assisted by a human to prove specific ground instances and then generalizes by induction. It uses a mixture of Cartesian and topological representations to represent a dot (equivalent to a point in Cartesian representation) as a diagrammatic object in the discrete space, and a line and an area (or region) as diagrammatic objects in the continuous space. Elementary shapes, such as, row, column, ell, and frame, are constructed from dots, while derived shapes, such as, square, triangle, rectangle, etc. are constructed from the elementary or other derived shapes. The vocabulary consists of atomic or one-step operations (e.g., *rotate, translate, cut, join, project from 3D to 2D, remove, insert a segment*, etc.). Spatial problems in this system are composite operations composed from the atomic ones, such as, draw a right-angled triangle, translate and rotate a triangle, etc. The system does not need to execute perceptions as information from a diagram is perceived by a human who decides what actions to be applied during the proof search. An example of a mathematical theorem proven by diamond is shown in Figure 6.

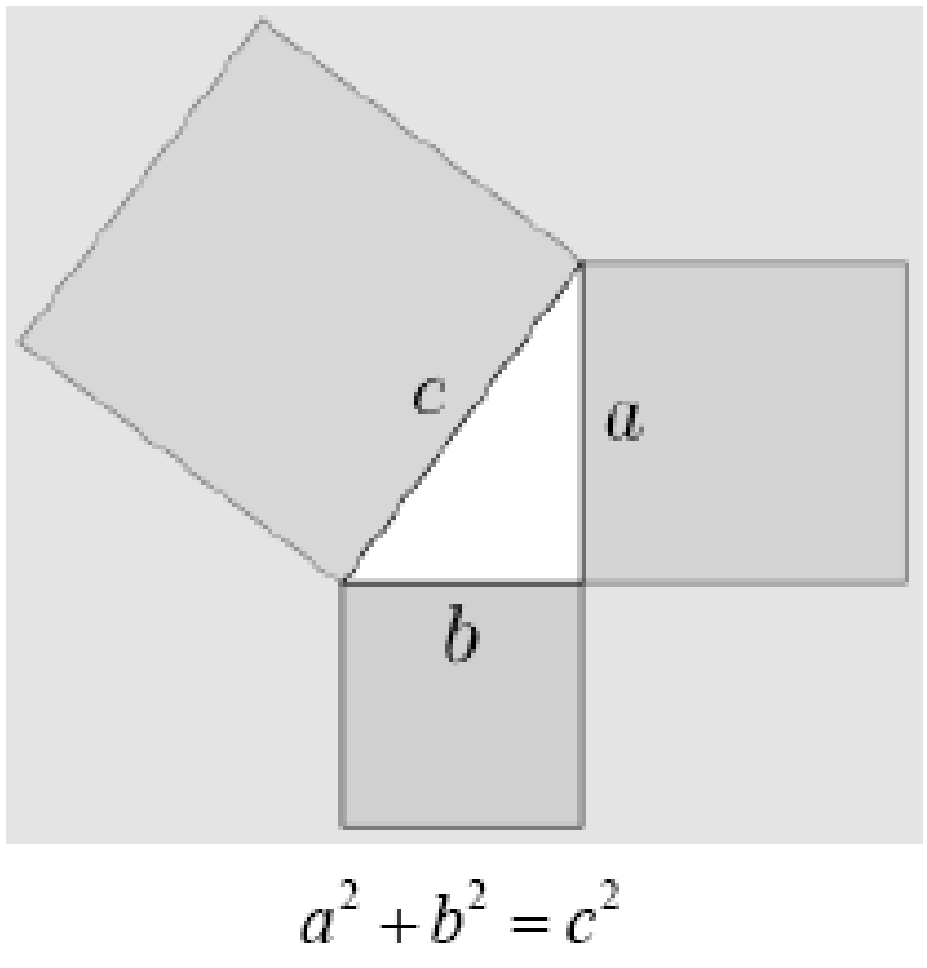

Figure 5: An example of a geometry theorem demonstrated by ARCHIMEDES.

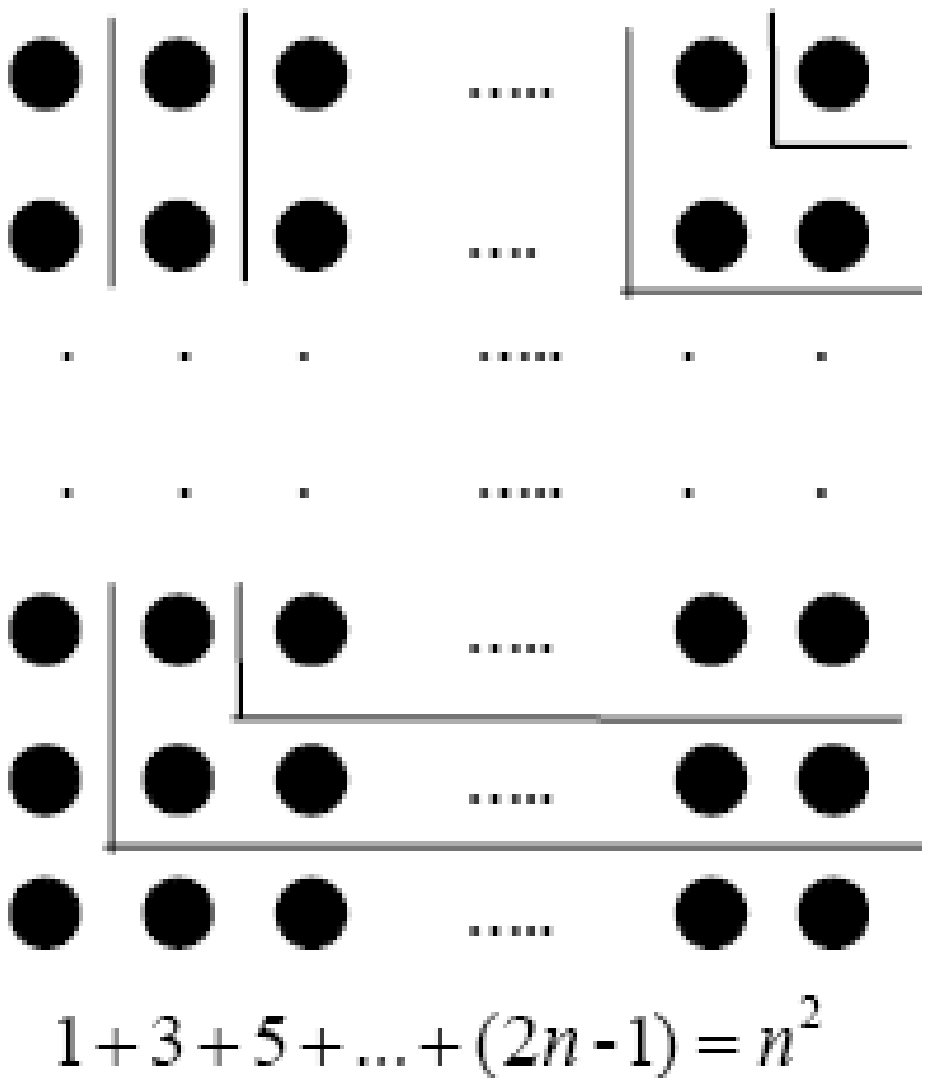

Figure 6: An example of a mathematical theorem proven by DIAMOND. The theorem is after Nelson (1993).

Georep (Ferguson & Forbus, 2000) takes as input a line drawing in vector graphics representation and creates a predicate calculus representation of the drawing's spatial relations. Five primitive shape types are recognized, namely line segments, circular arcs, circles and ellipses, splines (open and closed), and positioned text. Properties and relations, such as, proximity detection, orientation detection (e.g., *horizontal, vertical, above, beside*), parallelism, connectivity (e.g., *detecting corner, intersection, mid-connection, touch*), etc. are deployed to accomplish its task. The underlying representation is vector graphics or line drawings. Systems, such as, magi (Ferguson, 1994), juxta (Ferguson & Forbus, 1998), and coadd are built using georep for symmetry detection, critiquing diagrams based on their captions, and producing a description of the units, areas, and tasks from a course of action diagram, respectively. georep, due to the limitation of its task, does not need to execute any action. Examples of visual problems in georep include figuring out which cup contains more liquid (corresponding spatial problem: compare the areas of polygons representing the cups), determine whether a figure is symmetric or not (corresponding spatial problem: check whether one polygon is congruent to the reflection of the other polygon), etc. An example of ambush analysis by georep is shown in Figure 7.

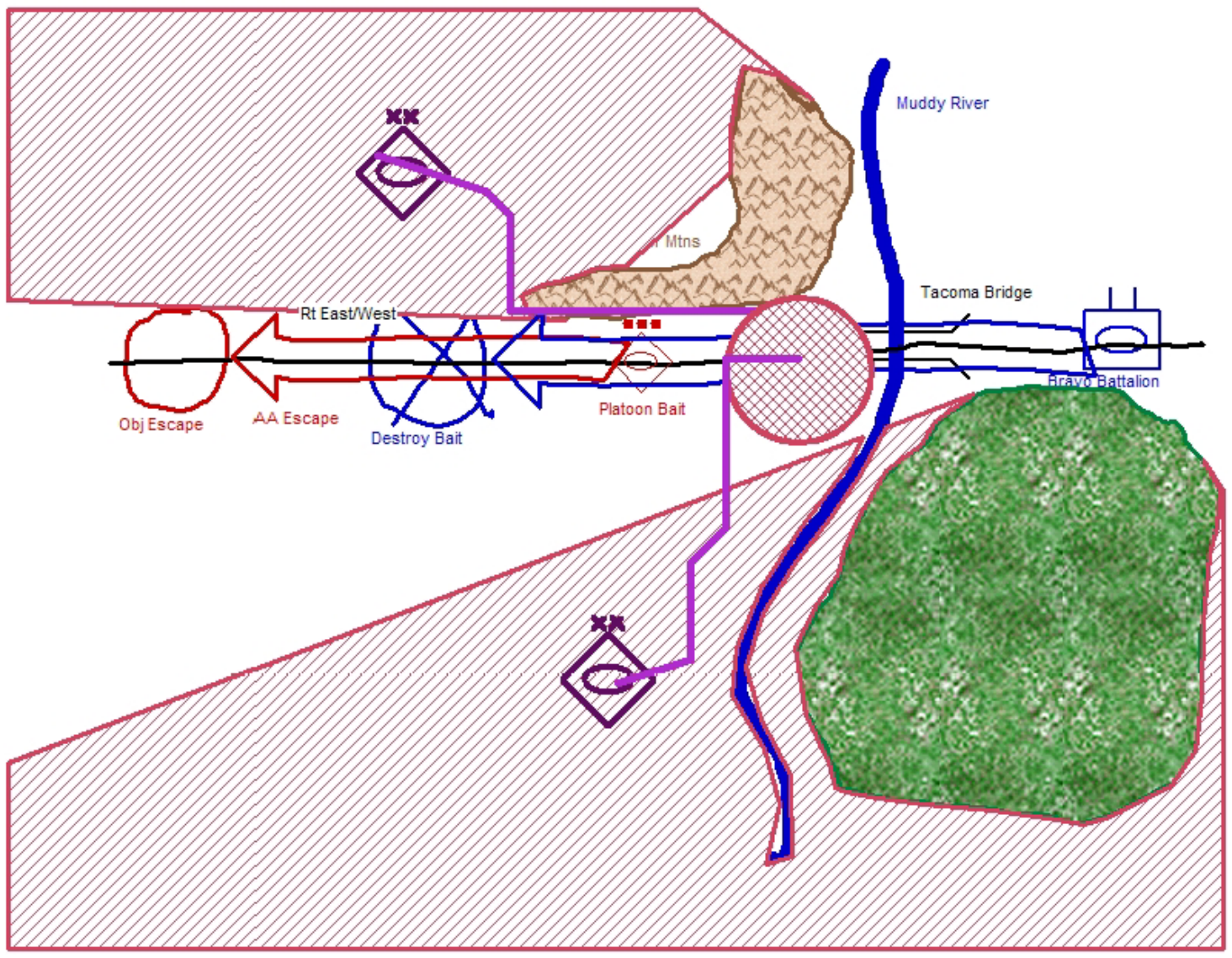

Figure 7: An example of ambush analysis by georep. Reproduced with permission from Forbus et al. (2003).

The preceding discussion leads to the observation that all DR systems require perceiving from and/or acting on diagrams, and that every perception/action requires solving a domain-independent spatial problem. Thus, a general-purpose DR system for solving problems for applications across multiple domains would require solving a large variety of non-trivial domain-independent spatial problems. These spatial problems can be described in terms of three diagrammatic objects – points, curves, regions, and spatial properties and relations involving them.

### 1.4 The Problem

How are the perceptions and actions solved in a DR system? Typically, the human developing a DR system identifies a priori the problem solving steps including a set of perceptions and actions, and hand-codes efficient algorithms for solving each of them. If the problem solving steps need to be altered in future and as a result, a new perception arises, the developer has to write another algorithm for obtaining its solution. Thus, algorithms need to be hand-coded for each perception/action. Clearly, this is inconvenient and time consuming in developing a DR system, and does not allow fast and easy experimentation with different problem solving strategies for the same problem. These drawbacks are further magnified when the goal is to build a general-purpose DR system where a very large variety of perceptions and actions are possible which is not feasible to ascertain a priori, and develop and store algorithms for. Hence, our goal is to investigate a spatial problem solver (SPS) for efficiently solving spatial problems implicit in perceptions/actions without human intervention.[1]

### 1.5 Contributions

In this paper, we make the following contributions:

1. We observe that the wide variety of visual perceptions/actions for DR applications can be transformed into domain/task-independent spatial problems. We developed a language for specifying spatial problems (i.e., spatial relations or actions) as quantified constraint satisfaction problems (QCSPs) in first-order logic using a fixed set of mathematical/logical operators in the real domain and an open-ended vocabulary of properties, relations and actions. Any spatial relation or action involving only points that can be expressed using those operators and real variables in first-order logic can be included in the vocabulary. Further, any spatial relation or action involving curves and/or regions that can be expressed using the relations $On(p, c)$ and/or $Inside(p, r)$ where $p$ is a point, $c$ is a curve, $r$ is a region, and any relation/action involving only points in first-order logic can be included in our vocabulary. The vocabulary grows richer as more spatial relations and actions are specified.

2. Any spatial relation or action that can be included in the vocabulary is solvable by our SPS. Real QCSPs are known to be computationally intractable, so a substantial part of the spatial problem solving literature concentrates on constraint satisfaction problems (CSPs). We developed a general framework for solving spatial problems specified as QCSPs. The framework bypasses the process of quantifier elimination (QE) – the computational bottleneck and a doubly exponential problem – by taking the help of previously solved *similar* spatial problems. We show that, if the symbolic solution to a problem can be expressed

1. The reader should keep the DR architecture in mind. As shown in Figure 1, there are two problem solvers – the main problem solver which will be always referred to as the “problem solver” (this might be a human) and the spatial problem solver which will be referred to as the “SPS” (this strictly has no human intervention). The problem solver is responsible for the entire problem solving strategy including converting domain-specific perceptions and actions into domain-independent spatial problems. The SPS is responsible only for solving the domain-independent spatial problems that it receives from the problem solver. It is important to not get confused between the roles played by the two.

concisely, quantifiers can be eliminated from spatial problems in low-order polynomial time using similar previously solved problems. The framework leaves room to be more efficient and convenient by incorporating future results in at least two possible directions – learning constraints from examples (automatic constraint acquisition) and carefully exploiting a rich portfolio of QE algorithms.

The rest of the paper is organized as follows. In the next section, we discuss the language for specifying a spatial problem to the SPS. Section 3 describes the SPS. Section 4 analyzes the computational complexity of the SPS. Section 5 shows how the proposed SPS can be augmented to a traditional AI problem solver (SOAR) for reasoning with diagrams in two real-world applications. Finally, we end with discussion and conclusion.

## 2. Specification Language

In this section, we discuss a high-level language that is finite, extensible, human-usable, and expressive enough to describe a wide variety of 2D spatial problems relevant to DR. The problems specified in this language will be accepted as input by the SPS and solved without human intervention. The specification language is independent of the SPS, i.e., the problem specification remains unchanged even if the underlying representation and reasoning strategy of the SPS change.

### 2.1 Diagrammatic Objects

The specification language recognizes three kinds of diagrammatic objects – points, curves, regions.

***Point.*** A point is the basic diagrammatic object. The other objects are defined in terms of a set of points.

***Curve.*** A curve is the set of points *on* it. We approximate a curve piecewise-linearly. Thus, if curve $c$ is approximated by the sequence of $n$ points $\{p_1, p_2, ...p_n\}$, then $c$ is the set of points that lies on its constituent line segments, i.e.

$$c \equiv \{p : On(p, \{p_1, p_2\}) \vee On(p, \{p_2, p_3\}) \vee ...On(p, \{p_{n-1}, p_n\})\}$$

where $p \leftarrow (x, y)$, $x, y \in \Re$, and $\{p_i, p_{i+1}\}$ is the line segment from $p_i$ to $p_{i+1}$. We call the points $\{p_1, p_2, ...p_n\}$ *vertex points.* For the sake of simplicity of specification, the problem solver will write the sequence of vertex points $\{p_1, p_2, ...p_n\}$ to specify a curve $c$.

***Region.*** A region is the set of points *inside* its boundary. The boundary of a region is a closed curve which is approximated piecewise-linearly. Thus, a region is a simple (convex or concave) polygon. Any simple polygon can be triangulated such that a point inside the region is inside one of the triangles. Thus, if the boundary of region $r$ is approximated by the sequence of $n$ points $\{p_1, p_2, ...p_n\}$, then

$$r \equiv \{p : Inside(p, \triangle(r)[1]) \vee Inside(p, \triangle(r)[2]) \vee ...Inside(p, \triangle(r)[m])\}$$

where $m$ is the number of triangles in region $r$ after triangulation, $\triangle(r)[i]$ is the $i^{th}$ triangle of $r$, and $p \leftarrow (x, y)$, $x, y \in \Re$. For the sake of simplicity of specification, the problem solver will write the sequence of vertex points $\{p_1, p_2, ...p_n\}$ of the boundary curve to specify a region $r$. Whether a sequence of vertex points corresponds to a curve or a region will be determined automatically by the system from the context of the property/relation predicate. More on how we define $On$ and $Inside$ in section 3.1.

Further, the SPS can be asked to recognize the kind of diagrammatic object(s) obtained as the solution to a spatial problem. This is achieved by the function $Recognize(\mathcal{D}_{ext})$ where $\mathcal{D}_{ext}$ is an external diagram (i.e., constituted of pixels unlike an abstract diagram). For example, the set of all points behind a curve $c$ with respect to a given point $p$ can be a region object or a curve object depending on the nature of $c$ and its location with respect to $p$. In order to recognize the output, the SPS colors the corresponding set of pixels on an external diagram where each pixel at some predetermined resolution corresponds to a point. The set of colored pixels are grouped such that two adjacent pixels always belong to the same group. Each group of pixels constitutes a diagrammatic object. The boundary pixels of each group is determined. If a group consists of less than three pixels, we consider it as a point object. If a group consists of more than two pixels and its width (both horizontal and vertical) is always less than three pixels, we consider it as a curve object. Otherwise, the group constitutes a region object.

### 2.2 Vocabulary

Unlike certain well-known qualitative spatial reasoning calculi (e.g., intersection calculus in Egenhofer, 1991, cardinal direction calculus in Frank, 1991, region connection calculus in Randell, Cui, & Cohn, 1992), we are not interested in finding a minimal set of spatial relations nor is our vocabulary based on a closed set of predicates. Rather, our vocabulary is based on a closed set of operators (to be discussed shortly in section 2.3). The spatial relations and actions that can be included in our vocabulary are as follows:

1. Any spatial relation or action involving only points that can be expressed using the fixed set of operators and real variables in first-order logic.

2. Any spatial relation or action involving points, curves or regions that can be expressed in first-order logic using the fixed set of operators, real variables, any relation/action from #1, and the relations $On(p, c)$ and/or $Inside(p, r)$ where $p$ is a point, $c$ is a curve, $r$ is a region.

3. Any spatial relation or action involving points, curves or regions that can be expressed in first-order logic using the fixed set of operators, real variables, and any relation/action from #1 and #2.

Thus, our vocabulary is open-ended and addition of new properties and relations is encouraged when a problem cannot be easily expressed using the existing ones. The observation is that, a human often encounters new perceptions/actions but most of them can be specified using the already known ones. However, having a large vocabulary helps specify new ones more conveniently. From the DR literature (Pisan, 1995; Tessler et al., 1995; Lindsay, 1998; Jamnik, 2001; Ferguson & Forbus, 2000; Chandrasekaran et al., 2004; Banerjee & Chandrasekaran, 2004), we have identified a vocabulary of properties, relations and actions based on their wide usage for expressing a variety of real-world spatial problems in different domains. The same vocabulary will be used in this paper as a starting point for specifying spatial problems. In what follows are a few examples of properties, relations and actions in our vocabulary.

***Properties.*** Associated with each kind of object are a few properties – location of a point; location, closedness and length of a curve; and location, area and periphery of a region, where the periphery of a region refers to its boundary curve. The user can also define particular shapes (e.g., circle, triangle, annulus, etc.) for curves and regions as appropriate for reasoning in his domain. Different shapes might have their own specific properties, such as, radius of a circle, height of a triangle, etc. which can be easily associated with the objects in our vocabulary by the user. DR also requires solving spatial problems concerning a discrete set of points. For such problems, properties, such as, $Centroid(S)$ and $Variance(S)$, where $S$ is a set of points, are included in the vocabulary.

***Relations.*** The vocabulary also contains a few widely used relations (or relational predicates) involving points, such as, $Leftof(p_1, p_2)$, $Topof(p_1, p_2)$, $Collinear(p_1, p_2, p_3)$, $Between(p_1, p_2, p_3)$ where $p_1$, $p_2$, $p_3$ are points. Any other relation involving points can be included in the vocabulary as needed. $On(p, c)$, where $p$ is a point and $c$ a curve, is the fundamental relation involving a curve while $Inside(p, r)$, where $p$ is a point and $r$ a region, is the fundamental relation involving a region in our vocabulary as any other relation involving curves or regions uses $On$ and/or $Inside$. Some of the relational predicates involving curves or regions in our vocabulary are $Intersect(c_1, c_2)$, $IntersectionPoints(q, c_1, c_2)$, $Touches(c_1, c_2)$, $Subcurveof(c_1, c_2)$ where $c_1$, $c_2$ are curves, and $Subregionof(r_1, r_2)$ where $r_1$, $r_2$ are regions.

***Actions.*** Further, there is a set of predicates for identifying emergent objects or modifications of existing objects. For example, $Translate(q, \mathcal{O}, t_x, t_y)$ returns a translation of object $\mathcal{O}$ for $t_x$ units along $x$-axis and $t_y$ units along $y$-axis, $Rotate(q, \mathcal{O}, c, \theta)$ returns a rotation of the object $\mathcal{O}$ with respect to point $c$ as center for $\theta$ degrees in the anti-clockwise direction, $Reflect(q, \mathcal{O}, \{a, b\})$ returns a reflection of object $\mathcal{O}$ with respect to the line segment $\{a, b\}$ (i.e., from point $a$ to point $b$), $Scale(q, \mathcal{O}, c, s_x, s_y)$ returns a scaling of the object $\mathcal{O}$ with respect to point $c$ for $s_x$ units along $x$-axis and $s_y$ units along $y$-axis. When $\mathcal{O}$ is a curve or region, each of these predicates is defined using the corresponding action involving a point and the predicates $On$ and/or $Inside$.

### 2.3 The Language

This is the language in which the problem solver (human or artificial) specifies a spatial problem to the SPS. The internal representations of objects, properties, relations, and the problem-solving strategies are hidden from the problem solver. The specification language remains unchanged even if the underlying representation or problem-solving strategy is changed. We use first-order predicate logic as the specification language, previously reported by Banerjee and Chandrasekaran (2007).

***Operators.*** The language recognizes a set of boolean operators $\{\wedge, \vee, \neg\}$, a set of arithmetic operators $\{+, -, \times, \div\}$, a set of relational operators $\{<, >, =, \neq\}$, and the quantifiers $\{\exists, \forall\}$. The brackets () are used to express precedence while the brackets {} are used to express a set. In this paper, we will often use certain combination of operators, such as, $\leq$, $\geq$, $\Rightarrow$, etc. for the sake of brevity.

***Domain.*** The language allows the problem solver to specify the domain as a set from which the variables can assume values. Unless otherwise stated, the domain is the real plane $\Re^2$ for a point variable and the real line $\Re$ for a non-diagrammatic variable.

***Functions.*** Further, the language recognizes two functions – $Maximize(f, \{x, y, ...\}, \mathcal{C})$ and $Minimize(f, \{x, y, ...\}, \mathcal{C})$, which maximizes and minimizes the function $f$ with respect to the variables $\{x, y, ...\}$ satisfying the boolean combination of constraints $\mathcal{C}$ (which might involve quantifiers) and returns the maximum and minimum value of $f$ respectively along with the conditions on the variables.

***Quantified Constraint Satisfaction Problem.*** An instance of a constraint satisfaction problem (CSP) consists of a tuple $< \mathcal{V}, D, \mathcal{C} >$ where $\mathcal{V}$ is a finite set of variables, $D$ is a domain, and $\mathcal{C}= \{\mathcal{C}_1, ...\mathcal{C}_k\}$ is a set of constraints. A constraint $\mathcal{C}_i$ consists of a pair $< \mathcal{S}_i, \mathcal{R}_i >$ where $\mathcal{S}_i$ is a list of $m_i$ variables and $\mathcal{R}_i$ is a $m_i$-ary relation over the domain $D$. The question is to decide whether or not there is an assignment mapping each variable to a domain element such that all the constraints are satisfied. All of the variables in a CSP can be thought of as being implicitly existentially quantified.

A useful generalization of the CSP is the quantified constraint satisfaction problem, where variables may be both existentially and universally quantified. An instance of the QCSP consists of a quantified formula in first-order logic, which consists of an ordered list of variables with associated quantifiers along with a set of constraints. A QCSP can be expressed as follows:

$$\phi(v_1, ...v_m) \equiv Q(x_n, ...x_1)\phi'(v_1, ...v_m, x_1, ...x_n)$$
$$Q(x_n, ...x_1) \equiv Q_n x_n, ...Q_1 x_1$$

where $Q_i \in \{\forall, \exists\}$, $\{x_1, ...x_n\}$ is the set of quantified variables, $\{v_1, ...v_m\}$ is the set of free variables, $\mathcal{V}= \{v_1, ...v_m, x_1, ...x_n\}$, and $\phi'$ is a quantifier-free expression called the matrix. Such representation of a quantified expression $\phi$, where it is written as a sequence of quantifiers followed by the matrix, is referred to as prenex form. Example of a QCSP is as follows:

$$Subcurveof(c_1, c) \equiv \forall p, On(p, c_1) \Rightarrow On(p, c)$$

where $c_1$, $c$ are curves in $\Re^2$. In this example, there are two constraints:

$$< \{p, c_1\}, On >$$
$$< \{p, c\}, On >$$

Further, $\mathcal{V}= \{p\}$ and $D = \Re^2$. The variables $c$, $c_1$ are given. The question here is to decide whether there is an assignment mapping $p$ to an element in $\Re^2$ such that the logical combination of constraints is not satisfied. If such an assignment exists, then $c_1$ is not a subcurve of $c$; otherwise it is.

***Decision, Function and Optimization problems.*** In the proposed specification language, a spatial problem $\phi$ is expressed as a QCSP where $\mathcal{V}$ consists of variables of type point, curve or region and $D = \Re^2$. Solving a spatial problem involves:

1. When there are no free variables in $\mathcal{V}$ (i.e., all variables in $\mathcal{V}$ are quantified), deciding whether or not there exists a mapping from $\mathcal{V}$ to $D$ satisfying $\mathcal{C}$.

2. When there are free variables in $\mathcal{V}$, computing the conditions on the free variables such that a mapping from $\mathcal{V}$ to $D$ satisfying $\mathcal{C}$ exists.

Thus, a spatial problem can be classified as a *decision* or a *function* or an *optimization* problem in the real domain. The first case constitutes a decision problem and yields a *True* or *False* solution. The second case constitutes a function problem which involves computing the diagrammatic object(s) described by the conditions on the free variables. If a spatial problem requires computing the "best" mapping from $\mathcal{V}$ to $D$ satisfying $\mathcal{C}$, it is called an optimization problem.

Let us consider an example. Given a curve $c$ and two points $p$, $q$, the spatial problem $BehindCurve(q, c, p)$ is defined as deciding whether or not $q$ is behind $c$ with respect to $p$. This might be specified as deciding whether or not the curve $c$ and line segment $\{p, q\}$ intersect. Thus,

$$BehindCurve(q, c, p) \equiv Intersect(c, \{p, q\})$$

For particular instances of $q$, $p$, $c$, the solution to this problem is $True$ or $False$, hence it is a decision problem (see Figure 8). For particular instances of $p$, $c$, and generalized coordinates of $q$ i.e., $q \leftarrow (x, y)$, the solution to the same problem is a logical combination of conditions involving $x$ and $y$, which when plotted constitutes a region object (see Figure 9). Hence, it is a function problem. While a decision problem merely requires checking whether or not a given instance of an object satisfies the constraints or not, a function problem requires computing all conditions for a general object to satisfy the constraints.

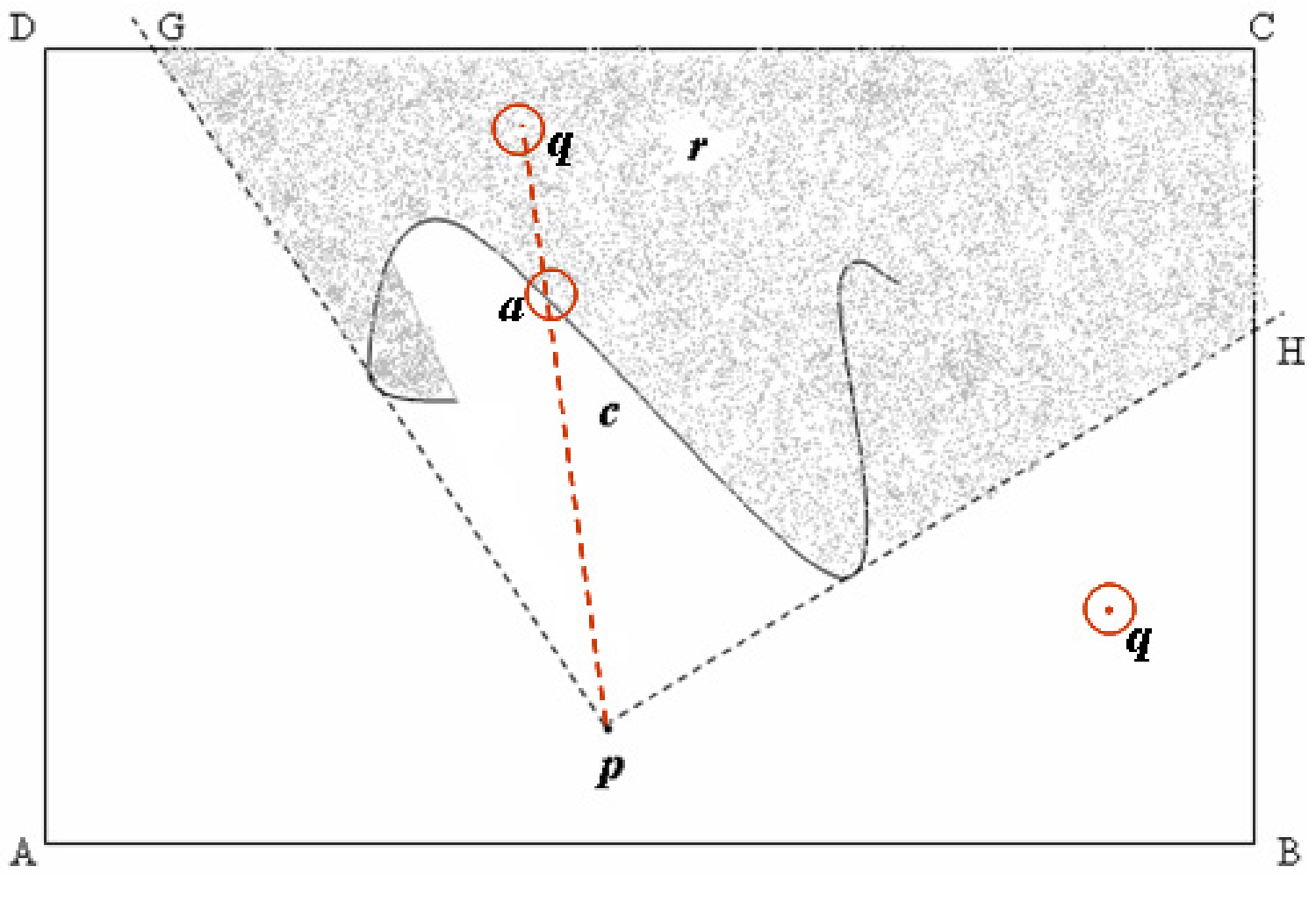

Figure 8: The *BehindCurve* as a decision problem. One of the points $q$ is behind $c$ with respect to $p$ while the other one is not.

Again, given a curve $c$ and two points $p$, $q$, the spatial problem $FurthestBehindCurve(q, c, p)$ is defined as deciding whether or not $q$ is the furthest point behind $c$ with respect to $p$. This might be specified as deciding whether or not $q$ lies behind $c$ with respect to $p$ and

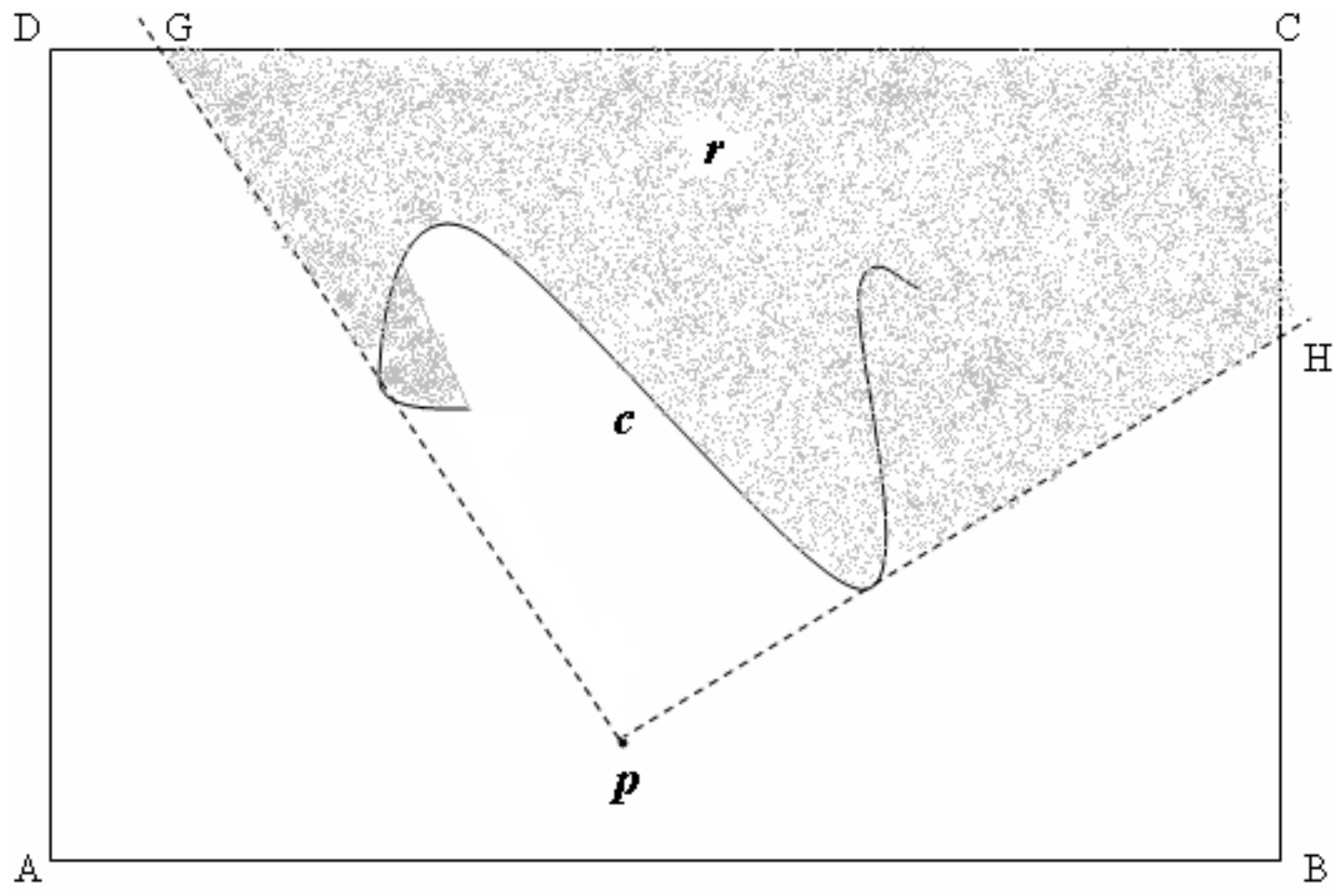

Figure 9: The *BehindCurve* as a function problem. The shaded region $r$ is behind $c$ with respect to $p$.

distance between $p$ and $q$ is maximum. Thus,

$FurthestBehindCurve(q, c, p) \equiv BehindCurve(q, c, p) \wedge \forall b, BehindCurve(b, c, p) \Rightarrow CompareDistance(b, p, q, p)$

$CompareDistance(a, b, c, d) \equiv Distance(a, b) \leq Distance(c, d)$

For particular instances of $q$, $p$, $c$, the solution to this problem is $True$ or $False$, hence it is a decision problem. For particular instances of $p$, $c$, and generalized coordinates of $q$ i.e., $q \leftarrow (x, y)$, the solution to the same problem is a logical combination of conditions involving $x$ and $y$, which when plotted constitutes a single point object, assuming there is only one furthest point behind $c$ with respect to $p$, which is dependent on the nature of $c$ and how the *Distance* function is defined (see Figure 10).

An alternative way of specifying the same problem $FurthestBehindCurve(q, c, p)$ is by explicitly asking to maximize the distance between $p$ and $q$ where $q$ satisfies the constraint $BehindCurve(q, c, p)$, written as:

$FurthestBehindCurve(q, c, p) \equiv Maximize(Distance(q, p), \{q\}, BehindCurve(q, c, p))$

This outputs the conditions involving $x$ and $y$, which constitutes a single point object. The $Maximize$ (or $Minimize$) function assumes the pool of candidates from which to choose the best are those that satisfy the set of constraints. This fact has to be stated explicitly if not using the $Maximize$ (or $Minimize$) function which makes the specification more difficult to come up with and also cumbersome. On the flip side, the specification of a problem using the $Maximize$ (or $Minimize$) function cannot be used as a decision problem. That is, whether or not a particular instance of an object is the best candidate that satisfies the constraints cannot be computed from this specification, unlike the former specification. A problem of this type, which computes the best candidate out of a pool of candidates, is called an optimization problem.

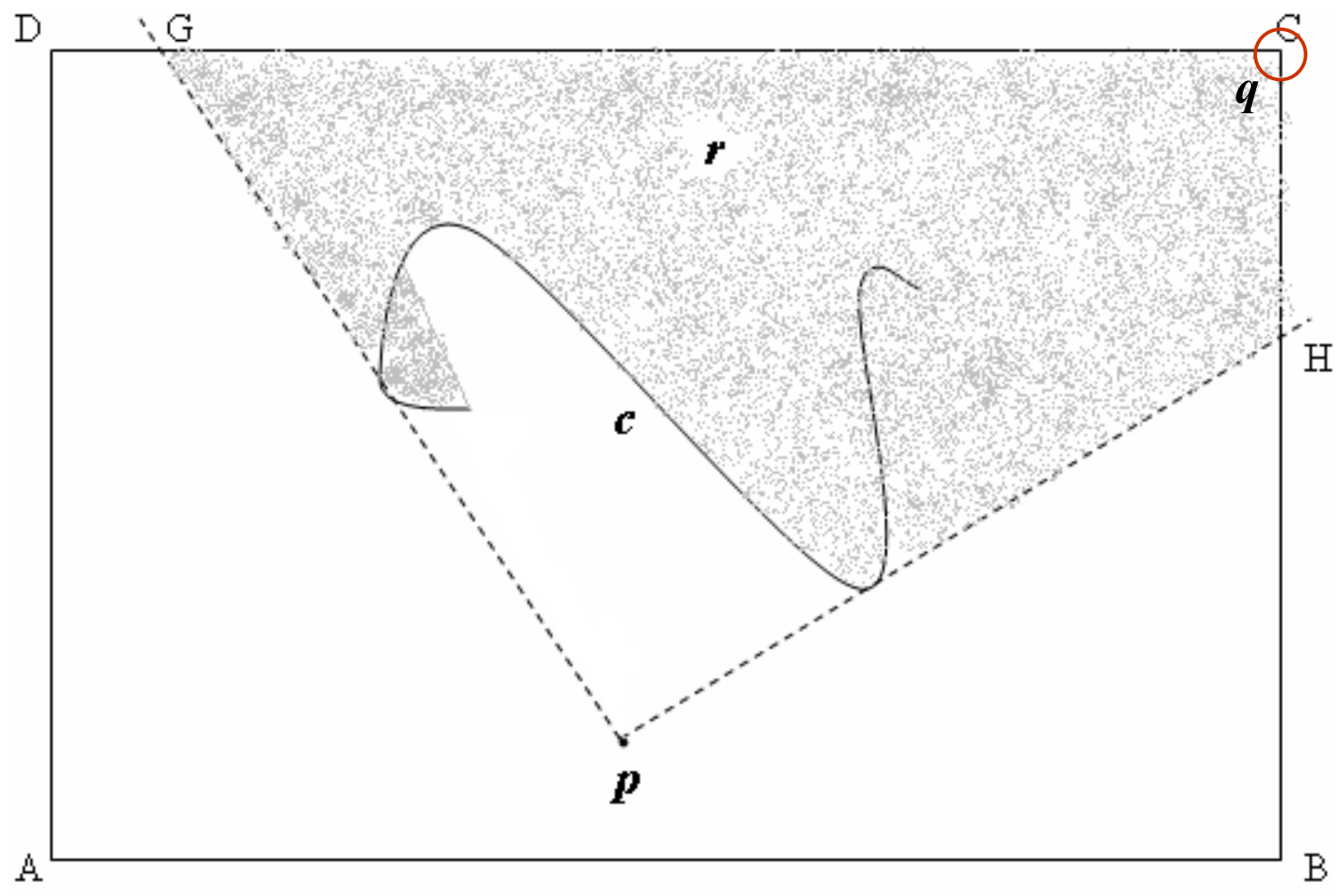

Figure 10: The $FurthestBehindCurve$ as an optimization problem. The point $q$ is the furthest point behind $c$ with respect to $p$.

**Definition 6.** ***Spatial Problem.*** *A spatial problem (or problem) is a QCSP where a variable (quantified or free) can only be of type point, and the domain is* $\Re^2$.

Thus, a spatial problem $\phi$ is a mapping from a diagram $\mathcal{D}$ satisfying a logical combination of constraints $\mathcal{C}$ to a set of booleans $\{True, False\}$ or real numbers $\Re$ or diagrammatic objects $\mathcal{D}'$, i.e.,

$$\phi : \mathcal{D} \xrightarrow{\mathcal{C}} \{True, False\} \cup \Re \cup \mathcal{D}'$$

Solving a spatial problem requires eliminating the quantifiers and solving algebraic equations/inequalities to arrive at the most simplified expression. The computational bottleneck in solving a spatial problem is quantifier elimination (QE) which is inherently doubly exponential (Davenport & Heintz, 1988). More recently, Brown and Davenport (2007) have shown that real QE is doubly-exponential even when there is only one free variable and all polynomials in the quantified input are linear. *In this paper, we will concentrate primarily on QE as part of spatial problem solving and hence, our solution will be an equivalent quantifier-free expression but not necessarily the most simplified one.* Theoretically, the best complexity for QE achieved so far is $O(s^{(l+1)\Pi(k_i+1)}d^{(l+1)\Pi k_i})$ where $s$ is the number of polynomials, their maximum degree is $d$ and coefficients are real, $l$ is the number of free variables, $k_i$ is the number of variables in the $i^{th}$ quantifier block while $k = \sum k_i$ is the number of quantified variables (Basu, Pollack, & Roy, 2003). However, this algorithm is too complicated to yet have a practical implementation. The most general and elaborately implemented method for real QE is the cylindrical algebraic decomposition or CAD (Collins

& Hong, 1991), complexity of which is $(sd)^{O(1)^{k-1}}$. Another implemented method, QE by virtual substitution (Weispfenning, 1988), is restricted to formulas in which the quantified variables occur at most quadratically. The complexity of this method is doubly exponential in the number of blocks of variables delimited by alternations of the existential and universal quantifiers. Thus, while there exist general algorithms for QE, for large real-world problems, it soon becomes too time consuming.

## 3. Spatial Problem Solver

In this paper, we concentrate on developing an efficient SPS without sacrificing its generality. The goal of our design of the SPS is to bypass the general QE algorithms as much as possible, either by taking the help of previously solved *similar* problems in memory to obtain the solution or by using a set of more practical algorithms each of which is developed for a limited class of problems. Here we describe the overall control mechanism of the SPS (see Figure 11).

In many domains, such as, military, spatial problems involve diagrammatic objects that are arbitrary shaped (e.g., mountainous regions) and often cannot be approximated enough by well-defined shapes so that the solution can reliably depend on the specifics of the shape. For example, the solution to the problem of finding all places behind a mountain where one can hide from the enemy depends critically on the particular shape of the mountain. Due to such nature of domains, we choose to represent curves piecewise-linearly and regions as polygons. Piecewise-linear curves and polygonal regions are unions of line segments and triangular regions respectively, which facilitate decomposition of complex problems into simpler and *similar* subproblems. We observe that similar subproblems involving both existential and universal quantifiers occur regularly in the spatial problem solving process which are solved by one of the QE algorithms (e.g., CAD), thereby incurring doubly exponential time. We minimize this enormous computational cost by reusing the solutions of subproblems previously solved.

Given a spatial problem $\phi$ in the specification language, the SPS replaces numerical values in the problem by symbolic variables, and then transforms the symbolic problem from specification to a modeling language (to be described shortly) by progressively replacing objects/predicates by *base objects/predicates* in their internal definitions. If a definition cannot be found, it flags an error and halts till provided. As a first step, the SPS decomposes $\phi$ into disjunctions and/or conjunctions of subproblems $\phi_i$ in prenex form. As we will see later, all of these subproblems $\phi_i$ are similar to each other in that if one of them can be solved, solution to any of the others can be computed from it. Next, it searches the memory for problems similar to $\phi_i$. The memory contains symbolic problems and their corresponding quantifier-free symbolic solutions. If $\phi_i$ can be mapped to one of these problems, its solution is readily obtained by reverse-mapping from the corresponding symbolic solution in memory. Obtaining a solution in such a way completely bypasses the QE process, which is the computational bottleneck of the SPS, thereby reducing the computational costs considerably. If SPS cannot map $\phi_i$ to any problem in memory, it sends $\phi_i$ to the problem classifier that classifies and sends it to the appropriate QE algorithm. The problem classifier and combination of QE algorithms have been borrowed from *Mathematica* (Wolfram, 2003). Once the SPS solves a new subproblem, the subproblem and its solution are stored in memory so that the solution can be used when a similar problem is encountered in future. Thus, the SPS grows more efficient as it solves more problems. Finally, the SPS computes the solution to the given problem $\phi$ by combining the solutions of all its subproblems.

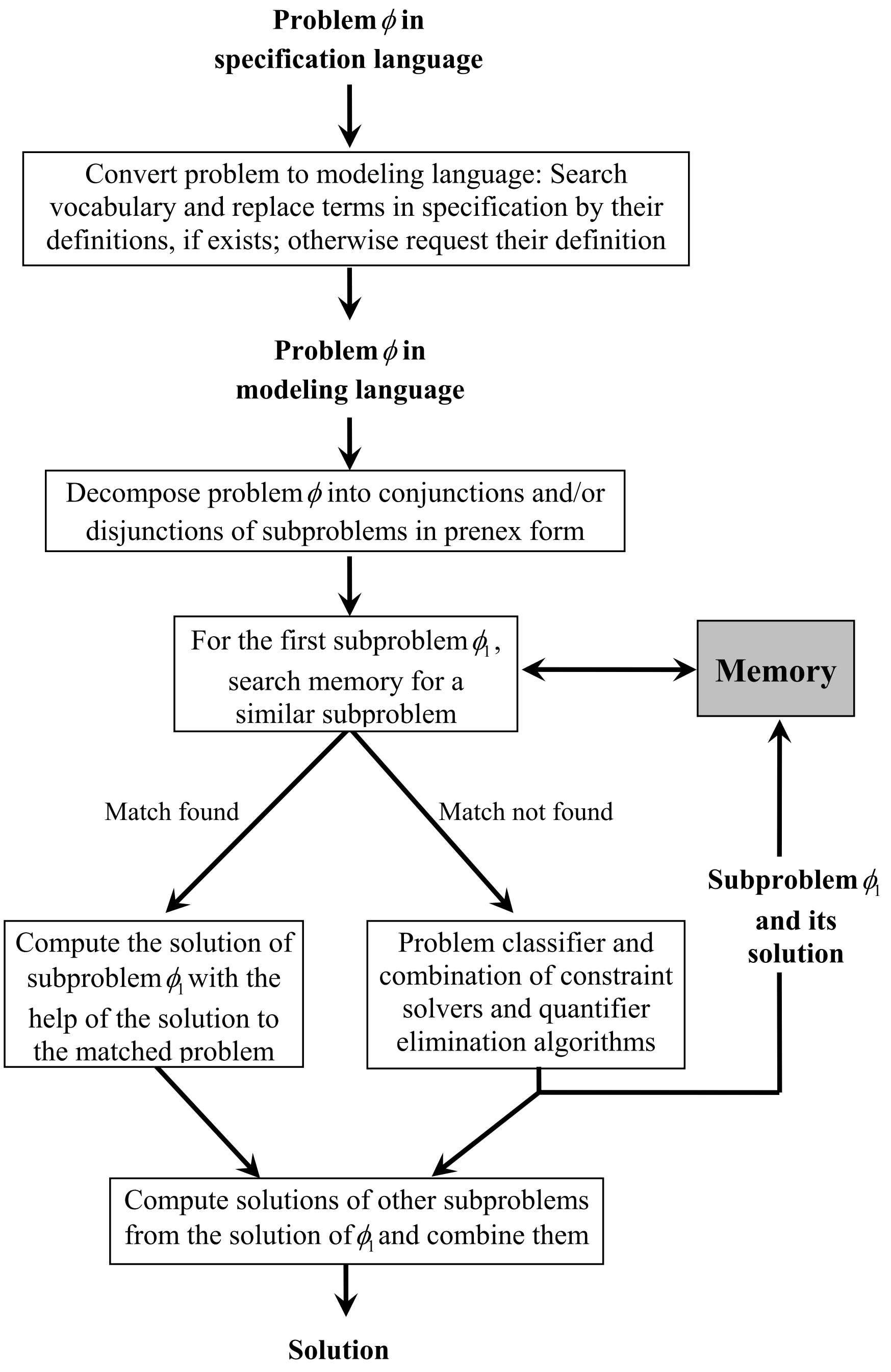

Figure 11: Flow diagram of our spatial problem solver.

Unfortunately, for some problems, quantifiers cannot be eliminated symbolically in reasonable time. The SPS tries for a prescribed time, after which it resorts to more practical methods, such as, techniques especially suited for low degree polynomials (e.g., Dolzmann, Sturm, & Weispfenning, 1998) and approximate methods for obtaining a subset of the solution sufficient for immediate purposes (e.g., Ratschan, 2006; Lasaruk & Sturm, 2006). It has been shown, in integer linear programming (e.g., Leyton-Brown, Nudelman, & Shoham, 2002) and satisfiability testing (e.g., Xu, Hutter, Hoos, & Leyton-Brown, 2008), that the best on-average solver can be out-performed by carefully exploiting a portfolio of possibly poorer on-average solvers, and accordingly, researchers have experimented with different ways of selecting a portfolio of solvers (see for example, Xu et al., 2008; Pulina & Tacchella, 2007; Sayag, Fine, & Mansour, 2006; Streeter, Golovin, & Smith, 2007; Gebruers, Hnich, Bridge, & Freuder, 2005; O'Mahony, Hebrard, Holland, Nugent, & O'Sullivan, 2008). As none of these work involve solving QCSPs over the real domain, they are not directly usable for our purposes and will not be further discussed in this paper. However, we do expect the same result to extend to QCSP solvers over the real domain, and building and smartly selecting from a portfolio of QCSP solvers is a promising line of future research. In our approach, once a subproblem is deemed symbolically unsolvable in the prescribed time, its specification is stored in memory so that in future, a similar problem can be directly subjected to practical methods, thereby saving the prescribed time.

### 3.1 Modeling Language

This is the language in which a problem is described in terms of the underlying representations of objects/properties/relations in a form that can be readily subjected to algebraic manipulation. The location of a point $p$ is represented as a pair $(x, y)$, $x, y \in \Re$, as its coordinates.

**Notation.** The $x$- and $y$-coordinates of a point $p$ are denoted by $p.x$ and $p.y$ respectively.

The distance between two points, $p$ and $q$, is given by

$$Distance(p, q) \equiv \sqrt{(p.x - q.x)^2 + (p.y - q.y)^2}$$

The location of a curve $c$ is represented by the sequence of vertex points $\{p_1, p_2, ...p_n\}$.

**Notation.** The number of vertex points in a curve $c$ is denoted by $\#(c)$, the $i^{th}$ vertex point is denoted by $c[i]$, and the $i^{th}$ line segment is denoted by $\{c[i], c[i+1]\}$.

A line segment $ls$ is specified by its pair of vertex (or terminal) points. The $x$- and $y$-coordinates of $ls$ are represented parametrically as

$$f_x(ls, t) \equiv ls[1].x + t \times (ls[2].x - ls[1].x)$$

$$f_y(ls, t) \equiv ls[1].y + t \times (ls[2].y - ls[1].y)$$

where $t$ is a parameter, $0 \le t \le 1$. The relation $On(p, ls)$, where $p$ is a point, is given by

$$On(p, ls) \equiv \exists t, 0 \le t \le 1 \wedge f_x(ls, t) = p.x \wedge f_y(ls, t) = p.y$$

Length of a line segment $ls$ is given by

$$Length(ls) \equiv Distance(ls[1], ls[2])$$

Length of a curve $c$ is given by

$$Length(c) \equiv \sum_{i \leftarrow 1}^{\#(c)-1} Length(\{c[i], c[i+1]\})$$

The location of a region $r$ is represented by the location of its periphery which is a piecewise linear closed curve. As discussed in section 2.1, internally a region is triangulated (computable in linear time as shown in Chazelle, 1991; Seidel, 1991[2]) with the aim of reducing and simplifying computations (more on this in section 3.2).

**Notation.** On triangulation, the number of triangles in a region $r$ is denoted by $\#_{\triangle}(r)$ while the $i^{th}$ triangle in $r$ is denoted by $\triangle(r)[i]$.

The area of a triangle $\triangle$ is given by

$$Area(\triangle) \equiv \tfrac{1}{2} \sum_{i \leftarrow 1}^{3} \triangle[i].x \times \triangle[i \backslash 3 + 1].y - \triangle[i \backslash 3 + 1].x \times \triangle[i].y$$

Note that the area of a triangle is positive if the sequence of vertex points on its periphery are given in a counter-clockwise direction, otherwise it is negative. Area of a region $r$ is given by

$$Area(r) \equiv \sum_{i \leftarrow 1}^{\#_{\triangle}(r)} Area(\triangle(r)[i])$$

The relation $Inside(p, \triangle)$, where $p$ is a point and $\triangle$ is a triangle, is given by

$$Inside(p, \triangle) \equiv \wedge_{i \leftarrow 1}^{3} Leftof(p, \{\triangle[i], \triangle[i \backslash 3 + 1]\})$$

$$Leftof(p, ls) \equiv Area(\{ls[1], ls[2], p\}) > 0$$

where $ls$ is a line segment.

The action $Translate(q, c, t_x, t_y)$ where $q \leftarrow (x, y)$, $c$ is a curve, and $t_x$, $t_y$ are real numbers, is given by

$$Translate(q, p, t_x, t_y) \equiv q.x = p.x + t_x \wedge q.y = p.y + t_y$$

$$Translate(q, c, t_x, t_y) \equiv \forall a, On(a, c) \Rightarrow Translate(q, a, t_x, t_y)$$

**Definition 7.** ***Base Object.*** *A base object is the simplest form of a diagrammatic object. A point is its own simplest form. A line segment is the simplest form of a curve. A triangular region is the simplest form of a region. Thus, internally, there are three base objects – point, line segment, triangle.*

**Definition 8.** ***Base Predicate.*** *A base predicate is a predicate which accepts only the base objects as arguments.*

2. Vik (2001) discusses an implementation in *Mathematica*.

Examples of base predicates include $Leftof(p_1, p_2)$, $Between(p_1, p_2, p_3)$, $On(p, ls)$, $Inside(p, \triangle)$, where $p$, $p_1$, $p_2$, $p_3$ are points, $ls$ is a line segment, $\triangle$ is a triangular region.

**Lemma 1.** $On(p, c) \equiv \vee_{i \leftarrow 1}^{\#(c)-1} On(p, \{c[i], c[i+1]\})$

*Proof.* The proof follows from our representation of a curve, as described in section 2.1. □

**Lemma 2.** $Inside(p, r) \equiv \vee_{i \leftarrow 1}^{\#_{\triangle}(r)} Inside(p, \triangle(r)[i])$

*Proof.* The proof follows from our representation of a region, as described in section 2.1. □

The relations included in the vocabulary are internally defined in terms of base predicates. For example, the predicate, $Intersect(c_1, c_2)$ where $c_1$, $c_2$ are curves, is defined in terms of base predicates as

$$Intersect(c_1, c_2)$$

$$\equiv \exists a, On(a, c_1) \wedge On(a, c_2)$$

$$\equiv \exists a, \vee_{i \leftarrow 1}^{\#(c_1)-1} On(a, \{c_1[i], c_1[i+1]\}) \wedge \vee_{j \leftarrow 1}^{\#(c_2)-1} On(a, \{c_2[j], c_2[j+1]\})$$

### 3.2 Decomposing a Problem

**Definition 9.** ***Decomposition.*** *Decomposition is the process of replacing the relational predicates, involving free variables of types curve and region, in a spatial problem (quantified expression) by conjunctions/disjunctions of base predicates and taking those conjunctions/disjunctions to the front of the expression. The expression following those conjunctions/disjunctions is a subproblem.*

***Example.*** Decomposition of the problem $Intersect(c_1, c_2) \equiv \exists a, On(a, c_1) \wedge On(a, c_2)$ occurs as follows:

$$Intersect(c_1, c_2) \equiv \exists a, On(a, c_1) \wedge On(a, c_2)$$

$$\equiv \exists a, \vee_{i \leftarrow 1}^{\#(c_1)-1} On(a, \{c_1[i], c_1[i+1]\}) \wedge \vee_{j \leftarrow 1}^{\#(c_2)-1} On(a, \{c_2[j], c_2[j+1]\}) \quad (before\ decomposition)$$

$$\rightarrow \vee_{i \leftarrow 1}^{\#(c_1)-1} \vee_{j \leftarrow 1}^{\#(c_2)-1} \exists a, On(a, \{c_1[i], c_1[i+1]\}) \wedge On(a, \{c_2[j], c_2[j+1]\}) \quad (after\ decomposition)$$

$$\equiv \vee_{i \leftarrow 1}^{\#(c_1)-1} \vee_{j \leftarrow 1}^{\#(c_2)-1} Intersect'(\{c_1[i], c_1[i+1]\}, \{c_2[j], c_2[j+1]\})$$

where $Intersect'(\{c_1[i], c_1[i+1]\}, \{c_2[j], c_2[j+1]\})$ is a subproblem. However, the question arises – is $Intersect(c_1, c_2) \equiv \vee_{i \leftarrow 1}^{\#(c_1)-1} \vee_{j \leftarrow 1}^{\#(c_2)-1} Intersect'(\{c_1[i], c_1[i+1]\}, \{c_2[j], c_2[j+1]\})$? That is, can we replace the "$\rightarrow$" by a "$\equiv$"?

**Theorem 1.** *A problem is equivalent before and after decomposition if and only if it does not contain the following forms:*

*F1:* $\exists p, On(p,c) \wedge \neg Inside(p,r)$

*F2:* $\forall p, \neg On(p,c) \vee Inside(p,r)$ *(complement of F1)*

*F3:* $\exists p, Inside(p,r) \wedge \neg Inside(p,r_1)$

*F4:* $\forall p, \neg Inside(p,r) \vee Inside(p,r_1)$ *(complement of F3)*

*where $c$ is a curve, $r$, $r_1$ are regions, and $c$, $r$, $r_1$ are free variables.*

*Proof.* As discussed in section 2.2, in our framework, $On(p,c)$ and $Inside(p,r)$ are the two fundamental relations using which any other relation involving a curve or region is specified. Also, in our framework, point is the only quantifiable variable, $\{\forall, \exists\}$ are the only quantifiers, and $\{\wedge, \vee, \neg\}$ are the boolean operators sufficient to express any boolean expression. Thus, any spatial problem involving only curves (and points but no regions) is a logical combination of smaller problems of the following form:

$$Qp, Rel(p,c) \quad \text{where } Q \in \{\exists, \forall\},\ Rel \in \{On, \neg On\}$$

Any spatial problem involving only regions (and points but no curves) is a logical combination of smaller problems of the following form:

$$Qp, Rel(p,c) \quad \text{where } Q \in \{\exists, \forall\},\ Rel \in \{Inside, \neg Inside\}$$

Any spatial problem involving both curves and regions (and points) is a logical combination of smaller problems of the following form:

$$Qp, Rel_1(p,c) \odot Rel_2(p,r)$$
where $Q \in \{\exists, \forall\},\ Rel_1 \in \{On, \neg On\},\ Rel_2 \in \{Inside, \neg Inside\},\ \odot \in \{\wedge, \vee\}$

Any spatial problem involving two curves, $c$ and $c_1$, (and points) is a logical combination of smaller problems of the following form:

$$Qp, Rel_1(p,c) \odot Rel_2(p,c_1)$$
where $Q \in \{\exists, \forall\},\ Rel_1 \in \{On, \neg On\},\ Rel_2 \in \{On, \neg On\},\ \odot \in \{\wedge, \vee\}$

Again, any spatial problem involving two regions, $r$ and $r_1$, (and points) is a logical combination of smaller problems of the following form:

$$Qp, Rel_1(p,r) \odot Rel_2(p,r_1)$$
where $Q \in \{\exists, \forall\},\ Rel_1 \in \{Inside, \neg Inside\},\ Rel_2 \in \{Inside, \neg Inside\},\ \odot \in \{\wedge, \vee\}$

We symbolically solved each of the above problems (56 total) in two ways – directly and by decomposing – for $p \leftarrow (x,y)$, $c \leftarrow \{p_1, p_2, ...p_n\}$ $(n \geq 2)$, $p_i \leftarrow (x_{p_i}, y_{p_i})$, $c_1 \leftarrow \{a_1, a_2, ...a_u\}$

$(u \geq 2)$, $a_i \leftarrow (x_{a_i}, y_{a_i})$, $r \leftarrow \{q_1, q_2, ...q_m\}$ $(m \geq 3)$, $q_i \leftarrow (x_{q_i}, y_{q_i})$, and $r_1 \leftarrow \{b_1, b_2, ...b_v\}$ $(v \geq 3)$, $b_i \leftarrow (x_{b_i}, y_{b_i})$. It turned out that the solutions from the two ways were equivalent for all problems, except the four cases stated in the theorem statement. Note that $F2$ is the specification for computing whether a curve $c$ is entirely inside a region $r$ or not. Let $ls_i$ be the $i^{th}$ line segment of $c$ $(1 \leq i \leq n-1)$ and $\triangle_j$ be the $j^{th}$ triangular region of $r$ $(1 \leq j \leq m-2)$. We found

$$\forall p, \neg On(p, c) \vee Inside(p, r)$$

$$\equiv \forall p, \neg(\vee_{i \leftarrow 1}^{n-1} On(p, ls_i)) \vee (\vee_{j \leftarrow 1}^{m-2} Inside(p, \triangle_j))$$

$$\equiv \forall p, \wedge_{i \leftarrow 1}^{n-1} \vee_{j \leftarrow 1}^{m-2} (\neg On(p, ls_i) \vee Inside(p, \triangle_j))$$

$$\not\equiv \wedge_{i \leftarrow 1}^{n-1} \vee_{j \leftarrow 1}^{m-2} (\forall p, \neg On(p, ls_i) \vee Inside(p, \triangle_j))$$

That is because, for $c$ to be entirely inside $r$, it is not necessary for all line segments of $c$ to be inside a triangle of $r$; a line segment of $c$ can span across multiple triangles of $r$ and $c$ can still be inside $r$. Figure 12(a) shows an example where $c$ is inside $r$ but a line segment of $c$ spans across two triangles of $r$. In such a case, the solution to the problem will be $True$ when computed directly but will be $False$ when computed via decomposition. The first case $F1$ in the theorem statement can be explained similarly. The forms $F1$ and $F2$ can be rewritten as follows:

$$F1: \quad \exists p, On(p, c) \wedge \neg Inside(p, r)$$

$$\equiv \exists p, On(p, c) \wedge Inside(p, \overline{r}) \quad \text{where} \quad \overline{r} \equiv \mathcal{B} - r$$

$$F2: \quad \forall p, \neg On(p, c) \vee Inside(p, r)$$

$$\equiv \forall p, \neg On(p, c) \vee \neg Inside(p, \overline{r}) \quad \text{where} \quad \overline{r} \equiv \mathcal{B} - r$$

where $\mathcal{B}$ is the rectangular region (boundary) containing the diagram as discussed in section 1.2. Note that each of the rewritten forms is equivalent before and after decomposition.

Again, $F4$ is the specification for computing whether a region $r$ is entirely inside a region $r_1$ or not. Let $\triangle_{1,i}$ be the $i^{th}$ triangle of $r_1$ $(1 \leq i \leq v-2)$ and $\triangle_j$ be the $j^{th}$ triangular region of $r$ $(1 \leq j \leq m-2)$. We found

$$\forall p, \neg Inside(p, r) \vee Inside(p, r_1)$$

$$\equiv \forall p, \neg(\vee_{j \leftarrow 1}^{m-2} Inside(p, \triangle_j)) \vee (\vee_{i \leftarrow 1}^{v-2} Inside(p, \triangle_{1,i}))$$

$$\equiv \forall p, \wedge_{j \leftarrow 1}^{m-2} \vee_{i \leftarrow 1}^{v-2} (\neg Inside(p, \triangle_j) \vee Inside(p, \triangle_{1,i}))$$

$$\not\equiv \wedge_{j \leftarrow 1}^{m-2} \vee_{i \leftarrow 1}^{v-2} (\forall p, \neg Inside(p, \triangle_j) \vee Inside(p, \triangle_{1,i}))$$

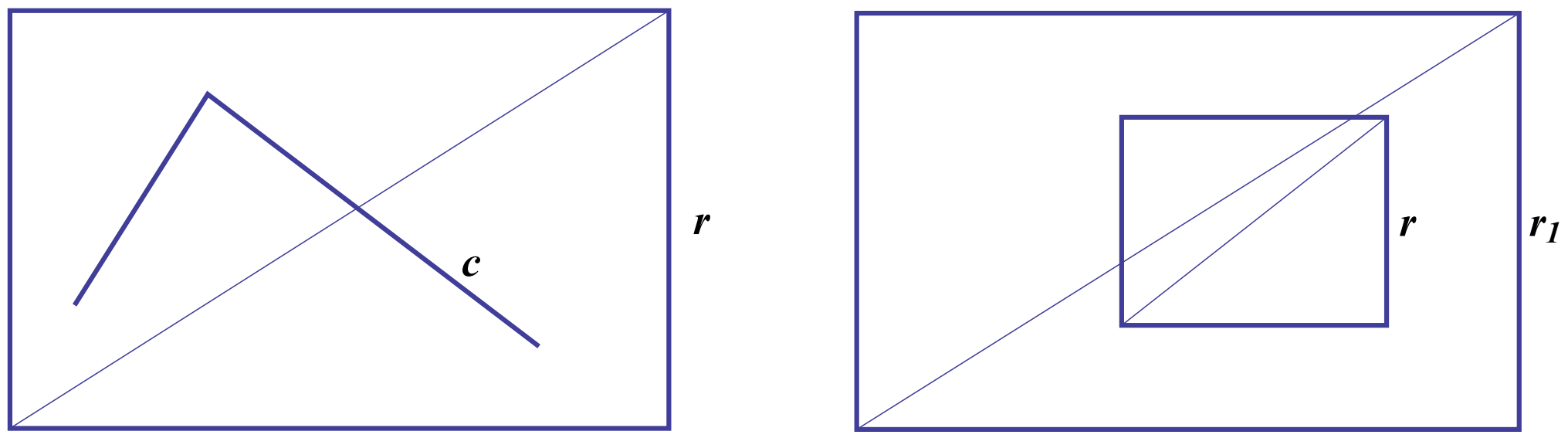

(a) Curve $c$ is inside region $r$ but each line segment of $c$ is not inside one triangle of $r$.

(b) Region $r$ is inside region $r_1$ but each triangle of $r$ is not inside one triangle of $r_1$.

Figure 12: Examples to show the decomposition of curves and regions for problems containing the forms $F1$, $F2$, $F3$, $F4$.

That is because, for $r$ to be entirely inside $r_1$, it is not necessary for all triangles of $r$ to be inside a triangle of $r_1$; a triangle of $r$ can span across multiple triangles of $r_1$ and $r$ can still be inside $r_1$. Figure 12(b) shows an example where $r$ is inside $r_1$ but a triangle of $r$ spans across two triangles of $r_1$. In such a case, the solution to problem will be $True$ when computed directly but will be $False$ when computed via decomposition. The third case $F3$ in the theorem statement can be explained similarly. The forms $F3$ and $F4$ can be rewritten as follows:

$$F3: \quad \exists p, Inside(p,r) \wedge \neg Inside(p,r_1)$$

$$\equiv \exists p, Inside(p,r) \wedge Inside(p,\overline{r_1}) \quad \text{where} \quad \overline{r_1} \equiv \mathcal{B} - r_1$$

$$F4: \quad \forall p, \neg Inside(p,r) \vee Inside(p,r_1)$$

$$\equiv \forall p, \neg Inside(p,r) \vee \neg Inside(p,\overline{r_1}) \quad \text{where} \quad \overline{r_1} \equiv \mathcal{B} - r_1$$

Again, each of the rewritten forms is equivalent before and after decomposition. □

**Theorem 2.** *Any subproblem resulting from decomposing a problem contains base predicates only.*

*Proof.* A problem $\phi$ is decomposable due to the presence of relational predicates, involving free variables of types curve and region, in its specification. As stated in section 2.2, a problem involving a curve or region can be specified in our framework using a relation(s) involving only points and the relation $On$ or $Inside$. Thus, a relation $Rel(q,c)$ involving a point $q$ and a curve $c$ can be rewritten as:

$Rel(q,c) \equiv \exists a, On(a,c) \wedge Rel'(q,a)$
or

$Rel(q,c) \equiv \forall a, On(a,c) \Rightarrow Rel'(q,a)$

where $Rel'$ is a base predicate involving the points $q$ and $a$. In both cases, the expression on the right-hand side of '$\equiv$' contains base predicates only. Let $\phi$ be a problem involving points and curves but no regions. Let us replace in $\phi$ each occurrence of non-base predicates involving a curve, such as, $Rel(q,c)$, by their equivalent expression consisting of $On$ and base predicates involving points and line segments only. Then the resulting expression for $\phi$ consists of base predicates involving points only and $On$. By lemma 1, a non-base $On$ can be rewritten as disjunctions of base $On$. Therefore, the resulting expression for $\phi$ consists of base predicates involving points only.

Similarly, a relation $Rel(q,r)$ involving a point $q$ and a region $r$ can be rewritten as:

$Rel(q,r) \equiv \exists a, Inside(a,r) \wedge Rel'(q,a)$
or
$Rel(q,r) \equiv \forall a, Inside(a,r) \Rightarrow Rel'(q,a)$

where $Rel'$ is a base predicate involving the points $q$ and $a$. If $\phi$ is a problem involving points and regions but no curves, replacing each occurrence of non-base predicates involving a region, such as, $Rel(q,r)$, by their equivalent expression consisting of $Inside$ and base predicates involving points only, and then using lemma 2, results in an expression for $\phi$ consisting of base predicates involving points and triangular regions only. Both the above processes can be employed when $\phi$ involves both curves and regions. Thus, any subproblem resulting from decomposing a problem will contain base predicates only. □

### 3.3 Mapping to a Similar Problem

**Definition 10.** ***Similarity.*** *We define two spatial problems (quantified expressions) to be similar if there exists a one-to-one correspondence between their variables (free and quantified).*

Given two similar problems, $\phi_1$ and $\phi_2$, and the solution $\varsigma_1$ of $\phi_1$, the goal is to construct a one-to-one mapping $\Gamma$ between the variables of $\phi_1$ and $\phi_2$ such that the solution of $\phi_2$ can be obtained by replacing the variables in $\varsigma_1$ by the corresponding variables, thereby completely bypassing the QE process – the computational bottleneck of SPS. The one-to-one mapping exists if $\phi_1$ and $\phi_2$ are logically equivalent. However, equivalence checking for logical expressions is NP-hard (Dershowitz & Jouannaud, 1990; Goldberg & Novikov, 2003). Thus, equivalence checking cannot be used to determine similarity efficiently.

***Problem features.*** Let $\phi'$ be the quantifier free expression when $\phi$ is expressed in prenex form, i.e.,

$$\phi(v_1, ...v_m) \equiv Q(x_n, ...x_1)\phi'(v_1, ...v_m, x_1, ...x_n)$$

where no variable $x_i$ appears more than once in $Q$ and $Q$ contains no redundant variables. A quantifier block $qb$ of $Q$ is a maximal contiguous subsequence of $Q$ where every variable in $qb$ has the same quantifier type. The quantifier blocks are ordered by their sequence of appearance in $Q$; $qb_1 \leq qb_2$ iff $qb_1$ is equal to or appears before $qb_2$ in $Q$. Each quantified variable $x_i$ in $\phi'$ appears in some quantifier block $qb(x_i)$, and the ordering of the quantifier blocks imposes a partial order on the quantified variables. The variables in the same quantifier block are unordered.

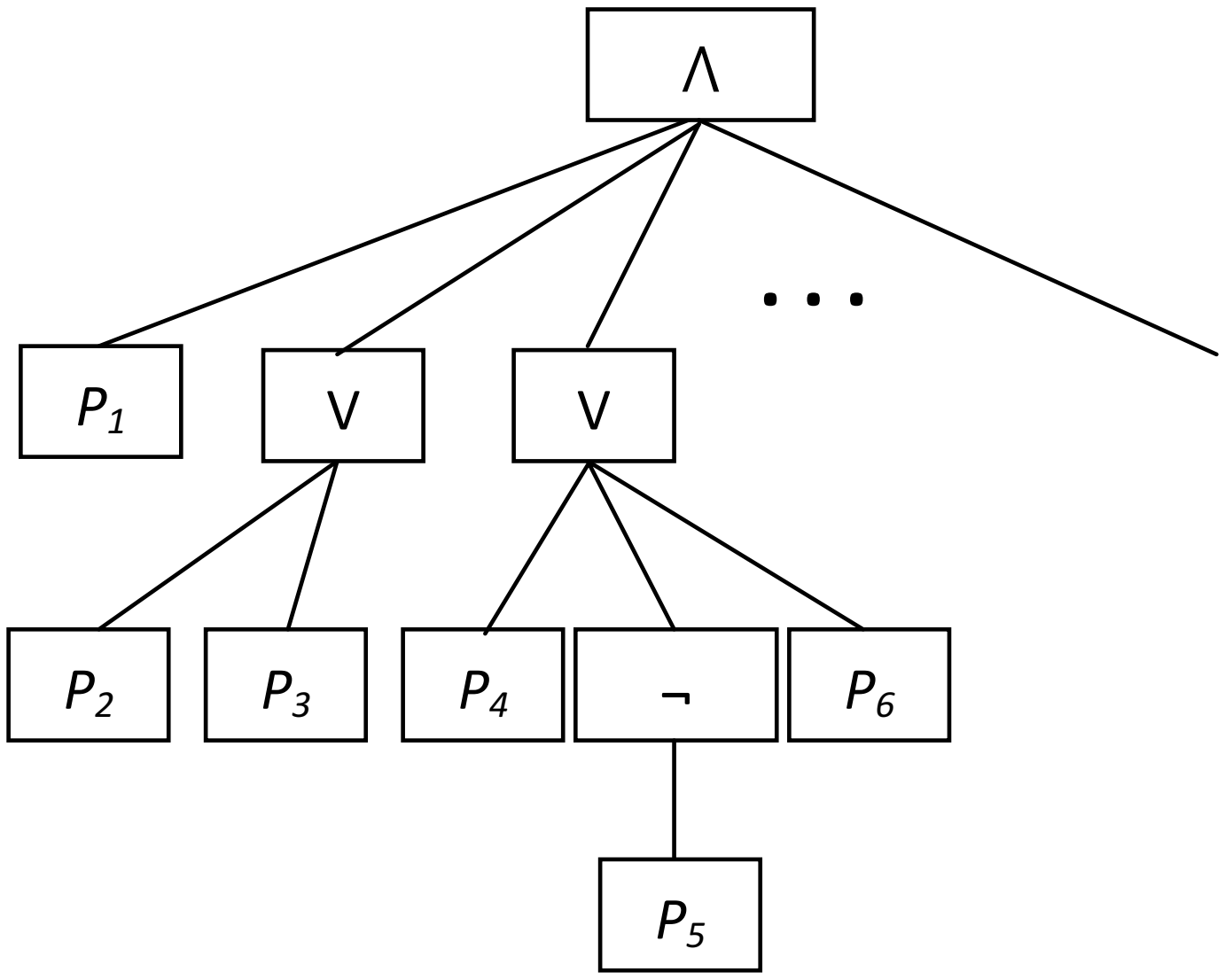

Figure 13: Parse tree for the matrix of a problem in conjunctive normal form.

Let $\phi_1 \equiv Q_1\phi'_1$ and $\phi_2 \equiv Q_2\phi'_2$ while $\tau_1$ and $\tau_2$ be the parse trees for $\phi'_1$ and $\phi'_2$ respectively. For example, the matrix $\phi'$ of a problem $\phi$ in conjunctive normal form might look like:

$$\phi' \equiv P_1 \wedge (P_2 \vee P_3) \wedge (P_4 \vee \neg P_5 \vee P_6) \wedge ...$$

where each $P_i$ is a predicate. If $\phi$ is a subproblem, each $P_i$ is a base predicate. The parse tree of the above sentence is shown in Figure 13.

Two trees, $\tau_1$ and $\tau_2$, are isomorphic if there exists a bijection $\chi : \tau_1 \rightarrow \tau_2$ that preserves adjacency and root vertex, i.e., $\chi(u)$ is adjacent to $\chi(w) \Leftrightarrow u$ is adjacent to $w$, and $\chi(root(\tau_1)) = root(\tau_2)$. It follows that two isomorphic trees have the same maximum height and the same number of vertices at any height. Let $l$ be the maximum height of $\tau$ and $\nu_i$ be the number of vertices at height $i$. The function $\xi$, defined as

$$\xi(< \lambda_1, ...\lambda_\kappa >) = \prod_{i \leftarrow 1}^{\kappa} \rho_i^{\lambda_i}$$

where $\lambda_i$ is an integer, <> denotes a sequence, and $\rho_i$ is the $i^{th}$ smallest prime number, maps a sequence of integers to an unique integer. From a problem $\phi$, a tuple $\vartheta(\phi)$ can be constructed as follows:

$$\begin{aligned}\vartheta(\phi) = \quad & < l, \\ & \xi(\#\text{ vertices at different heights of parse tree}), \\ & \#\text{ quantifier blocks}, \\ & \text{order of quantifier blocks}, \\ & \xi(\#\text{ variables in different quantifier blocks}) >\end{aligned}$$

**Definition 11.** ***Structural Equivalence.*** *Two spatial problems (quantified expressions), $\phi_1$ and $\phi_2$, are structurally equivalent if they satisfy all of the following conditions:*
*1. $\vartheta(\phi_1) = \vartheta(\phi_2)$*
*2. $\tau_1$ and $\tau_2$ are isomorphic to each other, where $\tau_1$ and $\tau_2$ are the parse trees of the matrices of $\phi_1$ and $\phi_2$ respectively.*
*3. The contents (predicate or boolean operator $\{\wedge, \vee, \neg\}$) of each pair of corresponding nodes of $\tau_1$ and $\tau_2$ are identical.*
*4. There exists a one-to-one correspondence between the variables in the arguments of predicates contained in each pair of corresponding nodes of $\tau_1$ and $\tau_2$. Moreover, any two mappings obtained from two pairs of corresponding nodes of $\tau_1$ and $\tau_2$ do not contradict each other.*

As we will see in section 4, structural equivalence of two problems can be computed in time linear in the size of their parse trees. Note that if two problems are structurally equivalent, they will be logically equivalent but not vice versa. For example, the expressions $(P \wedge \neg P) \vee Q$ and $Q$, where $P$ and $Q$ are base predicates, are logically equivalent, but not structurally equivalent since their parse trees are not isomorphic. In general, logical equivalence does not imply structural equivalence when there are redundancies (redundant variables and/or predicates) in one or both problems. For the sake of computational efficiency, we will use structural equivalence to determine the similarity of two problems.

**Theorem 3.** *All subproblems obtained by decomposing a problem are always similar.*

*Proof.* Let us assume, for a contradiction, there exists a problem $\phi$ that decomposes into subproblems such that two of them, $\phi_j$ and $\phi_k$, are dissimilar. Without loss of generality, assume that all subproblems of $\phi$ except $\phi_k$ are similar to $\phi_j$. Then,

$$\phi$$

$$\equiv (\vee_{i_1=1}^{n_1} \wedge_{i_2=1}^{n_2} \ldots \vee_{i_p=1}^{n_p} Q_{i_1 i_2 \ldots i_p} \phi_{i_1 i_2 \ldots i_p}) \odot Q_k \phi_k, \quad \odot \in \{\vee, \wedge\},$$
$$\phi_j \in \{\phi_{i_1 i_2 \ldots i_p} | 0 \leq i_1 \leq n_1, 0 \leq i_2 \leq n_2, \ldots 0 \leq i_p \leq n_p\},$$
$$Q_j \in \{Q_{i_1 i_2 \ldots i_p} | 0 \leq i_1 \leq n_1, 0 \leq i_2 \leq n_2, \ldots 0 \leq i_p \leq n_p\}$$

$$\equiv \vee_{i_1=1}^{n_1} \wedge_{i_2=1}^{n_2} \ldots \vee_{i_p=1}^{n_p} (Q_{i_1 i_2 \ldots i_p} Q_k)(\phi_{i_1 i_2 \ldots i_p} \odot \phi_k)$$

$$\equiv \vee_{i_1=1}^{n_1} \wedge_{i_2=1}^{n_2} \ldots \vee_{i_p=1}^{n_p} Q'_{i_1 i_2 \ldots i_p} \phi'_{i_1 i_2 \ldots i_p}$$

Thus, all subproblems of $\phi$ are similar which contradicts our assumption. Hence the proof follows. □

Intuitively, in the proposed framework, while a curve is represented by an arbitrary number of vertices, a line segment is always represented by its two end points. Similarly, while the periphery of a region is represented by an arbitrary number of vertices, the periphery of a triangular region is always represented by three vertices. Hence, two line segments or triangular regions are always represented similarly and differ only in the coordinates of their constituent vertices, unlike two curves or regions. The base predicates are defined in terms of base objects – points, line segments, and triangular regions. Thus, when a predicate is defined as conjunctions or disjunctions of base predicates, the base predicates are always similar. Decomposition of a problem into subproblems merely replaces one or more of its predicates by similar base predicates. Hence, all the subproblems have to be similar.

### 3.4 Memory Organization

Memory in the SPS is hierarchically organized and stores problems in disjoint classes based progressively on a problem's features in $\vartheta$ (see Figure 14). After decomposing a problem into subproblems and computing their $\vartheta$, if the subproblems have the same value for $\vartheta$, SPS checks whether or not their parse trees are isomorphic and a mapping exists between their variables. Since the memory hierarchy has a constant height, insertion of a problem or searching for a potential class of similar problems can be executed in constant time. Also, the features that classify the problems are discriminative enough to create a large number of classes (leaf nodes), each class containing only a few problems, thereby reducing search to a few problems belonging to a class.

## 4. Computational Complexity

We will now analyze the time complexity of the algorithms used in our framework. In our implementation, a problem $\phi$ is a data structure consisting of two fields – $ParseTree$ and $Solution$. The $ParseTree$ stores the lexicographically sorted parse tree of the matrix of $\phi$ while the $Solution$ stores the symbolic solution to $\phi$ in a concise form. A parse tree can be constructed in time $O(t)$ where $t$ is the number of base predicates and boolean operators $\{\wedge, \vee, \neg\}$ in $\phi$. The boolean operators occupy the non-leaf nodes in the parse tree while the base predicates occupy the leaf nodes. Lexicographically sorting a tree requires lexicographically sorting the contents of the children of each non-leaf node in the tree. Let $t'$ be the number of boolean operators and $t''_i$ be the number of children of the $i^{th}$ boolean operator. Thus, $\sum_{i \leftarrow 1}^{t'} t''_i = t - 1$. Note that since each base predicate is always followed by a boolean operator, $t = \kappa t'$ where $\kappa$ is a constant. Lexicographically sorting a list of the contents of the children of a node requires $O(t''_i log t''_i)$ time. Thus, the total time required for repeating this process for all non-leaf nodes is $\sum_{i \leftarrow 1}^{t'} O(t''_i log t''_i)$. Since the average number of children per node is $\frac{1}{t'} \sum_{i \leftarrow 1}^{t'} t''_i = \frac{t-1}{t'}$, the total time required to lexicographically sort a tree is $\sum_{i \leftarrow 1}^{t'} O(\frac{t-1}{t'} log \frac{t-1}{t'}) = O(t)$.

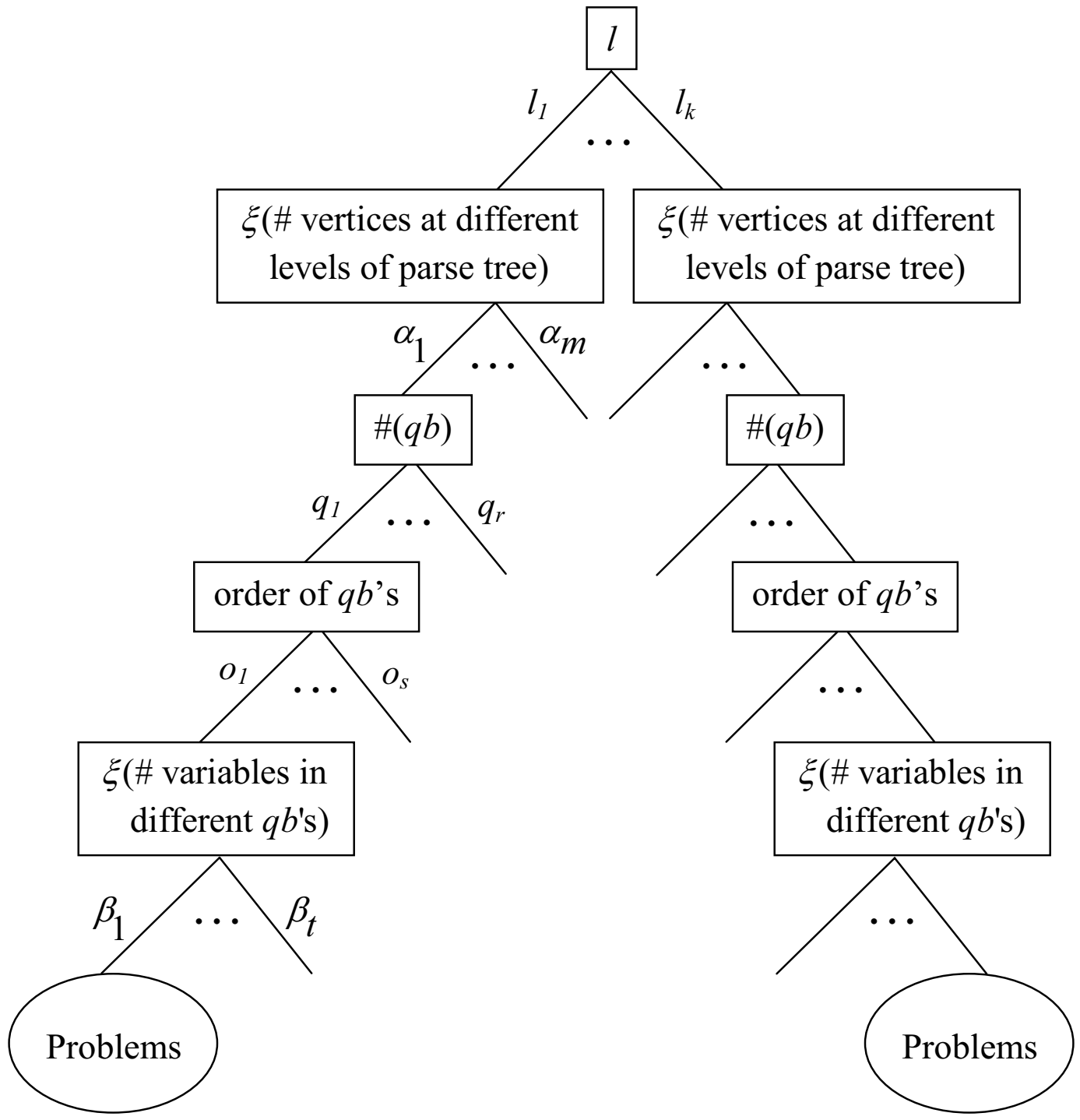

Figure 14: Hierarchical problem classification in memory. At each height in the hierarchy, the branches correspond to the different values of the features captured by $\vartheta$. For example, $l_1$ through $l_k$ correspond to the $k$ different maximum heights of parse trees for the matrices of spatial problems.

Given two problems – $\phi_1$, $\phi_2$, the algorithm $Similar(\phi_1, \phi_2)$ computes whether $\phi_1$ and $\phi_2$ are similar to each other or not (see Figure 15). Since computing $\vartheta$ requires $O(t)$ time, line 1 requires $O(t)$ time. Since checking whether two trees are isomorphic or not requires $O(t)$ time (as shown in Aho, Hopcroft, & Ullman, 1974), line 6 requires $O(t)$ time. Lines 9 through 11 requires $O(t)$ time. Thus, the algorithm runs in $O(t)$ time.

Given an unsolved problem $\phi$ and a similar solved problem $\phi_{similar}$, the algorithm $ComputeSolutionFromSimilarProblem(\phi, \phi_{similar})$ computes the solution to $\phi$ by variable mapping from $\phi_{similar}$ (see Figure 16). $VariableMap$ is a list where each entry is a pair $< v, v_{similar} >$, $v$ being a free variable in $\phi$ and $v_{similar}$ is the corresponding free variable in $\phi_{similar}$. Let the size of $VariableMap$ be $k'$. The lines 5 through 11 requires $O(tk')$ time since the number of nodes in $\phi_{similar}$ is $t$ and the number of arguments in any predicate is small. Lines 12 through 13 requires $O(\omega k')$ time where $\omega$ is the size of the solution to $\phi_{similar}$. Thus, the algorithm runs in $O(tk' + \omega k')$ time.

Finally, given an unsolved problem $\phi$ and a memory $Memory$ that stores problems hierarchically (as described in section 3.4), the algorithm $EliminateQuantifiers(\phi, Memory)$ computes the solution to $\phi$ by variable mapping from a similar problem in $Memory$, if such a problem exists; otherwise solves $\phi$ using a problem classifier and combination of constraint

```
Similar(φ1, φ2)
if ϑ(φ1) ≠ ϑ(φ2),
    return False
else
    τ1 ← φ1.ParseTree
    τ2 ← φ2.ParseTree
    if Isomorphic(τ1, τ2) = False,
        return False
    else
        for each node in τ1
            if the predicate or boolean operator at the corresponding node in τ2 does not match,
                return False
return True
```

Figure 15: Algorithm for deciding whether two problems – $\phi_1$, $\phi_2$ – are similar or not by computing structural equivalence. A problem is a quantified expression.

```
ComputeSolutionFromSimilarProblem(φ, φ_similar)
τ ← φ.ParseTree
τ_similar ← φ_similar.ParseTree
ς_similar ← φ_similar.Solution
ς ← ς_similar
for each node in τ_similar
    if the node contains a predicate (say P),
        for j ← 1 to # arguments in P
            v ← variable occupying j^th argument of P in τ
            v_similar ← variable occupying j^th argument of P in τ_similar
            if VariableMap.Contains(v) = False,
                VariableMap.Add(< v, v_similar >)
for i ← 1 to |VariableMap|
    Replace all occurrences of VariableMap[i, 2] in ς by VariableMap[i, 1]
return ς
```

Figure 16: Algorithm for computing the solution to a problem $\phi$ by mapping variables from a similar problem $\phi_{similar}$. The problem is a quantified expression and the solution is the equivalent quantifier-free expression.

solvers and QE algorithms (as described in section 3). The algorithm is shown in Figure 17.

Let there be $n$ subproblems to a problem. In a problem, some of the predicates are already base predicates while the rest are not which can be written as conjunctions/disjunctions of base predicates thereby leading to decomposition of a problem into subproblems. For example, in section 1.1, the problem $RiskyPortionsofPath(q, c_1, c_2, d)$ is defined in terms of the base predicate $DistanceLessThan(p, q, d)$ (i.e., $Distance(p, q) \leq d$) and the non-base predicates $On(q, c_1)$ and $On(p, c_2)$. Each of these non-base predicates can be written as disjunctions of base predicates, such as, $On(q, \{c_1[i], c_1[i+1]\})$ and $On(q, \{c_2[j], c_2[j+1]\})$, respectively, thereby leading to decomposition of $RiskyPortionsofPath$ into subproblems. Each of the subproblems inherits the base predicates from the problem (e.g., $DistanceLessThan(p, q, d)$) and also includes the new base predicates (e.g., $On(q, \{c_1[i], c_1[i+1]\})$, $On(q, \{c_2[j], c_2[j+1]\})$) obtained from the non-base predicates. Let $\alpha$ be the number of polynomials in the base predicates of a problem and $\beta$ be the number of polynomials due to the newly obtained base predicates in a subproblem. Since all subproblems are similar, each of them will have $\alpha + \beta$ polynomials. The total number of polynomials $s$ in a problem is $O(\alpha + \beta n)$.

Let $d$ be the maximum degree of any polynomial in a subproblem. Since all subproblems are similar, each of them will have maximum degree $d$. The maximum degree of polynomials in a problem will also be $d$ if objects are represented piecewise-linearly, in which case $d \leq 2$. If the objects are not represented piecewise-linearly, the degree will be much larger than two which might lead to a situation where the problem might not be solvable in reasonable time.

Let $k$ be the number of quantified variables in a problem. Then each subproblem also has $k$ quantified variables. Let the computational complexity of using a general QE algorithm for solving a problem be $T(n)$ while that for solving a subproblem is $T(1)$, where $T$ is a doubly exponential function, such as, when using CAD, $T(n) = (sd)^{O(1)^{k-1}}$. Note that $T(n) \gg nT(1)$, i.e., it is more efficient to solve each subproblem using a general QE algorithm than to solve the whole problem using the same algorithm.

In algorithm $EliminateQuantifiers(\phi, Memory)$, lines 4 through 7 require $O(n)$ time. Lines 8 and 9 require $O(t)$ time each. Since line 13 requires $O(t)$ time, lines 11 through 16 require $O(mt)$ time. Line 18 requires time $T(1)$ while lines 20 through 23 require $O((n-1)(tk' + \omega k'))$ time. Thus, the entire algorithm runs in $O(T(1) + mt + (n-1)(tk' + \omega k'))$ time. Note that $\omega$ is the size of the symbolic solution, and if the symbolic solution can be expressed concisely, $\omega$ is small. Since the number of boolean operators is of the order of number of base predicates and each base predicate is defined in terms of at least one polynomial, $t = O(s) = O(\alpha + \beta n)$. Thus, the complexity of the algorithm is $O(T(1) + (m + (n-1)k')s + (n-1)\omega k')$. It can be seen that

$$nT(1) > T(1) + (m + (n-1)k')s + (n-1)\omega k'$$

$$or, \quad ((\alpha + \beta)d)^{O(1)^{k-1}} > (\tfrac{m}{n-1} + k')(\alpha + \beta n) + \omega k'$$

is true provided $\omega$ is not large. That is, it is more efficient to solve a problem by variable mapping than to solve each subproblem using a general QE algorithm provided the size of the stored symbolic solution is not large. For every decomposable problem, the complexity of QE can be reduced as above.

$EliminateQuantifiers(\phi, Memory)$
1. $\phi.Solution \leftarrow \emptyset$
2. Decompose $\phi$ into subproblems $\phi_i \equiv Q_i\phi_i'$, $1 \leq i \leq n$

where $\phi \equiv \vee_{i_1=1}^{n_1} \wedge_{i_2=1}^{n_2} \ldots \vee_{i_p=1}^{n_p} Q_i\phi_i'$, $n = \sum_{k \leftarrow 1}^{p} n_k$

3. $i \leftarrow 1$
4. for $k \leftarrow p$ to 1
5. for $j \leftarrow 1$ to $n_k$
6. $\odot_i \leftarrow$ the $i^{th}$ operator ($\wedge$ or $\vee$) in decomposed $\phi$ from right to left
7. $i \leftarrow i + 1$
8. Construct the parse tree for the matrix $\phi_1'$
9. Compute $\vartheta(\phi)$
10. $flag \leftarrow 0$
11. for $j \leftarrow 1$ to $m$
12. $\phi_{Memory,j} \leftarrow j^{th}$ problem in $Memory$
13. if $Similar(\phi_{Memory,j}, \phi_1) = True$,
14. $\phi_{similar} \leftarrow \phi_{Memory,j}$
15. $flag \leftarrow 1$
16. break loop
17. if $flag = 0$,
18. $\phi_1.Solution \leftarrow ComputeSolutionFromQEAlgorithms(\phi_1)$
19. $\phi_{similar} \leftarrow \phi_1$
20. for $i \leftarrow (2 \times flag + 2)\backslash 3$ to $n$
21. $\phi_i.Solution \leftarrow ComputeSolutionFromSimilarProblem(\phi_i, \phi_{similar})$
22. $\phi.Solution \leftarrow \phi.Solution \odot_i \phi_i.Solution$
23. return $\phi.Solution$

Figure 17: Algorithm for computing the solution to a spatial problem $\phi$ taking the help of previously solved similar problems in $Memory$, thereby bypassing quantifier elimination whenever possible. The problem is a quantified expression and the solution is the equivalent quantifier-free expression.

When a problem is encountered by the SPS for the first time, it is solved by decomposing into subproblems, solving the first subproblem using a general QE algorithm and then obtaining the solution of the rest of the subproblems by mapping their variables to the first subproblem. Since a subproblem and its solution are stored in memory, if a similar subproblem is encountered in future, the SPS bypasses the QE algorithm completely and solves it by variable mapping. In such a case, line 18 of the algorithm is never executed, and the time complexity of solving the problem is

$$(m + nk')s + n\omega k'$$

which is a considerable savings compared to the complexity of solving the entire problem using a general QE algorithm (e.g., complexity of CAD is $(sd)^{O(1)^{k-1}}$), provided $\omega$ is not large. As the SPS solves more problems, the probability to encounter a similar problem in memory increases thereby leading to the above scenario which incurs a complexity of low order polynomial as compared to doubly exponential.[3]

***Example.*** To illustrate the problem solving process, let us consider the spatial problem $BehindCurve(q, c, p)$ (described in section 2.3). For a point $p \leftarrow (p_x, p_y)$ and a curve $c \leftarrow \{p_1, p_2, ...p_n\}$ where $p_i \leftarrow (x_i, y_i)$ is a point, decomposition of the problem occurs as follows:

$$\phi$$

$$\equiv BehindCurve(q, c, p)$$

$$\equiv Intersect(c, \{p, q\})$$

$$\equiv \exists a, On(a, c) \wedge On(a, \{p, q\})$$

$$\equiv \exists a, (\vee_{i=1}^{n-1} On(a, \{p_i, p_{i+1}\})) \wedge On(a, \{p, q\})$$

$$\equiv \vee_{i=1}^{n-1} (\exists a, On(a, \{p_i, p_{i+1}\}) \wedge On(a, \{p, q\}))$$

$$\equiv \vee_{i=1}^{n-1} (Q_i \phi'_i)$$

$$\equiv \vee_{i=1}^{n-1} \phi_i$$

Thus there are $(n-1)$ subproblems $\phi_i$, where

$$\phi'_i \equiv On(a, \{p_i, p_{i+1}\}) \wedge On(a, \{p, q\})$$

$$Q_i \equiv \exists a$$

$$\phi_i \equiv \exists a, On(a, \{p_i, p_{i+1}\}) \wedge On(a, \{p, q\})$$

From Figure 18, $\vartheta(\phi_i) =< 2, 2^1 3^3, 1, < \exists >, 2^1 >$ for $i = 1, 2, ...n-1$. By theorem 3, all $\phi_i$'s are similar since they are the subproblems of the same problem. If the SPS does

3. It should be noted that approximating a continuous curve by a sequence of line segments has its drawbacks. For example, a point $p$ that is *on* a continuous curve $c$ might not be *on* the piecewise-linear approximation of $c$. The SPS can accept a parameter that specifies the maximum length of a line segment to be used in the approximation. As of our current implementation, we leave the onus of determining this maximum length on the problem solver. In this context, it deserves mention that loss of information is inevitable in almost any kind of approximation. For example, when the space in a diagram is approximated by a finite number of pixels, as shown by Banerjee and Chandrasekaran (2010), the diagrammatic objects lose certain spatial information that might be detrimental to spatial problem solving which can be avoided by knowing the minimum allowable resolution (or maximum length of one side of a square pixel).

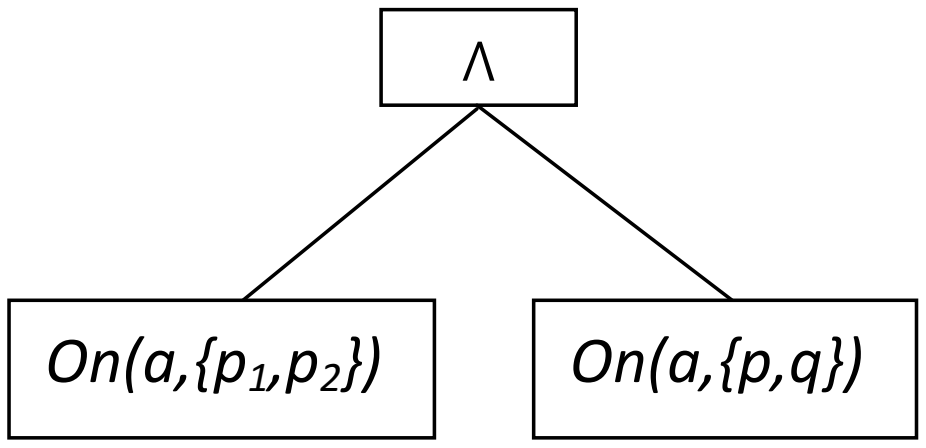

Figure 18: Parse tree for the matrix of the first subproblem of the *BehindCurve* problem.

not find a problem in memory similar to the first subproblem $\phi_1$, it is sent to the problem classifier who sends it to the appropriate QE algorithm. The problem definition, its tuple $\vartheta$, parse tree, and solution are then stored in memory as follows:

$$\phi_1(q, \{p_1, p_2\}, p) \equiv \exists a, On(a, \{p_1, p_2\}) \wedge On(a, \{p, q\})$$

$$\phi_1((x, y), \{(x_1, y_1), (x_2, y_2)\}, (p_x, p_y))$$

$\equiv (p_x - x < 0 \wedge p_x - x_1 \leq 0 \wedge x_1 - x \leq 0 \wedge p_y x_1 - p_y x + p_x y - x_1 y - p_x y_1 + x y_1 = 0) \vee (x - p_x < 0 \wedge x_1 - p_x \leq 0 \wedge x - x_1 \leq 0 \wedge p_y x_1 - p_y x + p_x y - x_1 y - p_x y_1 + x y_1 = 0) \vee ...$

where the arguments of $\phi_1$ are the free variables. The other subproblems are solved by replacing the variables in $\phi_1$ by the mapped variables. If a problem similar to $\phi_1$ is found in memory, $\phi_1$ will also be solved by replacing the mapped variables, just as the other subproblems.

Note that, for example, the *BehindCurve* problem, in the absence of an appropriate vocabulary of properties/relations, would have been specified as (see redlog in Weispfenning, 2001):

$$BehindCurve((x, y), \{(p_{1,x}, p_{1,y}), (p_{2,x}, p_{2,y}), ...(p_{n,x}, p_{n,y})\}, (p_x, p_y))$$

$\equiv \exists a_x, a_y, t, 0 \leq t \leq 1 \wedge p_x + t(x - p_x) = a_x \wedge p_y + t(y - p_y) = a_y \wedge \vee_{i \leftarrow 1}^{n-1} (\exists t_i, 0 \leq t_i \leq 1 \wedge p_{i,x} + t_i(p_{i+1,x} - p_{i,x}) = a_x \wedge p_y + t_i(p_{i+1,y} - p_{i,y}) = a_y)$

Here the total number of quantifiers is $n + 3$, dependent on the number of line segments forming the curve which can be huge for complicated curves as in many real-world applications. In our SPS, due to appropriate decomposition of problems into subproblems, the number of quantifiers in any subproblem is always fixed (4 in this case) irrespective of the spatial complexity of the object(s) (curve in this case). The symbolic solutions of these simple subproblems can be stored for future use which is not possible in systems like redlog. Needless to say, though solving the problem using both the systems produce the same solution, ours is much faster.

## 5. Applications

In this section, we illustrate how the SPS can be deployed in conjunction with a problem solver, human or artificial (such as, soar), for solving spatial problems without human intervention as needed for DR. Two applications will be considered – entity re-identification and ambush analysis – that are deemed very important in the military domain. The subproblems that the SPS autonomously decomposes each spatial problem into will be shown. Problems in military domain involve a wide variety of objects with arbitrary properties and relations, and hence, help to illustrate the expressiveness of the specification language and the efficiency and generality of the SPS.

For the implementation, we used bisoar, due to Kurup and Chandrasekaran (2007), a bimodal version of soar (Laird et al., 1987), where the problem solver uses two kinds of operators – predicate-symbolic operators that are applied on information in predicate-symbolic form and perception-like operators that are applied on a diagram – to bring about state transitions to reach the goal state from an initial state. A human is responsible for providing the broad problem solving strategy for a class of problems; given a specific problem from that class, bisoar uses the predicate-symbolic and perception-like operators accordingly. Since we have used bisoar in a number of different domains (e.g., military, Euclidean geometry, physics, civil engineering; Banerjee & Chandrasekaran, 2007 provide some examples) and still continue to do so, it knows several different problem solving strategies and operators, both predicate-symbolic and perception-like. The emphasis of this section is not on how efficiently bisoar solves problems but on how efficiently the perception-like operators can be executed without incorporating any knowledge that jeopardizes the generality of a general-purpose problem solver. For each spatial problem, we will compare the performance of our proposed SPS with the CAD algorithm in terms of actual computation time which is determined by taking the average of at least 10 runs. As we will see, the SPS excels by a significant margin in most cases.

### 5.1 Entity Re-identification

The entity re-identification problem is a core task in the US Army's All-Source Analysis System (ASAS). ASAS receives a new report about sighting of an entity $T_3$ of type $T$ (e.g. tanks). The task is to decide if the new sighting is the same as any of the entities in its database of earlier sightings, or an entirely new entity. Reasoning has to dynamically integrate information from different sources – database of sightings, mobility of vehicles, sensor reports, terrain and map information – to make the decision. We will follow a novel capability using failure of expectation: If $H$ were true, $O$ should have been observed, but since it was not, $H$ is likely not the case, where $H$ and $O$ are hypotheses and observations respectively (Josephson & Josephson, 1996; Chandrasekaran et al., 2004). In the following, we consider a simple version of the problem to illustrate how the task is solved using DR and the spatial problems involved therein.

Figure 19(a) shows the terrain of interest – mountainous with the closed regions marking impassable areas for entities of type $T$ (e.g., tanks). Let $T_3$ be an entity newly sighted at time $t_3$ located at point $p_3$ while $T_1$, $T_2$ are the two entities that were located at points $p_1$, $p_2$ when last sighted at times $t_1$, $t_2$ respectively. $T_1$ and $T_2$ were retrieved from the database as having the potential to be $T_3$ based on their partial identity information. Also, in the

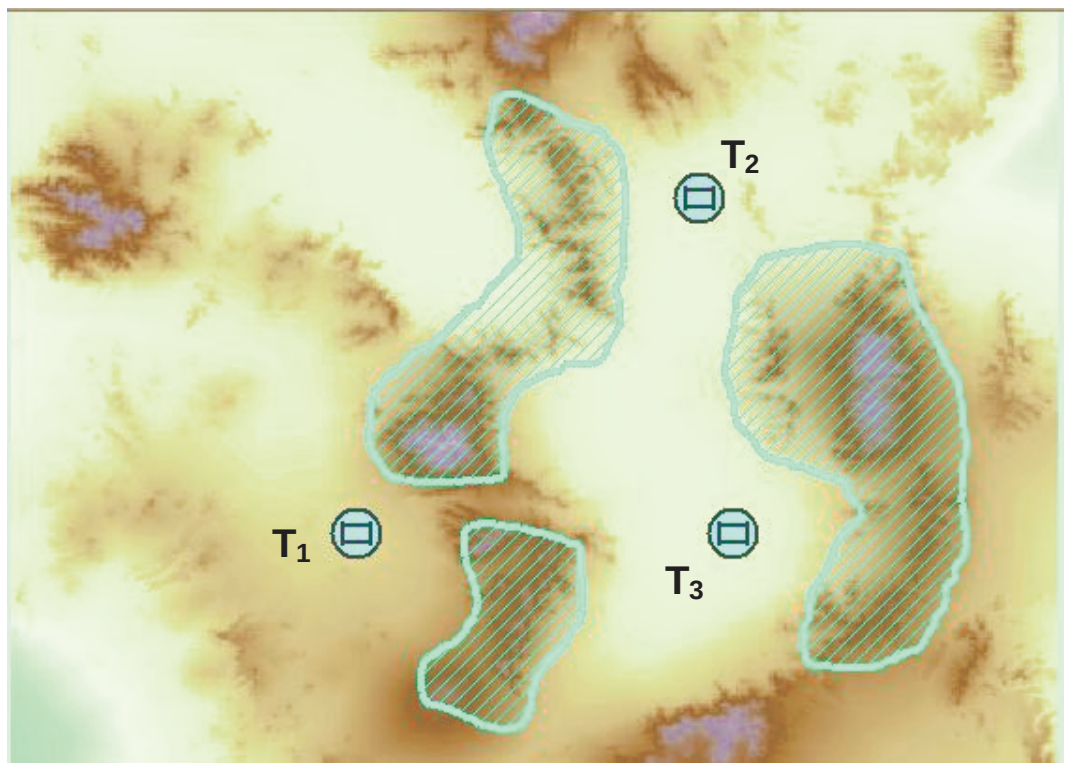

(a) Terrain, impassable regions, and sighted tanks.

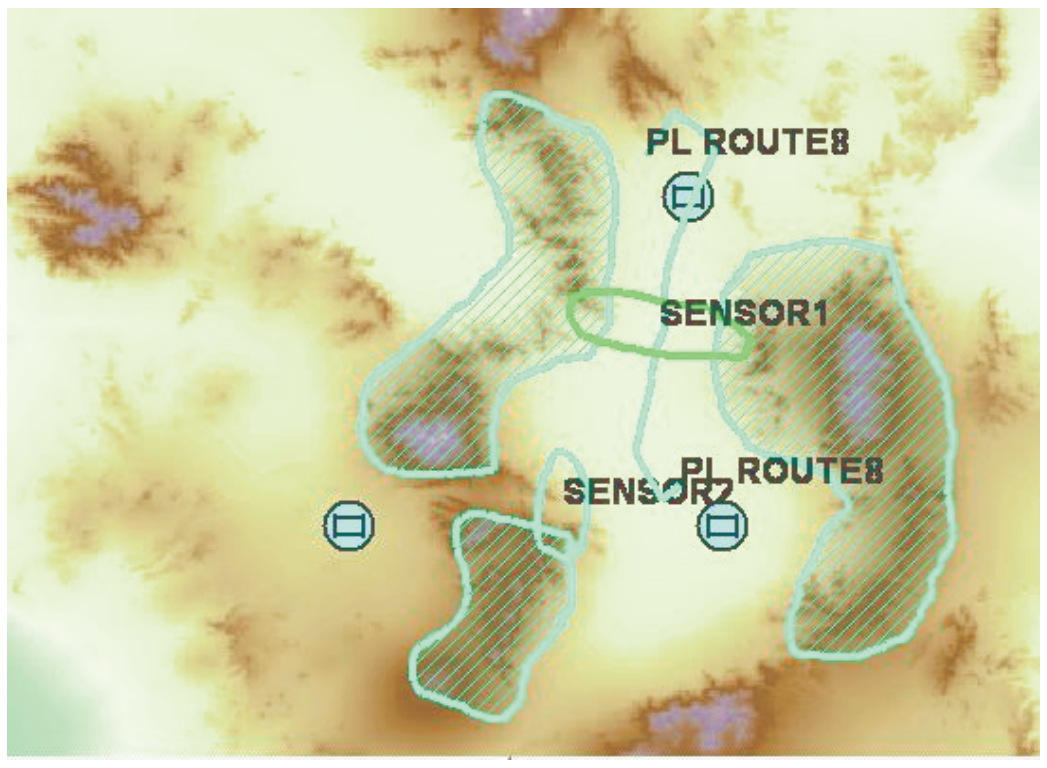

(b) A short path between $T_2$ and $T_3$.

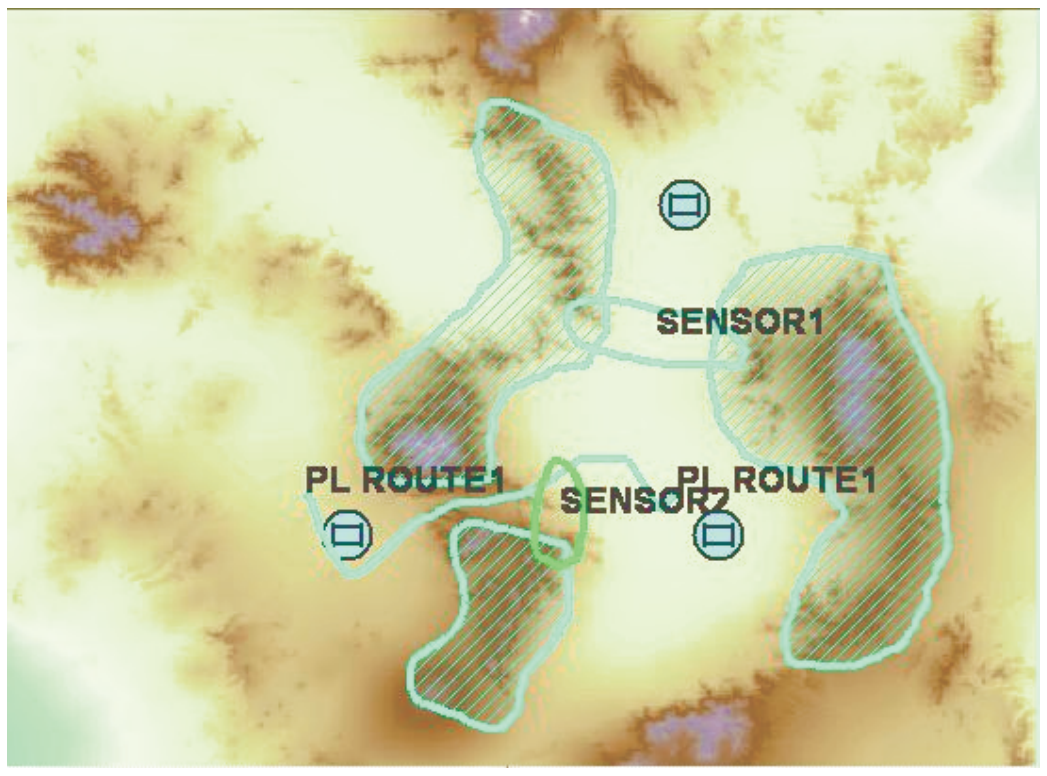

(c) A short path between $T_1$ and $T_3$.

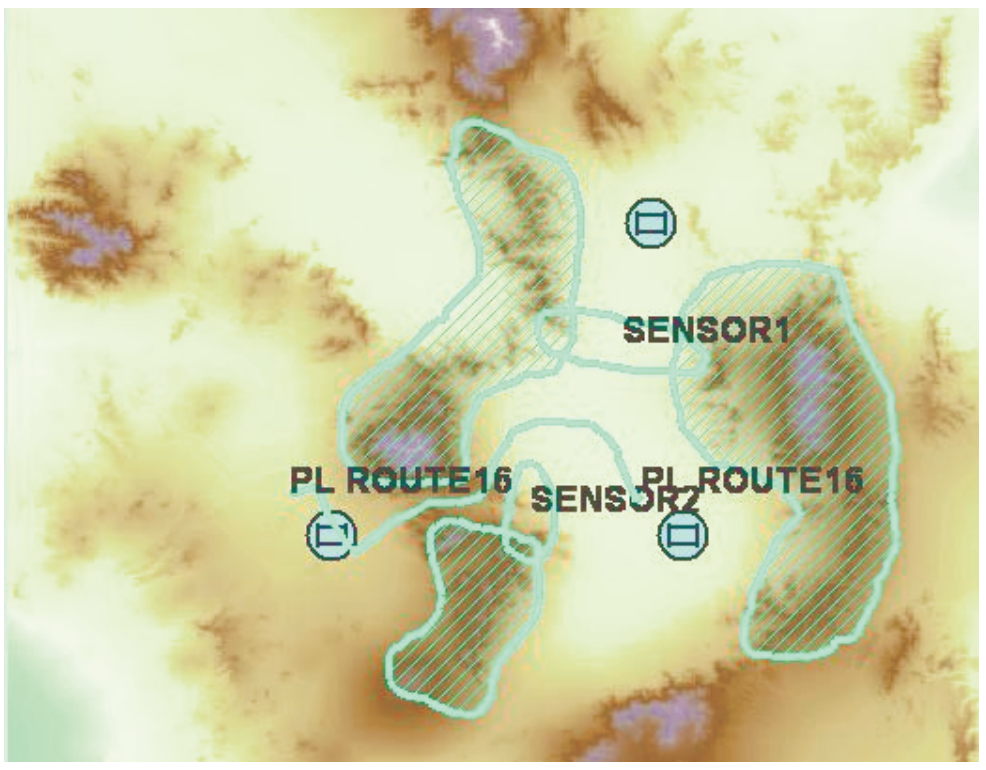

(d) A path from the only plausible homotopy class.

Figure 19: Reasoning steps for entity re-identification

area of interest, there are three enemy regions or obstacles $\{r_1, r_2, r_3\}$ (as shown in Figure 19(a)) with a given firepower/sight range $d$ of the enemy. Reasoning proceeds as follows. If $T_1$ can reach $p_3$ within the time $t_3 - t_1$, then $T_3$ might be $T_1$. Similarly for $T_2$. Since each mountainous region (or obstacle) is a hiding place for enemies with a firepower range $d$, the existence of an entity shows that it most probably did not traverse through a territory within the firepower range. Further, there might be sensor fields that report to the database when they sense entities. If no entity was sensed by a sensor field between the times $t_1$ and $t_3$, then $T_1$ could not have followed a path that passed through that sensor field. Such constraints have to be taken into account while reasoning. All information might not be available in the database at once. In what follows is a simple scenario and a discussion of the spatial problems as they occur.

The problem solver (e.g., a commander) wants to know whether there exists a contiguous safe region containing the points $p_1$ and $p_3$. He specifies the problem $SafeRegion$ as follows:

$$SafeRegion(q, \{r_1, r_2, ...r_n\}, d)$$

$\equiv \forall a, (\vee_{i \leftarrow 1}^{n} Inside(a, r_i)) \Rightarrow Distance(q, a) \geq d$

$\equiv \forall a, \neg(\vee_{i \leftarrow 1}^{n} \vee_{j \leftarrow 1}^{\#\triangle(r_i)} Inside(a, \triangle(r_i)[j])) \vee Distance(q, a) \geq d$

$\equiv \forall a, (\wedge_{i \leftarrow 1}^{n} \wedge_{j \leftarrow 1}^{\#\triangle(r_i)} \neg Inside(a, \triangle(r_i)[j])) \vee Distance(q, a) \geq d$

$\equiv \wedge_{i \leftarrow 1}^{n} \wedge_{j \leftarrow 1}^{\#\triangle(r_i)} \forall a, \neg Inside(a, \triangle(r_i)[j]) \vee Distance(q, a) \geq d$

$\equiv \wedge_{i \leftarrow 1}^{n} \wedge_{j \leftarrow 1}^{\#\triangle(r_i)} \forall a, Inside(a, \triangle(r_i)[j]) \Rightarrow Distance(q, a) \geq d$

$\equiv \wedge_{i \leftarrow 1}^{n} \wedge_{j \leftarrow 1}^{\#\triangle(r_i)} SafeRegion'(q, \{\triangle(r_i)[j]\}, d)$

where $q \leftarrow (x, y)$. Decomposition of the problem by the SPS is shown above. The subproblem is symbolically solved and the solution stored in memory along with the subproblem specification. In order to compare the actual times required to solve the problem, we constructed a very simple diagram consisting of four polygonal regions depicting obstacles (see Figure 20(a)). The four regions are

$r_1 \leftarrow \{(10, 10), (30, 10), (30, 30), (10, 30)\}$,

$r_2 \leftarrow \{(-20, 0), (0, 0), (-10, 20)\}$,

$r_3 \leftarrow \{(0, 20), (10, 40), (-10, 40)\}$,

$r_4 \leftarrow \{(50, 20), (70, 20), (80, 40), (60, 50), (40, 40)\}$,

while $d \leftarrow 2$. Triangulation of the regions produced seven triangles. Once a subproblem is symbolically solved and stored, solving the problem required 0.25 seconds while solving the same using the CAD algorithm required 5.5 seconds.

The diagram with the shaded safe region is input to the *Recognize* function which computes the vertices and boundaries of the shaded region, as shown in Figure 20(b). Next the problem solver wants to know whether there exists a path between points $p_1$ and $p_3$ safely avoiding the obstacles and enemy firepower range, and whether that path can be traversed in time $t_3 - t_1$. Let $v$ be the velocity of the sighted entity – a piece of symbolic knowledge available from the database. Then, the maximum length of path traversable in the given time is $L = v \times (t_3 - t_1)$. Let $l \ll L$ be a rational number. Then, the problem of path existence between two points $s$ and $t$ such that the path lies inside a region $r$ and is less than a given length $l$ can be specified as:

$PathExists(s, t, r, l)$

$\equiv \exists q, Inside(q, r) \wedge Distance(s, q) + Distance(q, t) \leq l$

$\equiv \exists q, (\vee_{i \leftarrow 1}^{\#\triangle(r)} Inside(q, \triangle(r)[i])) \wedge Distance(s, q) + Distance(q, t) \leq l$

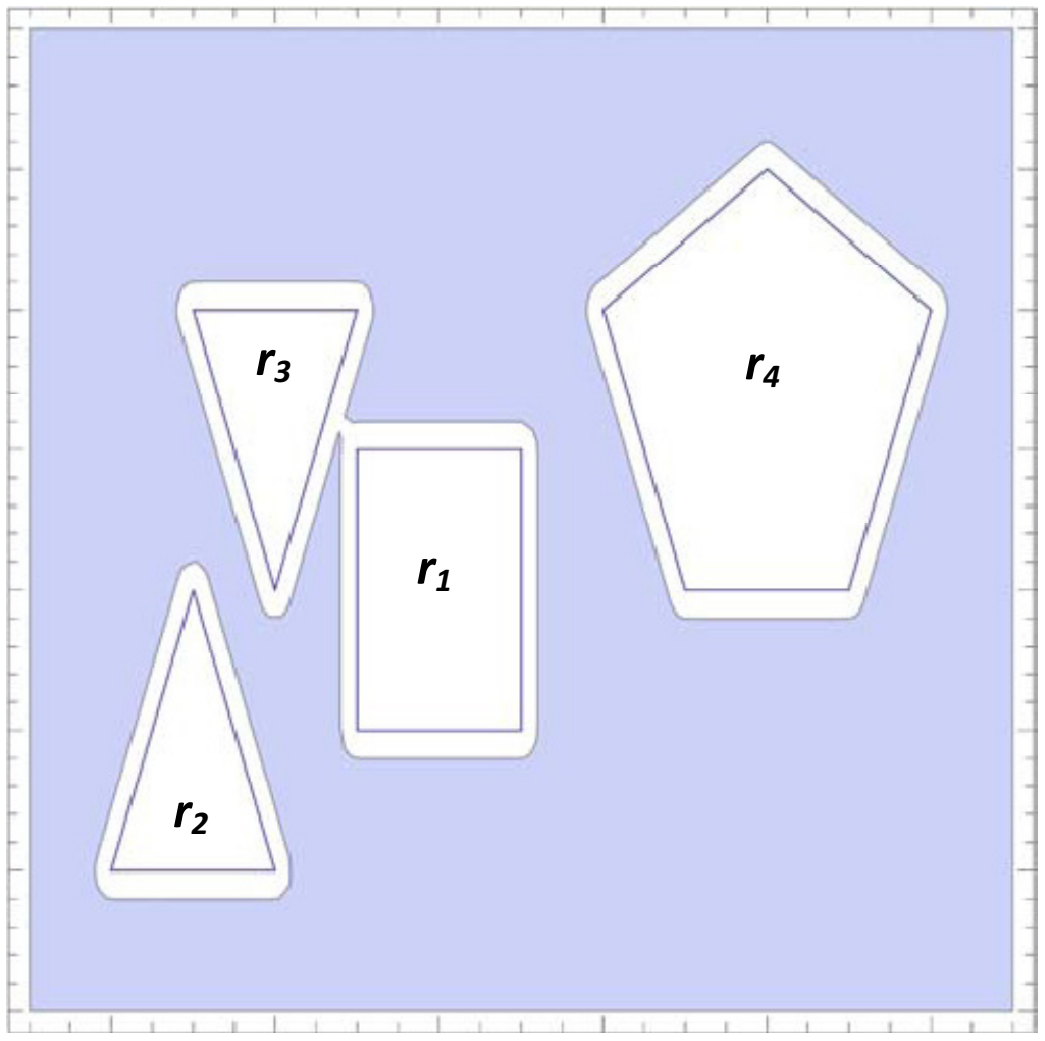

(a) The unshaded polygons are obstacles. The shaded region is the safe region, as computed by the SPS.

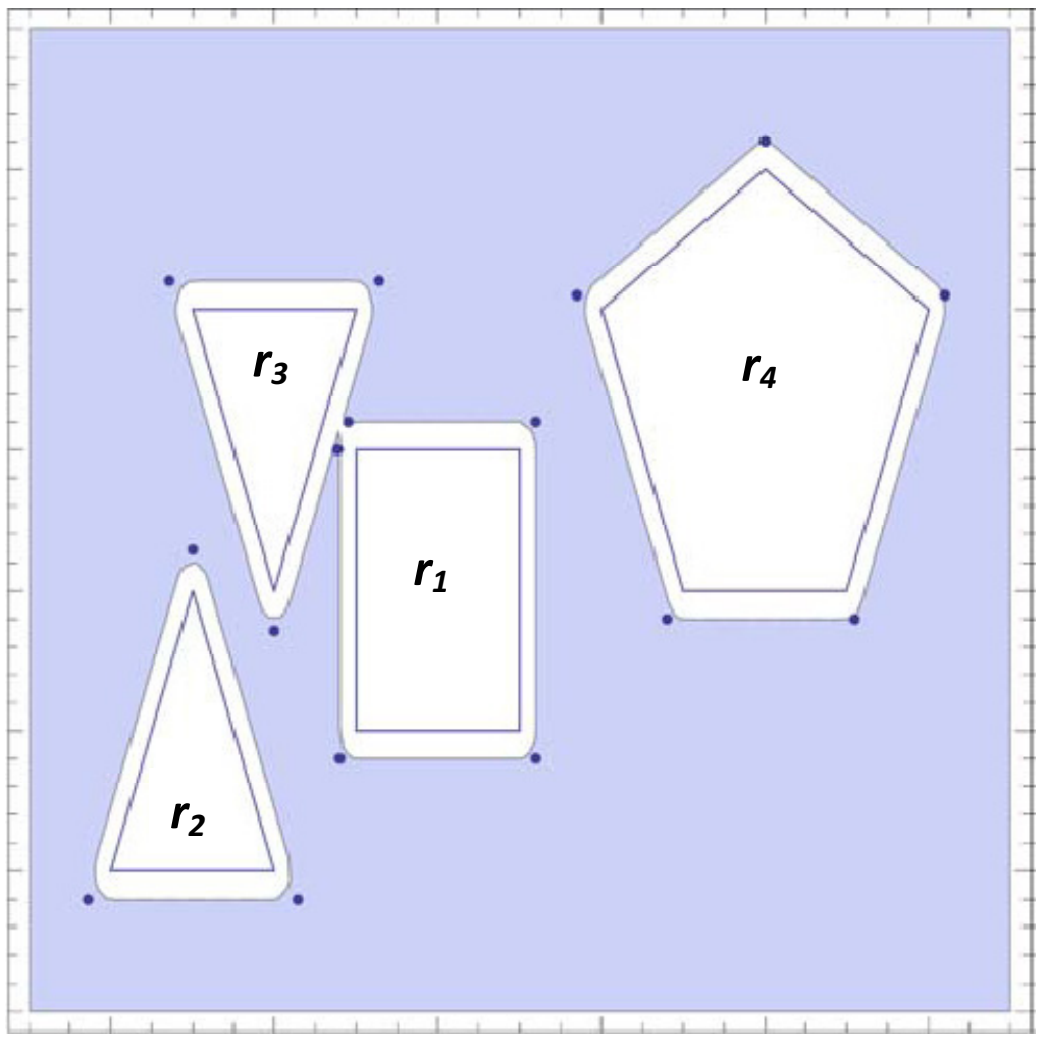

(b) The points shown are vertices of the boundaries of the safe region as computed by the *Recognize* function.

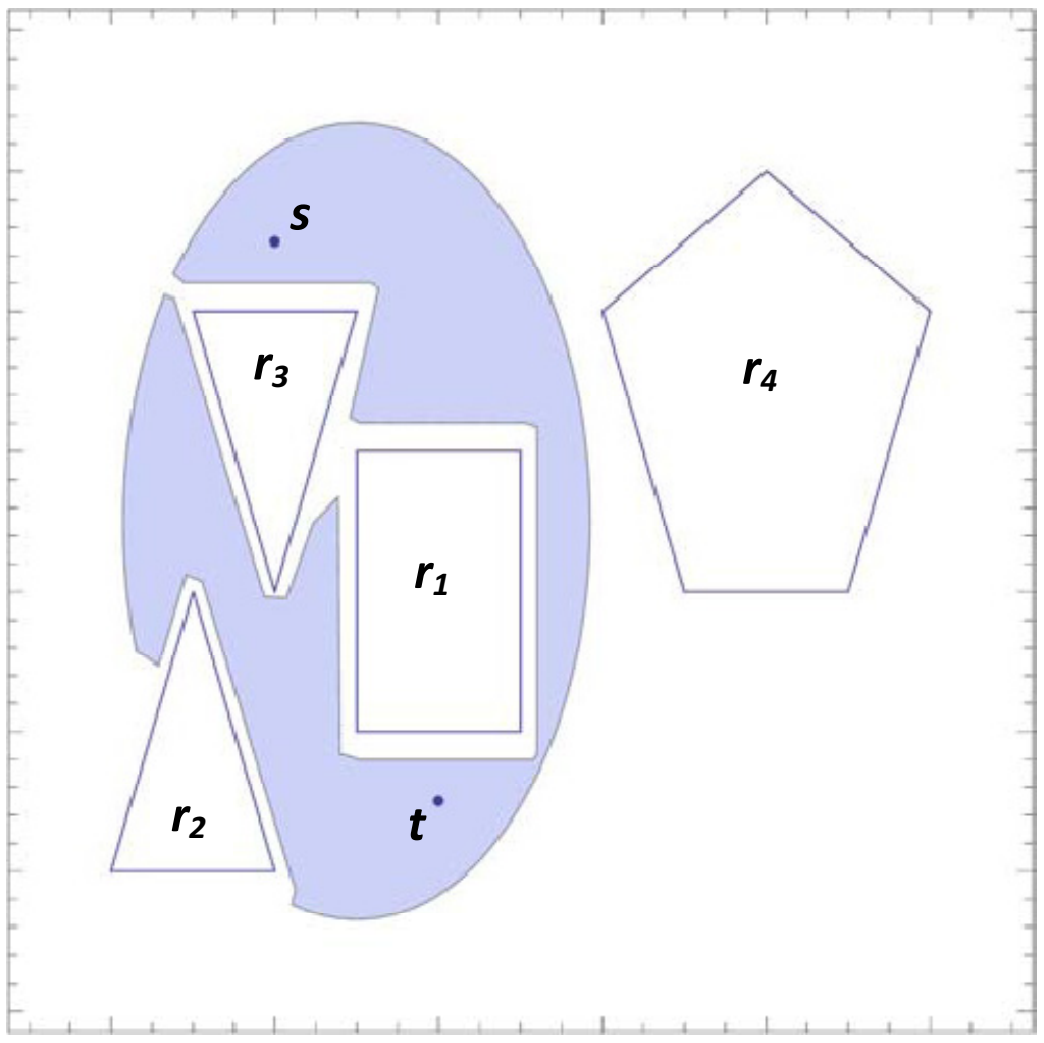

(c) Paths lying in the safe region and less than a given length between two points, as computed by the SPS.

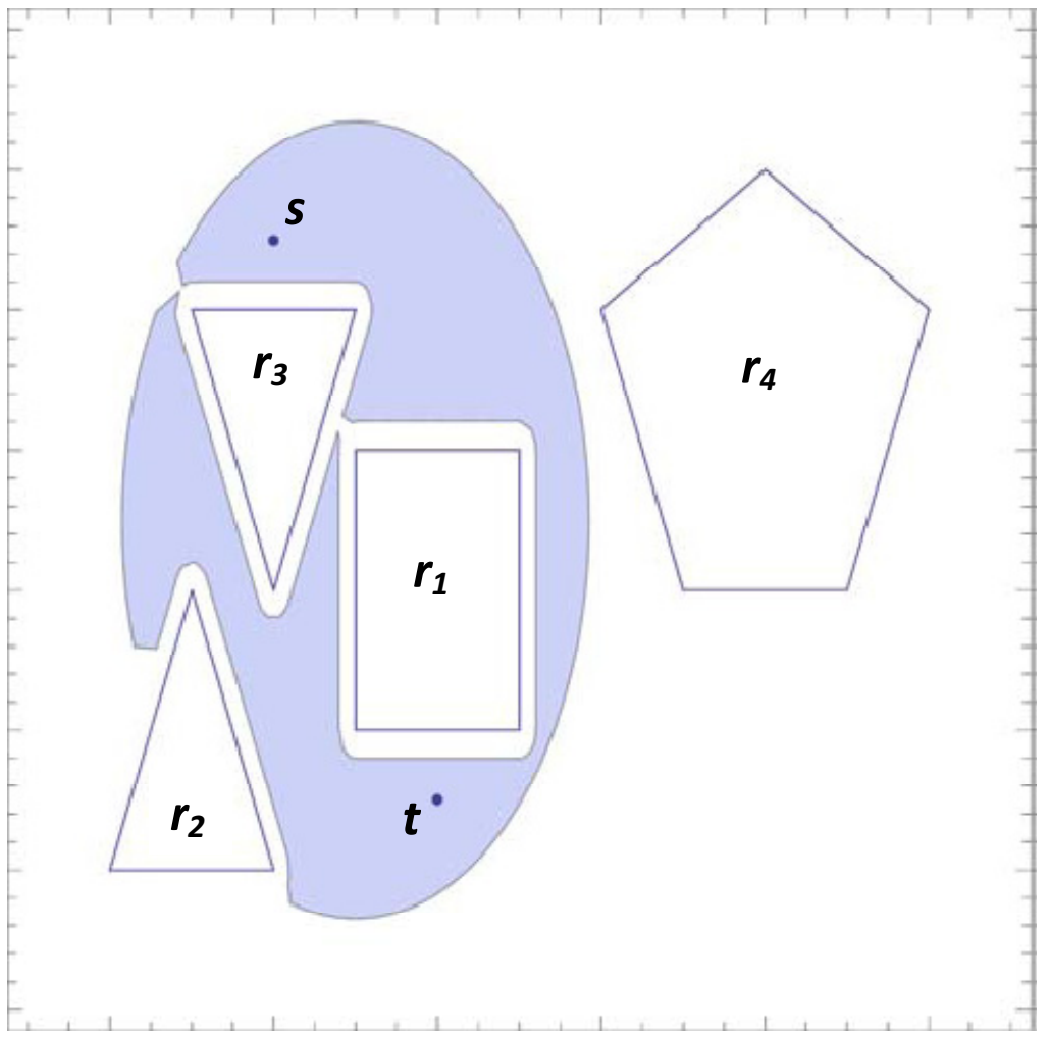

(d) Paths lying in the safe region and less than a given length between two points, as computed by the CAD algorithm.

Figure 20: A simplified scenario to illustrate the performance of the proposed SPS as compared to the CAD algorithm for entity re-identification.

$$\equiv \vee_{i \leftarrow 1}^{\#_{\triangle}(r)} \exists q, Inside(q, \triangle(r)[i]) \wedge Distance(s, q) + Distance(q, t) \leq l$$

$$\equiv \vee_{i \leftarrow 1}^{\#_{\triangle}(r)} PathExists'(s, t, \triangle(r)[i], l)$$

Decomposition of the problem by the SPS is shown above. The subproblem is symbolically solved and stored. Again, we resort to the simple diagram in Figure 20 to compare the actual computation times for the $PathExists(s,t,r,l)$ problem, where $s \leftarrow (0,45)$, $t \leftarrow (20,5)$, $r \leftarrow Recognize(SafeRegion((x,y),\{r_1,r_2,...r_n\},2))$, and different sets of regions $r_i$ and different values of $l$. Triangulation of $r$ produced 8, 7, 7, 9 and 24 triangles for $\{r_1\}$, $\{r_2\}$, $\{r_3\}$, $\{r_4\}$ and $\{r_1,r_2,r_3,r_4\}$ respectively. Once a subproblem is symbolically solved and stored, the computation times required for solving the problem using the proposed SPS is significantly less than that using the CAD algorithm (see Table 1).

Table 1: Comparison of computation times (in seconds) between the CAD algorithm and our proposed SPS for the $PathExists(s,t,r,l)$ problem, where $s \leftarrow (0,45)$, $t \leftarrow (20,5)$, $r \leftarrow Recognize(SafeRegion((x,y),\{r_1,r_2,...r_n\},2))$. A 2.8 GHz PC with 4 GB RAM, 5356 MB virtual memory and 32-bit operating system was used. The implementation was done in *Mathematica*. Below, "res" refers to result, "$T$" refers to $True$, "$F$" refers to $False$, and "$OOM$" refers to out of memory.

| $l$ | $\{r_1\}$ CAD,SPS,res | $\{r_2\}$ CAD,SPS,res | $\{r_3\}$ CAD,SPS,res | $\{r_4\}$ CAD,SPS,res | $\{r_1,r_2,r_3,r_4\}$ CAD,SPS,res |
|---|---|---|---|---|---|
| $\sqrt{100}$ | $2.78, 0.41, F$ | $498.22, 0.53, F$ | $470.74, 0.5, F$ | $OOM, 0.47, F$ | $OOM, 1.42, F$ |
| $\sqrt{500}$ | $2.77, 0.42, F$ | $482.77, 0.44, F$ | $476.97, 0.5, F$ | $OOM, 0.48, F$ | $OOM, 1.42, F$ |
| $\sqrt{1000}$ | $2.66, 0.39, F$ | $118.97, 0.55, T$ | $135.03, 0.49, T$ | $OOM, 0.49, T$ | $OOM, 1.42, F$ |
| $\sqrt{1009}$ | $2.28, 0.42, T$ | $119.28, 0.52, T$ | $134.75, 0.5, T$ | $OOM, 0.52, T$ | $OOM, 1.42, F$ |
| $\sqrt{1010}$ | $2.28, 0.41, T$ | $120.06, 0.53, T$ | $135.02, 0.52, T$ | $OOM, 0.47, T$ | $OOM, 1.41, T$ |
| $\sqrt{2000}$ | $1.88, 0.39, T$ | $120.38, 0.42, T$ | $135.3, 0.38, T$ | $OOM, 0.44, T$ | $OOM, 1.19, T$ |
| $\sqrt{4000}$ | $1.88, 0.34, T$ | $120.73, 0.39, T$ | $135.08, 0.34, T$ | $OOM, 0.31, T$ | $OOM, 1.0, T$ |
| $\sqrt{8000}$ | $1.88, 0.33, T$ | $121.58, 0.34, T$ | $135.03, 0.33, T$ | $OOM, 0.39, T$ | $OOM, 0.98, T$ |
| $\sqrt{16000}$ | $1.88, 0.33, T$ | $121.45, 0.34, T$ | $135.13, 0.36, T$ | $OOM, 0.36, T$ | $OOM, 0.92, T$ |

In general, the $PathExists(s,t,r,l)$ problem can be specified as:

$PathExists(s,t,r,l)$

$$\equiv \exists q_1,q_2,...q_n, (\forall a, On(a,\{s,q_1,q_2,...q_n,t\}) \Rightarrow Inside(a,r)) \wedge Length(\{s,q_1,q_2,...q_n,t\}) \leq l$$

$$\equiv \exists q_1,q_2,...q_n, (\forall a, \neg On(a,c) \vee Inside(a,r)) \wedge Length(c) \leq l$$

$$\equiv \exists q_1,q_2,...q_n, (\forall a, \neg On(a,c) \vee \neg Inside(a,\overline{r})) \wedge Length(c) \leq l$$

$$\equiv \exists q_1,q_2,...q_n, (\forall a, \neg On(a,c) \vee \neg(\vee_{k\leftarrow 1}^{\#_\triangle(\overline{r})} Inside(a,\triangle(\overline{r})[k]))) \wedge Length(c) \leq l$$

$$\equiv \exists q_1,q_2,...q_n, (\forall a, \neg On(a,c) \vee (\wedge_{k\leftarrow 1}^{\#_\triangle(\overline{r})} \neg Inside(a,\triangle(\overline{r})[k]))) \wedge Length(c) \leq l$$

$\equiv \wedge_{k\leftarrow 1}^{\#\triangle(\overline{r})} \exists q_1, q_2, ...q_n, (\forall a, \neg On(a, c) \vee \neg Inside(a, \triangle(\overline{r})[k])) \wedge Length(c) \leq l$

$\equiv \wedge_{k\leftarrow 1}^{\#\triangle(\overline{r})} PathExists'(s, t, \triangle(\overline{r})[k], l)$

where $c \equiv \{s, q_1, q_2, ...q_n, t\}$ and $\overline{r} \equiv \mathcal{B}{-}r$. Note that even though $c$ is a curve, $On(a, c)$ cannot be decomposed since $c$ is not a free variable (see definition of *Decomposition* in section 3.2). Also, note that the above problem contains the form $F2$ discussed in *Theorem* 1, so $\overline{r}$ has been used.

If there exists a path between points $p_1$ and $p_3$ safely avoiding the obstacles and enemy firepower range such that it can be traversed in time $t_3 - t_1$, then the problem solver wants to compute the path(s). The problem can be specified as:

$FindPath(q, s, t, r, l)$

$\equiv Inside(q, r) \wedge Distance(s, q) + Distance(q, t) \leq l$

$\equiv (\vee_{i\leftarrow 1}^{\#\triangle(r)} Inside(q, \triangle(r)[i])) \wedge Distance(s, q) + Distance(q, t) \leq l$

$\equiv \vee_{i\leftarrow 1}^{\#\triangle(r)} Inside(q, \triangle(r)[i]) \wedge Distance(s, q) + Distance(q, t) \leq l$

$\equiv \vee_{i\leftarrow 1}^{\#\triangle(r)} FindPath'(q, s, t, \triangle(r)[i], l)$

where $q \leftarrow (x, y)$. Since there are no quantifiers, solving the problem by decomposition and variable mapping does not achieve reduction in computation time by any significant amount. The region consisting of all paths that satisfy the constraints ($l \leftarrow \sqrt{1010}$) is shown in Figure 20(c). The quality of the solution depends on the *Recognize* function. For example, the solution shown in Figure 20(d) is more accurate than in Figure 20(c) as the *Recognize* function failed to determine the vertices of the safe region accurately. An alternate definition of the semi-linear motion planning problem can be found in Weispfenning (2001), where a semi-linear path consists of $n$ translations along straight lines each of which is parallel to one of the given $k$ vectors.

From the results, the problem solver infers that $T_3$ might be $T_1$. Next he repeats the same for entities $T_3$ and $T_2$, and finds that there exists a path between points $p_2$ and $p_3$ safely avoiding the obstacles and enemy firepower range such that it can be traversed in time $t_3 - t_2$. So $T_3$ might be $T_2$ as well. The sensor database informs that there are two sensor fields – `SENSOR1`, `SENSOR2` – in the area of interest but there has been no report from them of any passing vehicle. Problem solver wants to verify whether any of the paths passes through any of the sensor fields. He specifies the problem $Intersect(r_1, r_2)$ to compute the intersection of two regions $r_1$ and $r_2$.

$IntersectRegions(r_1, r_2)$

$\equiv \exists q, Inside(q, r_1) \wedge Inside(q, r_2)$

$\equiv \exists q, (\vee_{i\leftarrow 1}^{\#\triangle(r_1)} Inside(q, \triangle(r_1)[i])) \wedge (\vee_{i\leftarrow 1}^{\#\triangle(r_2)} Inside(q, \triangle(r_2)[j]))$

$$\equiv \vee_{i \leftarrow 1}^{\#_\triangle(r_1)} \vee_{j \leftarrow 1}^{\#_\triangle(r_2)} IntersectRegions'(\triangle(r_1)[i], \triangle(r_2)[j])$$

He computes the problem $IntersectRegions(paths_{13}, s_1)$ where $paths_{13} \leftarrow Recognize(FindPath(q, p_1, p_3, r, l))$ and $s_1$ is the region covered by SENSOR1. In our scenario in Figure 19(c), the solution is $True$. Next the problem solver wants to know whether there exists a path between points $p_1$ and $p_3$ safely avoiding the obstacles and enemy firepower range such that it can be traversed in time $t_3 - t_1$. He computes $PathExists(p_1, p_3, r_{13}, l)$, where $r_{13} \leftarrow Recognize(paths_{13} - s_1)$, which returns $True$. The inference follows that $T_3$ might be $T_2$. The same reasoning is repeated for $T_3$ and $T_2$; $Intersect(paths_{23}, s_2)$ returns $True$ while $PathExists(p_2, p_3, r_{23}, l)$ returns $False$ (see Figure 19(b)). The inference follows that $T_3$ cannot be $T_1$. Hence, the problem solver identifies $T_3$ as $T_2$.

The entity reidentification problem could also have been solved by computing the shortest paths between the pairs $p_1$, $p_3$ and $p_2$, $p_3$ avoiding the sensors and checking whether their lengths satisfy the time constraints. That requires computing the shortest path between two points $p_1$ and $p_3$ safely avoiding the obstacles and enemy firepower range (i.e., lying entirely within the safe region $r$). Since such a path will not have any loop and will share its intermediate vertices, if any, with the vertices of $r$, the path can have at most $\#(r)$ intermediate vertices. Let $S \leftarrow r \cup \{p_1, p_3\}$, $m \leftarrow \#(S)$, and $c \leftarrow \{q_1, q_2, ...q_m\}$ be the shortest path, where $q_1 \equiv p_1$, $q_m \equiv p_3$, $q_i \in r$ $(2 \leq i \leq m-1)$. Then, the problem of computing the shortest path can be specified as

$$FindShortestPath(r, c)$$

$$\equiv Minimize(Length(c), \{c[2], c[3], ...c[m-1]\}, CurveInsideRegion(c, r))$$

where $CurveInsideRegion(c, r)$ is the constraint that can be specified and decomposed as follows.

$$CurveInsideRegion(c, r)$$

$$\equiv \forall a, On(a, c) \Rightarrow Inside(a, r)$$

$$\equiv \forall a, \neg On(a, c) \vee Inside(a, r)$$

$$\equiv \forall a, \neg On(a, c) \vee \neg Inside(a, \overline{r})$$

$$\equiv \forall a, \neg(\vee_{i \leftarrow 1}^{\#(c)-1} On(a, \{c[i], c[i+1]\})) \vee \neg(\vee_{j \leftarrow 1}^{\#_\triangle(\overline{r})} Inside(a, \triangle(\overline{r})[j])))$$

$$\equiv \wedge_{i \leftarrow 1}^{\#(c)-1} \wedge_{j \leftarrow 1}^{\#_\triangle(\overline{r})} \forall a, \neg On(a, \{c[i], c[i+1]\}) \vee \neg Inside(a, \triangle(\overline{r})[j])$$

$$\equiv \wedge_{i \leftarrow 1}^{\#(c)-1} \wedge_{j \leftarrow 1}^{\#_\triangle(\overline{r})} CurveInsideRegion'(\{c[i], c[i+1]\}, \triangle(\overline{r})[j])$$

where $\overline{r} \equiv \mathcal{B}-r$. Since the above problem is of the form $F2$, $\overline{r}$ had to be used. Once a subproblem is symbolically solved and stored, solving the problem $CurveInsideRegion(c, r)$, where

$$c \leftarrow \{(0,45),(14,42),(35,42),(15,35),(34,32),(36,19),(47,15),(87,15),(30,-7),(20,5)\},$$

$$r \leftarrow Recognize(SafeRegion((x,y),\{r_1,r_2,r_3,r_4\},2)),$$

by SPS required 3.11 seconds while solving the same using the CAD algorithm required 175.01 seconds (see Figure 21(a)). The shortest path obtained by solving the $FindShortestPath(r, c)$ problem is shown in Figure 21(b).

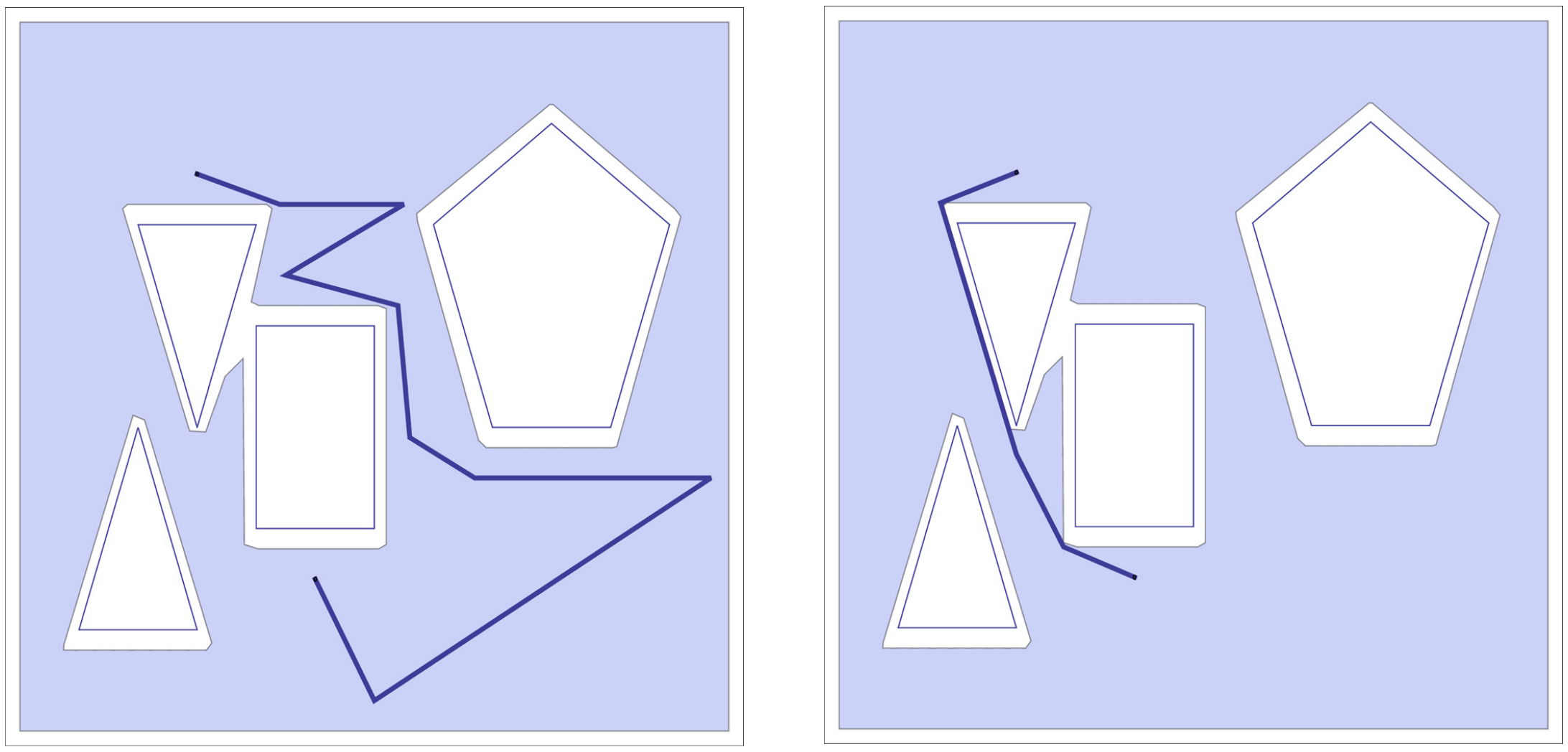

(a) A path $c$ between two points lying inside the shaded region $r$.

(b) Shortest path between two points as computed from the $FindShortestPath(r, c)$ problem.

Figure 21: Paths between two points lying inside the safe (shaded) region.

### 5.2 Ambush Analysis

There are two main factors – range of firepower and sight – that determine the area covered by a military unit. Presence of terrain features, such as, mountains, limit these factors and allow units to hide from opponents. These hidden units not only enjoy the advantage of concealing their resources and intentions from the opponents but can also attack the opponents catching them unawares if they are traveling along a path that is within the sight and firepower range of the hidden units, thereby ambushing them. Thus, it is of utmost importance for any military unit to a priori determine the areas or portions of a path prone to ambush before traversing them. We had already described in section 1.1 how a problem solver (e.g., an army commander) reasons using diagrams to figure out the safest path to transport his troops from one base camp to another in a given time. In this section, given a curve or region as a hiding place and the firepower and sight ranges, we show how the regions and portions of path prone to ambush is efficiently computed by the proposed SPS.

Given a curve $c$ and the firepower and sight range $d$, the spatial problem $RiskyRegion(q,c,d)$ is defined as the set of all points covered by that range from $c$. Thus, the problem specification is:

$RiskyRegion(q,c,d)$

$\equiv \exists a, On(a,c) \wedge Distance(a,q) \leq d$

$\equiv \exists a, (\vee_{i\leftarrow 1}^{\#(c)-1} On(a,\{c[i],c[i+1]\})) \wedge Distance(a,q) \leq d$

$\equiv \vee_{i\leftarrow 1}^{\#(c)-1} \exists a, On(a,\{c[i],c[i+1]\}) \wedge Distance(a,q) \leq d$

$\equiv \vee_{i\leftarrow 1}^{\#(c)-1} RiskyRegion'(q,\{c[i],c[i+1]\},d)$

where $q \leftarrow (x,y)$. In order to compare the actual computation times required to solve the problem, we constructed a very simple diagram consisting of two curves, *path* and *mntn*, where

$path \leftarrow \{(-25,-10),(-5,-10),(-3,-15),(-7,-17),(-2,-18),(2,-18),(7,-15),$
$(3,-12),(5,-10),(40,-10)\}$

$mntn \leftarrow \{(-5,5),(-7,2),(-9,9),(-6,12),(0,4),(2,3),(15,5),(25,12),(30,20)\}$

The solution to the problem $RiskyRegion(q,mntn,d)$ is the shaded region shown in Figure 22(a) where *mntn* is an obstacle for hiding (e.g., mountain range) and $d \leftarrow 15$. The problem $RiskyRegion(q,r,d)$ for a region $r$ can be specified by replacing the predicate $On(p,c)$ by $Inside(p,r)$.

Again, given a curve $c_1$ as a path, a curve $c_2$ for hiding, and a firepower range $d$, the problem $RiskyPortionsofPath(q,c_1,c_2,d)$ is defined as parts of $c_1$ covered by that range from $c_2$. Thus,

$RiskyPortionsofPath(q,c_1,c_2,d)$

$\equiv On(q,c_1) \wedge \exists p, On(p,c_2) \wedge Distance(p,q) \leq d$

$\equiv \exists p, (\vee_{i\leftarrow 1}^{\#(c_1)-1} On(q,\{c_1[i],c_1[i+1]\})) \wedge (\vee_{j\leftarrow 1}^{\#(c_2)-1} On(p,\{c_2[j],c_2[j+1]\})) \wedge Distance(p,q) \leq d$

$\equiv \vee_{i\leftarrow 1}^{\#(c_1)-1} \vee_{j\leftarrow 1}^{\#(c_2)-1} \exists p, On(q,\{c_1[i],c_1[i+1]\})) \wedge On(p,\{c_2[j],c_2[j+1]\}) \wedge Distance(p,q) \leq d$

$\equiv \vee_{i\leftarrow 1}^{\#(c_1)-1} \vee_{j\leftarrow 1}^{\#(c_2)-1} RiskyPortionsofPath'(q,\{c_1[i],c_1[i+1]\},\{c_2[j],c_2[j+1]\},d)$

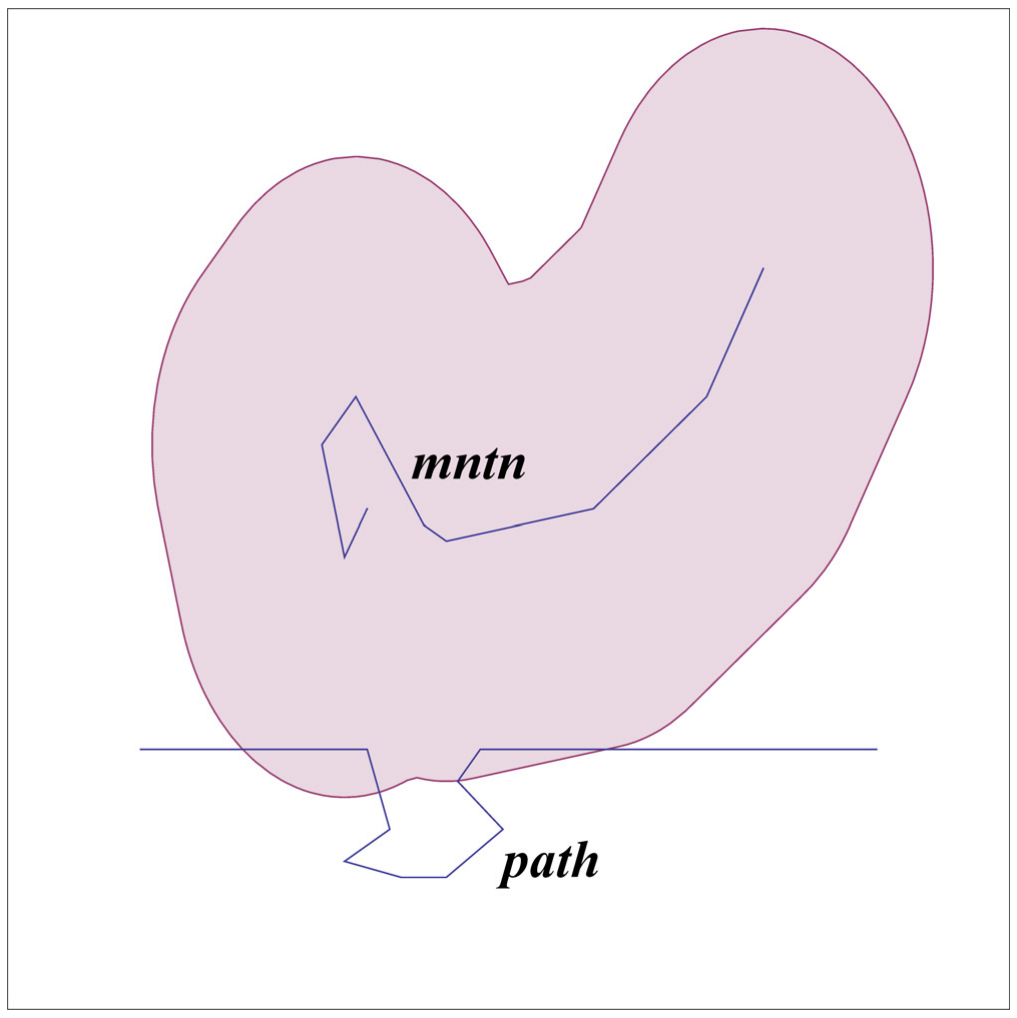

(a) The shaded region, as computed from the problem $RiskyRegion(q, mntn, 15)$, is the risky region prone to ambush due to enemies hiding at $mntn$. The portions of $path$ inside the risky region are the risky portions of the path.

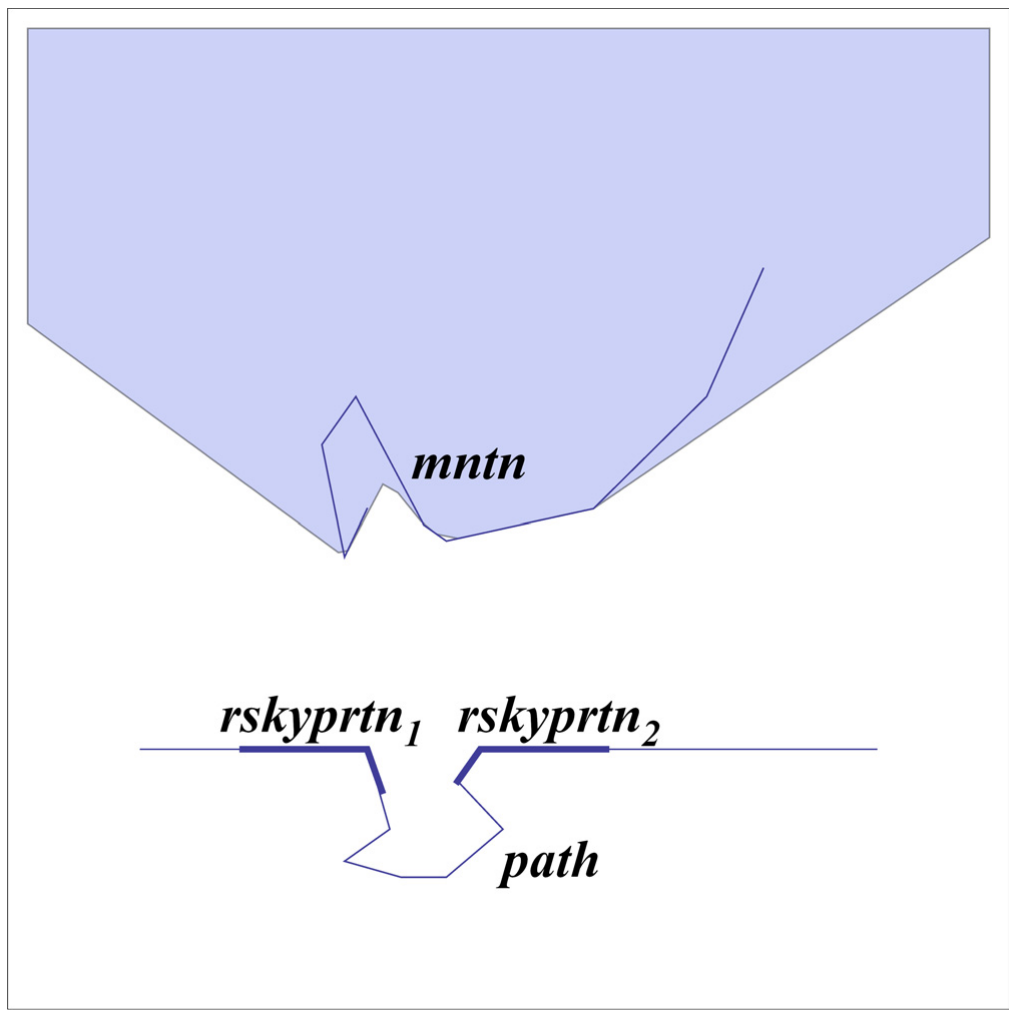

(b) The bold parts of $path$, as computed from the problem $RiskyPortionsofPath(q, path, mntn, 15)$, are the risky portions of the path. The shaded region, as computed from the problem $BehindCurvewrtRiskyPath(q, mntn, path)$, is where enemies could be hiding from troops traveling on $path$.

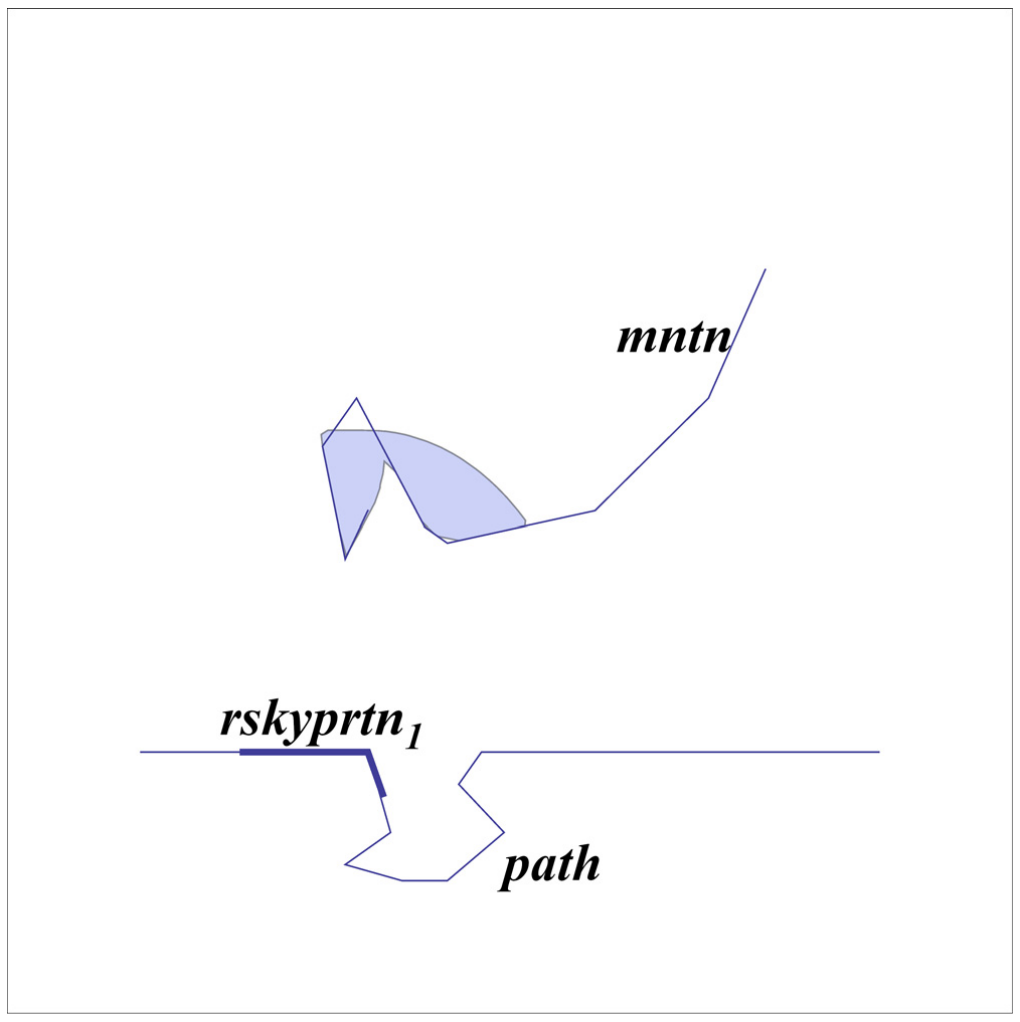

(c) Troops traveling on $rskyprtn_1$, a risky portion (in bold) of $path$, should be careful of being ambushed by enemies hiding in the shaded region because $rskyprtn_1$ is within firepower range from that region, computed from the problem $BehindCurvewrtRiskyPathDistance(q, mntn, rskyprtn_1, 20)$.

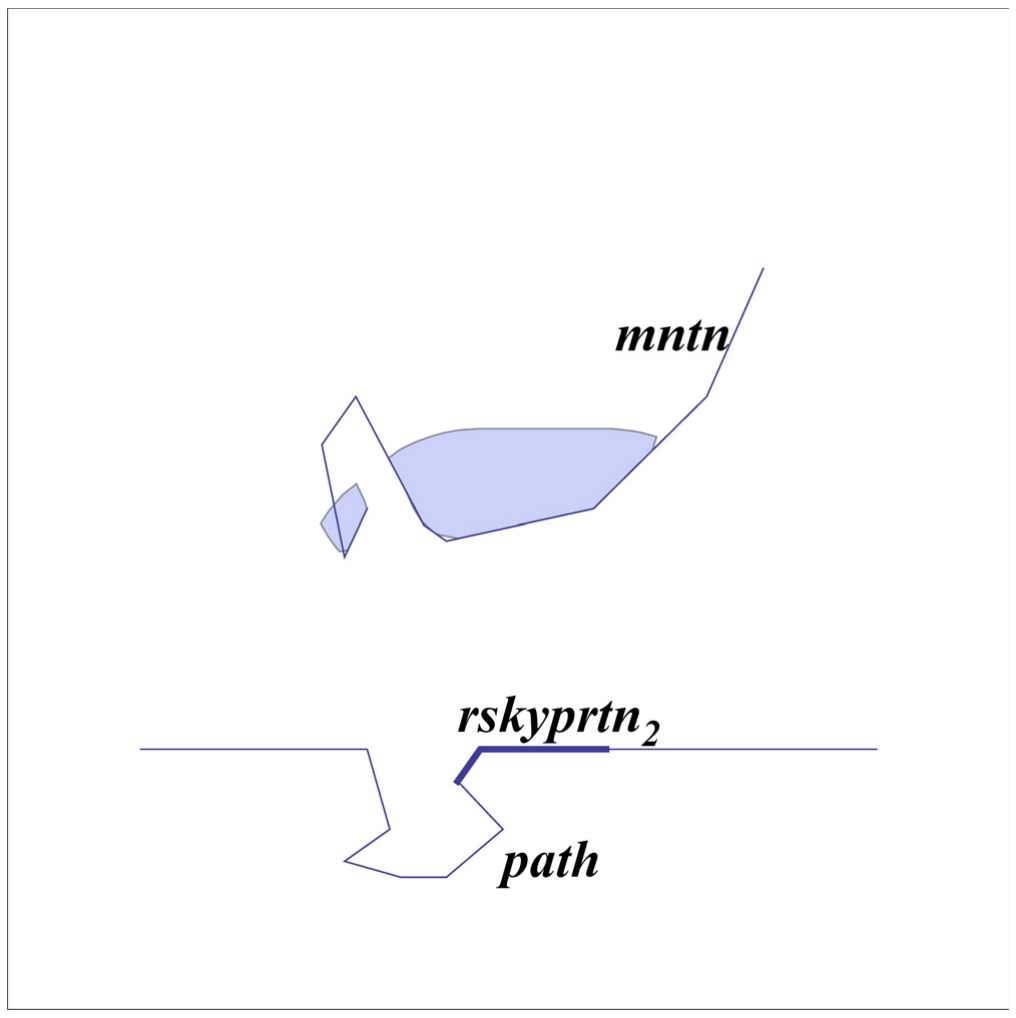

(d) Troops traveling on $rskyprtn_2$, a risky portion (in bold) of $path$, should be careful of being ambushed by enemies hiding in the shaded region because $rskyprtn_2$ is within firepower range from that region, computed from the problem $BehindCurvewrtRiskyPathDistance(q, mntn, rskyprtn_2, 20)$.

Figure 22: A simplified scenario to illustrate the performance of the proposed SPS for ambush analysis.

where $q \leftarrow (x, y)$. Alternatively, the same problem can be specified as

$$RiskyPortionsofPath(q, c_1, r_2, d)$$

$$\equiv On(q, c_1) \wedge Inside(q, r_2)$$

$$\equiv (\vee_{i \leftarrow 1}^{\#(c_1)-1} On(q, \{c_1[i], c_1[i+1]\})) \wedge (\vee_{j \leftarrow 1}^{\#_{\triangle}(r_2)} Inside(q, \triangle(r_2)[j]))$$

$$\equiv \vee_{i \leftarrow 1}^{\#(c_1)-1} \vee_{j \leftarrow 1}^{\#_{\triangle}(r_2)} On(q, \{c_1[i], c_1[i+1]\}) \wedge Inside(q, \triangle(r_2)[j])$$

$$\equiv \vee_{i \leftarrow 1}^{\#(c_1)-1} \vee_{j \leftarrow 1}^{\#_{\triangle}(r_2)} RiskyPortionsofPath'(q, \{c_1[i], c_1[i+1]\}, \triangle(r_2)[j], d)$$

where $r_2 \leftarrow Recognize(RiskyRegion((x, y), c_2, d))$ and $q \leftarrow (x, y)$. The solution to the problem $RiskyPortionsofPath(q, path, mntn, d)$, where $d \leftarrow 15$, is the parts of *path* inside the shaded region shown in Figure 22(a). Figure 22(b) shows the risky portions of the path – $rskyprtn_1$, $rskyprtn_2$ – in bold as obtained from $Recognize(RiskyPortionsofPath(q, c_1, c_2, d))$.

$$rskyprtn_1 \leftarrow \{(-16, -10), (-5, -10), (-3.7, -12.6)\}$$

$$rskyprtn_2 \leftarrow \{(3, -12), (5, -10), (16.1, -10)\}$$

Note that the latter specification is free from quantifiers while the former is not. However, the solution computed from the latter specification might have less accuracy than the same from the former due to the use of *Recognize* function. If the hiding place is a region $r$ instead of the curve $c_2$, the problem $RiskyPortionsofPath(q, c_1, r, d)$ can be specified by replacing the predicate $On(p, c_2)$ by $Inside(p, r)$. The portions of the path marked in Figure 2(c) is computed from this specification.

The region behind $c_2$ where the enemies might be hiding is the set of all points that are behind $c_2$ with respect to each point on the risky portions of curve $c_1$. Thus, if $c$ is a risky portion of a path, we have

$$BehindCurvewrtRiskyPath(q, c_2, c)$$

$$\equiv \exists a, On(a, c) \wedge BehindCurve(q, c_2, a)$$

$$\equiv \exists a, On(a, c) \wedge Intersect(c_2, \{a, q\})$$

$$\equiv \exists a, On(a, c) \wedge (\exists b, On(b, c_2) \wedge On(b, \{a, q\}))$$

$$\equiv \exists a, b, (\vee_{i \leftarrow 1}^{\#(c)-1} On(a, \{c[i], c[i+1]\})) \wedge (\vee_{j \leftarrow 1}^{\#(c_2)-1} On(b, \{c_2[j], c_2[j+1]\})) \wedge On(b, \{a, q\})$$

$$\equiv \vee_{i \leftarrow 1}^{\#(c)-1} \vee_{j \leftarrow 1}^{\#(c_2)-1} (\exists a, b, On(a, \{c[i], c[i+1]\}) \wedge On(b, \{c_2[j], c_2[j+1]\}) \wedge On(b, \{a, q\}))$$

$\equiv \vee_{i \leftarrow 1}^{\#(c)-1} \vee_{j \leftarrow 1}^{\#(c_2)-1} BehindCurvewrtRiskyPath'(q, \{c_2[j], c_2[j+1]\}, \{c[i], c[i+1]\})$

where
$q \leftarrow (x, y)$. The solution to the problem $BehindCurvewrtRiskyPath(q, mntn, rskyprtn_1) \vee BehindCurvewrtRiskyPath(q, mntn, rskyprtn_2)$ is the shaded region shown in Figure 22(b). If the hiding place is a region $r$ instead of the curve $c_2$, the problem $BehindCurvewrtRiskyPath(q, r, c)$ can be specified by replacing the predicate $On(p, c_2)$ by $Inside(p, r)$.

However, the enemies might be hiding not anywhere behind a mountain but within a distance from where they can ambush the friendly units. Hence, a more reasonable problem for the commander from the friendly side to compute would be $BehindCurvewrtRiskyPathDistance(q, c_2, c, d)$ where $d$ is the distance from where the enemies can ambush them. The problem is specified as:

$BehindCurvewrtRiskyPathDistance(q, c_2, c, d)$

$\equiv \exists a, On(a, c) \wedge BehindCurve(q, c_2, a) \wedge Distance(a, q) \leq d$

$\equiv \exists a, On(a, c) \wedge Intersect(c_2, \{a, q\}) \wedge Distance(a, q) \leq d$

$\equiv \exists a, On(a, c) \wedge (\exists b, On(b, c_2) \wedge On(b, \{a, q\})) \wedge Distance(a, q) \leq d$

$\equiv \exists a, b, (\vee_{i \leftarrow 1}^{\#(c)-1} On(a, \{c[i], c[i+1]\})) \wedge (\vee_{j \leftarrow 1}^{\#(c_2)-1} On(b, \{c_2[j], c_2[j+1]\})) \wedge On(b, \{a, q\}) \wedge Distance(a, q) \leq d$

$\equiv \vee_{i \leftarrow 1}^{\#(c)-1} \vee_{j \leftarrow 1}^{\#(c_2)-1} (\exists a, b, On(a, \{c[i], c[i+1]\}) \wedge On(b, \{c_2[j], c_2[j+1]\}) \wedge On(b, \{a, q\})) \wedge Distance(a, q) \leq d$

$\equiv \vee_{i \leftarrow 1}^{\#(c)-1} \vee_{j \leftarrow 1}^{\#(c_2)-1} BehindCurvewrtRiskyPathDistance'(q, \{c_2[j], c_2[j+1]\}, \{c[i], c[i+1]\}, d)$

where $q \leftarrow (x, y)$.
The solutions to the problems $BehindCurvewrtRiskyPathDistance(q, mntn, rskyprtn_1, d)$ and $BehindCurvewrtRiskyPathDistance(q, mntn, rskyprtn_2, d)$, where $d \leftarrow 20$, are the shaded regions shown in Figure 22(c), 22(d) respectively. If the hiding place is a region $r$ instead of the curve $c_2$, the problem $BehindCurvewrtRiskyPathDistance(q, r, c, d)$ can be specified by replacing the predicate $On(p, c_2)$ by $Inside(p, r)$. A comparison between the CAD algorithm and our proposed SPS of actual times required to compute the problems relevant to ambush analysis as discussed above is shown in Table 2.

Table 2: Comparison of computation times (in seconds) between the CAD algorithm and our SPS for the different problems relevant to ambush analysis. A 2.8 GHz PC with 4 GB RAM, 5356 MB virtual memory and 32-bit operating system was used. The implementation was done in *Mathematica*. All of the following are function problems where $q \leftarrow (x, y)$.

| *Problem* | SPS | CAD |
|---|---|---|
| $RiskyRegion(q, mntn, 15)$ | 0.11 | 0.11 |
| $RiskyPortionsofPath(q, path, mntn, 15)$ | 0.3 | 0.48 |
| $BehindCurve(q, mntn, (5, -10))$ | 0.27 | 0.71 |
| $BehindCurvewrtRiskyPath(q, mntn, path)$ | 50.2 | 102.88 |
| $BehindCurvewrtRiskyPath(q, mntn, rskyprtn_1)$ | 8.04 | 11.89 |
| $BehindCurvewrtRiskyPath(q, mntn, rskyprtn_2)$ | 10.92 | 17.48 |
| $BehindCurvewrtRiskyPathDistance(q, mntn, rskyprtn_1, 20)$ | 6.16 | 16.08 |
| $BehindCurvewrtRiskyPathDistance(q, mntn, rskyprtn_2, 20)$ | 6.3 | 15.33 |

## 6. Discussion

Spatial problem solving has been an area of active research since Sutherland's sketchpad (1963). The need to access, communicate and manipulate spatial information precisely (much as engineers and scientists do) using a high-level language (much as common people use) has been one of the frontiers in AI. It has been well-known that such capabilities are offered by first-order predicate logic and that, first-order logic is generally intractable except for limited domains. Under the umbrella of Qualitative Spatial Reasoning (QSR), researchers have investigated a plethora of spatial calculi, the most prominent of which are mereotopological calculi (Clarke, 1981; Bennett, 1997), cardinal direction calculus (Frank, 1991, 1992; Skiadopoulos & Koubarakis, 2004), double cross calculus (Freksa, 1992), 4- and 9-intersection calculi (Egenhofer, 1991; Egenhofer & Franzosa, 1991), flip-flop calculus (Ligozat, 1993), dipole calculus (Moratz, Renz, & Wolter, 2000; Schlieder, 1995; Dylla & Moratz, 2005), and the various region connection calculi (Randell et al., 1992; Bennett, Isli, & Cohn, 1997; Gerevini & Nebel, 2002; Cohn, Bennett, Gooday, & Gotts, 1997; Duntsch, Wang, & McCloskey, 1999; Gerevini & Renz, 1998). There are two main points of distinction between QSR and our approach to spatial problem solving as reported in this paper.

1. The different QSR calculi emphasize different aspects of space, such as, ontological issues, topology, distance, orientation, shape, etc. Depending on the spatial aspect of interest, the calculus is based on a minimal set of spatial relations. For example, the 9-intersection calculus (Egenhofer & Franzosa, 1991) is based on nine spatial relations $\{r_0, r_1, r_3, r_6, r_7, r_{10}, r_{11}, r_{14}, r_{15}\}$ between two spatial regions, the double cross calculus (Freksa, 1992) is based on fifteen spatial relations $\{lf, lp, lc, ll, lb, sf, sp, sc, sl, sb, rf, rp, rc, rl, rb\}$ among three points, etc. Our framework is not based on a minimal set of spatial relations; it is based on a fixed set of mathematical/logical operators (see section 2.3). Any spatial relation among points that can be expressed using real variables and the fixed set of operators in first-order logic is included in our vocabulary. Any spatial relation involving curves and/or regions that can be expressed in first-order logic using spatial relations among points and the relations *On* and *Inside* is included in our vocabulary.

2. The spatial problems of interest to the QSR community are CSPs involving either points (e.g., double cross calculus) or regions (e.g., 4- and 9-intersection calculi, region connection calculi) and a closed set of their properties/relations often limited to the binary domain. A general-purpose SPS for helping a human perceive from and act on diagrams in different real-world applications will need to solve QCSPs involving points, curves and regions and an open-ended vocabulary of their properties/relations/actions over the entire real domain, which is what our framework offers. Since QE is the computational bottleneck of our SPS, we concentrate our efforts on the real QE algorithms, as discussed towards the end of section 2.3.

Naturally, the question arises – how convenient is it for a human to specify a spatial problem as a QCSP? While we acknowledge that the process of specifying a spatial problem as a QCSP is not as effortless as explaining it to another human in a natural language, we have taken the first step in making the process less strenuous by offering a vocabulary of predicates that is open-ended. Not all QCSP-solving systems, such as, REDLOG (Dolzmann & Sturm, 1999) and QEPCAD (Brown, 2003), offer such a vocabulary for spatial problem solving which makes it difficult for a user to specify a problem as he has to dig deep into an ocean of equations and inequalities and cannot communicate naturally in terms of high-level predicates.[4] We are still far from building systems that can understand communication in natural language. However, research in automatic constraint acquisition from examples is already underway. Vu and O'Sullivan (2008) discuss recent advances in that direction. While we did not use any of those ideas or results in our work, it is not difficult to see how those ideas in conjunction with the work reported in this paper will be able to build a more convenient and efficient spatial problem solving framework.

## 7. Conclusion

DR requires perceiving specified information from a diagram or modifying/creating objects in a diagram in specified ways according to problem solving needs. A number of DR systems have been built in the last couple of decades, in each of which the developers have ascertained a priori and hand-coded the required perceptions and actions. This approach of building DR systems defeats the very purpose of open-ended exploration – the essence of human-like problem solving. Our goal, in this paper, was to develop a general and efficient framework for executing perceptions and actions as relevant to reasoning with $2D$ diagrams across a wide variety of domains and tasks. We make two important contributions:

1. We observe that the wide variety of visual perceptions/actions for DR applications can be transformed into domain/task-independent spatial problems. This observation makes it possible to use the well-established constraint satisfaction framework for spatial problem solving. We developed a language in which to specify spatial problems as QCSPs in the real domain using an open-ended vocabulary of properties, relations and actions involving three kinds of diagrammatic objects – points, curves, regions. Solution to a spatial problem is the equivalent simplified quantifier-free expression. That reduces the goal to developing a general and efficient SPS for solving $2D$ spatial problems without human intervention.

4. To be fair, REDLOG and QEPCAD were not developed for solving only spatial problems but any QCSP.

2. The spatial problems were specified as QCSPs in first-order logic. QE, an inherently doubly exponential problem, was the computational bottleneck of the SPS. We represented the objects (points, curves, regions) as configuration of simple elements to facilitate decomposition of complex problems into simpler and *similar* subproblems. We showed that, if the symbolic solution to a subproblem can be expressed concisely, QE can be achieved in low-order polynomial time by storing problems and their solutions in memory so that when a similar problem is encountered in future, it can be solved by mapping its solution from a similar previously solved problem. The SPS grows more efficient as it solves more problems. Even though we used the CAD algorithm for QE and compared the complexity results with that of CAD's, this approach is by no means limited to any particular algorithm. The complexity of any QE algorithm can be significantly improved for spatial problem solving by using the idea of problem decomposition and variable mapping, as discussed in this paper.

The framework leaves room to be more efficient and convenient by incorporating future results in at least two possible directions – learning constraints from examples (automatic constraint acquisition) and carefully exploiting a rich portfolio of QE algorithms for solving new problems.

## Acknowledgments

This research was partially supported by participation in the Advanced Decision Architectures Collaborative Technology Alliance sponsored by the U.S. Army Research Laboratory under Cooperative Agreement DAAD19-01-2-0009. We thank the anonymous reviewers for their constructive comments.